\pdfoutput=1   
\documentclass{article}

\usepackage[T1]{fontenc}
\usepackage{microtype}
\usepackage{graphicx}
\usepackage{subcaption}
\usepackage{booktabs}
\usepackage{longtable}   
\usepackage{hyperref}

\usepackage[preprint]{icml2026}

\usepackage{amsmath,amssymb,amsthm,mathtools}
\usepackage{enumitem}
\newcommand{\AmuseFirst}{0.23}
\newcommand{\AmuseLast}{0.50}
\newcommand{\BasePencilEightFirst}{0.47}
\newcommand{\BasePencilEightLast}{0.84}
\newcommand{\BasePhamEightMax}{1.15}
\newcommand{\BasePhamEightMin}{1.00}
\newcommand{\CodeRatioMax}{1.07}
\newcommand{\CodeRatioMin}{1.06}
\newcommand{\CodeSmin}{0.33}
\newcommand{\CodedMax}{512}
\newcommand{\CodedMin}{64}
\newcommand{\ContrastJacMax}{28\%}
\newcommand{\ConvBig}{662}

\newcommand{\LawDimOneFirst}{0.055}
\newcommand{\LawDimOneLast}{0.68}

\newcommand{\LawGapDevA}{8\%}
\newcommand{\LawGapDevB}{1\%}
\newcommand{\LawGapEnergyRatio}{3.8}
\newcommand{\LawGapErrRatio}{4.1}

\newcommand{\LowCappedPhrase}{all $54$ runs}
\newcommand{\LowConvLong}{37}
\newcommand{\LowCritGainMax}{0.19\%}
\newcommand{\LowDs}{32,64,128}
\newcommand{\LowLongChangeMax}{3\%}
\newcommand{\LowLongChangeMed}{0\%}
\newcommand{\LowN}{54}
\newcommand{\LowReps}{3}
\newcommand{\LowRestartSpreadMax}{14\%}
\newcommand{\LowRestartSpreadMed}{1\%}
\newcommand{\LowTwelvePredMed}{0.93}
\newcommand{\NBig}{684}

\newcommand{\PluginDs}{128,256}
\newcommand{\PluginPilotMax}{1.4\%}
\newcommand{\PluginPilotRel}{0.57}
\newcommand{\QPconvLat}{1.03}
\newcommand{\QPconvLatgamma}{0.028}

\newcommand{\SobiThreeMax}{0.020}
\newcommand{\SobiThreeMin}{0.007}
\newcommand{\SobiTwoFirst}{0.047}
\newcommand{\SobiTwoLast}{0.074}
\newcommand{\SobidMax}{128}
\newcommand{\SobidMin}{16}
\newcommand{\SweepsMedian}{13}
\newcommand{\UnequalJacMax}{41\%}
\newcommand{\WorstK}{4}
\newcommand{\WorstNmin}{22}
\newcommand{\WorstRatio}{400}
\newcommand{\WorstSEhi}{18\%}
\newcommand{\WorstSElo}{14\%}

\newcommand{\CodesCrossGrowth}{1.0}
\newcommand{\CodesCrossTarget}{0.01}
\newcommand{\CodesEwGrowth}{1.0}
\newcommand{\CodesK}{16}
\newcommand{\CodesRatioFourRange}{1.01}
\newcommand{\ContConv}{89}
\newcommand{\ContMaxChange}{5\%}
\newcommand{\ContMeanChange}{0.1\%}
\newcommand{\ContRuns}{90}
\newcommand{\DriftGapMinus}{0.0036}
\newcommand{\DriftGapMinusC}{0.0026}
\newcommand{\DriftGapPlus}{0.0013}
\newcommand{\DriftGapPlusC}{0.0024}
\newcommand{\DriftGapPop}{0.0025}
\newcommand{\DriftPredMinus}{0.0036}
\newcommand{\DriftPredPlus}{0.0014}
\newcommand{\DriftReps}{400}
\newcommand{\DriftRhoMinus}{0.10}
\newcommand{\DriftRhoMinusC}{0.17}
\newcommand{\DriftRhoPlus}{0.31}
\newcommand{\DriftRhoPlusC}{0.18}
\newcommand{\DriftSE}{0.0001}
\newcommand{\DriftTGapMinus}{0.00411}
\newcommand{\DriftTGapMinusC}{0.00279}
\newcommand{\DriftTGapMinusCB}{0.00278}
\newcommand{\DriftTGapPlus}{0.00127}
\newcommand{\DriftTGapPlusC}{0.00241}
\newcommand{\DriftTGapPlusCB}{0.00276}
\newcommand{\DriftTGapPop}{0.00279}
\newcommand{\DriftTPredMinus}{0.00411}
\newcommand{\DriftTPredPlus}{0.00147}
\newcommand{\DriftTReps}{300}
\newcommand{\DriftTResid}{2.4}
\newcommand{\DriftTSEB}{0.00003}
\newcommand{\FKEqualA}{1.16}
\newcommand{\FKEqualB}{1.14}
\newcommand{\FKPredEight}{1.04, 1.08, 1.15}
\newcommand{\FKPredTwo}{1.17, 1.31, 1.59}
\newcommand{\FKSimEight}{1.03, 1.07, 1.14}
\newcommand{\FKSimTwo}{1.16, 1.32, 1.67}

\newcommand{\LargeCrossFivedMin}{64}
\newcommand{\LargeCrossTwentyGrowth}{1.2}
\newcommand{\LargeDMax}{1000}
\newcommand{\LargeRepsMin}{6}
\newcommand{\LargeSecMax}{21}
\newcommand{\LargeSecMin}{6}
\newcommand{\LowPhamGainMax}{34\%}
\newcommand{\LowPhamGainMin}{28\%}
\newcommand{\LowPhamGainOtherMax}{13\%}
\newcommand{\LowPhamGainOtherMin}{5\%}
\newcommand{\MeanRiskCrossGrowth}{1.4}

\newcommand{\MeanRiskEightRatioMax}{1.04}
\newcommand{\MeanRiskEightRatioMin}{1.03}
\newcommand{\MeanRiskEwGrowth}{1.5}
\newcommand{\MeanRiskFiftyMax}{0.91}
\newcommand{\MeanRiskFiftyMin}{0.78}
\newcommand{\MeanRiskHundredLargeMax}{1.78}
\newcommand{\MeanRiskHundredLargeMin}{1.77}
\newcommand{\MeanRiskHundredPredMax}{0.46}
\newcommand{\MeanRiskHundredPredMin}{0.36}
\newcommand{\MeanRiskHundredRatioMax}{1.78}
\newcommand{\MeanRiskHundredRatioMin}{1.67}
\newcommand{\MeanRiskLowConv}{zero}
\newcommand{\MeanRiskLowRuns}{146}
\newcommand{\MeanRiskProvedGrowth}{11.1}
\newcommand{\MeanRiskQNinetyMax}{16\%}
\newcommand{\MeanRiskRatioMax}{1.13}
\newcommand{\MeanRiskRatioMin}{1.08}
\newcommand{\MeanRiskSEMax}{3\%}
\newcommand{\MeanRiskTwoFirst}{1.47}
\newcommand{\MeanRiskTwoLast}{1.22}
\newcommand{\MeanRiskTwoPredMax}{0.23}
\newcommand{\MeanRiskTwoPredMin}{0.15}
\newcommand{\MeanRiskdMax}{1000}
\newcommand{\MeanRiskdMin}{27}
\newcommand{\PencilInitAllConverged}{all}
\newcommand{\PencilInitConvMax}{0.024}
\newcommand{\PencilInitConvMin}{0.020}
\newcommand{\PencilInitConvRelDiffBound}{10^{-10}}
\newcommand{\PencilInitDmax}{256}
\newcommand{\PencilInitDs}{16, 32, 64, 128, 256}
\newcommand{\PencilInitMinimiserSinBound}{10^{-13}}
\newcommand{\PencilInitPencilFirst}{0.43}
\newcommand{\PencilInitPencilLast}{0.84}
\newcommand{\PencilInitReps}{six}
\newcommand{\PencilInitSweepOneFirst}{0.066}
\newcommand{\PencilInitSweepOneLast}{0.39}
\newcommand{\PencilInitSweepThreeMaxExcess}{13\%}
\newcommand{\PencilInitSweepTwoFirst}{0.024}
\newcommand{\PencilInitSweepTwoLast}{0.078}
\newcommand{\PencilInitSweepsMax}{17}
\newcommand{\RandCrossGrowth}{1.0}
\newcommand{\RandCrossTarget}{0.1}
\newcommand{\RandEwGrowth}{1.1}
\newcommand{\RandK}{4}
\newcommand{\RandRatioFourRange}{1.06--1.12}
\newcommand{\RanddMax}{512}
\newcommand{\RanddMin}{64}
\newcommand{\SegD}{32}
\newcommand{\SegFlatPRMin}{4.8}
\newcommand{\SegFlatRatioMax}{1.04}
\newcommand{\SegFourRatio}{1.07}
\newcommand{\SegKmax}{256}
\newcommand{\SegLastNk}{3.1}
\newcommand{\SegLastRatio}{1.34}
\newcommand{\SegRatio}{800}
\newcommand{\SegReps}{eight}
\newcommand{\SegRiseK}{128}
\newcommand{\SegRiseNk}{6.2}
\newcommand{\SegRiseRatio}{1.17}
\newcommand{\SegTwoLoss}{0.37}
\newcommand{\StepGammaEightMax}{4}
\newcommand{\StepGammaEightMin}{1.3}
\newcommand{\StepGammaFourMax}{100}
\newcommand{\StepLabEightMax}{2.13}
\newcommand{\StepLabEightMin}{1.82}
\newcommand{\StepLabFourMax}{4.84}
\newcommand{\StepLabFourMin}{3.13}
\newcommand{\StepPolarEightMax}{1.18}
\newcommand{\StepPolarEightMin}{1.09}
\newcommand{\StepPolarFourMax}{2.19}
\newcommand{\StepPolarFourMin}{1.40}
\newcommand{\StepStepEightMax}{1.06}
\newcommand{\StepStepEightMin}{1.00}
\newcommand{\StepStepFourMax}{1.42}
\newcommand{\StepStepFourMin}{0.99}
\newcommand{\StressArHalfMax}{1.17}
\newcommand{\StressArHalfMin}{1.02}
\newcommand{\StressArHighConv}{zero}
\newcommand{\StressArHighHigh}{1.11}
\newcommand{\StressArHighLow}{2.65}
\newcommand{\StressArHighNeff}{43}
\newcommand{\StressArHighTau}{4.6}
\newcommand{\StressContrastMax}{1.09}
\newcommand{\StressContrastMin}{0.99}
\newcommand{\StressCorrFloorHigh}{6}
\newcommand{\StressCorrFloorLow}{4}
\newcommand{\StressD}{64}
\newcommand{\StressK}{8}
\newcommand{\StressMixFloor}{22}
\newcommand{\StressMixFloorOverEps}{2.2}
\newcommand{\StressMixSmallMax}{1.22}
\newcommand{\StressReps}{eight}
\newcommand{\StressSpearmanMax}{0.78}
\newcommand{\StressSpearmanMin}{0.54}
\newcommand{\UnkEtoELowMax}{2.6}
\newcommand{\UnkEtoELowMin}{1.8}
\newcommand{\UnkLargeD}{216}
\newcommand{\UnkLargeGrid}{210{,}464}
\newcommand{\UnkLargeRuns}{two}
\newcommand{\UnkLargeTime}{9}
\newcommand{\UnkLowRatio}{1600}
\newcommand{\UnkMissMax}{10}
\newcommand{\UnkPilotDev}{3\%}
\newcommand{\UnkPilotDs}{27,64}
\newcommand{\UnkPilotFound}{in all six runs}
\newcommand{\UnkPilotRatioMax}{0.89}
\newcommand{\UnkPilotRatioMin}{0.80}
\newcommand{\UnkPilotRuns}{six}
\newcommand{\UnkPilotSminMax}{0.90}
\newcommand{\UnkPilotSminMin}{0.67}
\newcommand{\UnkProbeErrMax}{0.20}
\newcommand{\UnkRatioMin}{6400}
\newcommand{\UnkRatioOracleMax}{4\%}
\newcommand{\UnkSensDs}{27,64}
\newcommand{\UnkSensGridFactor}{7.9}
\newcommand{\UnkSensRuns}{six}
\newcommand{\UnkSensTimeFactor}{8}
\newcommand{\UnkTimeFirst}{4}
\newcommand{\UnkdMax}{216}
\newcommand{\WhCond}{41}
\newcommand{\WhEquivD}{27}
\newcommand{\WhEquivExp}{-13}
\newcommand{\WhEwh}{0.011}
\newcommand{\WhExcessHighMax}{6\%}
\newcommand{\WhExcessHighMin}{4\%}
\newcommand{\WhExcessLowMax}{15\%}
\newcommand{\WhExcessLowMin}{10\%}
\newcommand{\WhExcessMidOne}{2.9\%}
\newcommand{\WhExcessTopOne}{2.3\%}

\newcommand{\WhFoHalfOff}{11\%}
\newcommand{\WhFoRandEwh}{0.007}
\newcommand{\WhFoRandGamma}{0.03}
\newcommand{\WhFoRandMax}{1.02}
\newcommand{\WhFoRandMin}{1.00}
\newcommand{\WhFoReps}{30{,}000}
\newcommand{\WhFoTiltEwhMax}{0.15}
\newcommand{\WhFoTiltGamma}{3\cdot10^{-4}}
\newcommand{\WhFoTiltMax}{1.00}
\newcommand{\WhFoTiltMin}{0.99}
\newcommand{\WhFoTiltSE}{0.005}
\newcommand{\WhGammaHighMax}{0.08}
\newcommand{\WhGammaLowMax}{0.30}
\newcommand{\WhGammaLowMin}{0.22}
\newcommand{\WhPredExcessOne}{2.1\%}

\newcommand{\FmriBartlettL}{30}
\newcommand{\FmriD}{200}
\newcommand{\FmriDeltaBarMFourHundred}{0.723}

\newcommand{\FmriEnergyNNSpearman}{0.94}
\newcommand{\FmriFloorFitted}{128}
\newcommand{\FmriFloorMedian}{0.59}
\newcommand{\FmriFloorSpearman}{0.073}
\newcommand{\FmriFracLinMFifty}{26\%}
\newcommand{\FmriFracLinMTen}{8\%}
\newcommand{\FmriFracLinMTwoHundred}{64\%}
\newcommand{\FmriFracPointOneMFifty}{12\%}
\newcommand{\FmriFracPointOneMFourHundred}{46\%}
\newcommand{\FmriFramesEnv}{286176}
\newcommand{\FmriFramesSubjectEnv}{352}
\newcommand{\FmriGapSpearmanEnergyAllMsub}{0.84}
\newcommand{\FmriGapSpearmanEnergyLinMsub}{0.89}
\newcommand{\FmriGapSpearmanNNAllMsub}{0.78}
\newcommand{\FmriGapSpearmanNNLinMsub}{0.72}
\newcommand{\FmriGroupSweeps}{1000}
\newcommand{\FmriHardestI}{72}
\newcommand{\FmriHardestJ}{179}

\newcommand{\FmriLinCountMFifty}{52}
\newcommand{\FmriLinCountMFourHundred}{199}
\newcommand{\FmriMmax}{400}
\newcommand{\FmriMmin}{10}
\newcommand{\FmriMsub}{50}
\newcommand{\FmriN}{813}
\newcommand{\FmriNNShareMedian}{5\%}
\newcommand{\FmriNNShareQninety}{9\%}
\newcommand{\FmriNNShareQten}{2\%}
\newcommand{\FmriNrest}{799}
\newcommand{\FmriNtau}{50}

\newcommand{\FmriPairCrossedMFifty}{0.071}
\newcommand{\FmriPairMatchedMFifty}{0.091}
\newcommand{\FmriPhamLossMax}{0.97}
\newcommand{\FmriPredBarMFourHundred}{0.118}
\newcommand{\FmriPredLin}{0.3}
\newcommand{\FmriPredLinHalf}{0.15}
\newcommand{\FmriPredNmax}{350907}
\newcommand{\FmriPredNmin}{734}
\newcommand{\FmriPredPointOne}{0.1}
\newcommand{\FmriRatioLinMedian}{6.1}
\newcommand{\FmriRestDelta}{0.894}
\newcommand{\FmriRestPR}{1.54}
\newcommand{\FmriRsplit}{50}
\newcommand{\FmriSlopeLin}{0.731}
\newcommand{\FmriSmin}{0.0000570}
\newcommand{\FmriSpearmanLin}{0.84}
\newcommand{\FmriSpearmanLinMFifty}{0.92}
\newcommand{\FmriSpearmanLinMFourHundred}{0.88}
\newcommand{\FmriSplitSweeps}{300}
\newcommand{\FmriSubsetCount}{70}
\newcommand{\FmriSubsetHighDelta}{0.909}
\newcommand{\FmriSubsetHighEbar}{2769}
\newcommand{\FmriSubsetLowDelta}{0.884}
\newcommand{\FmriSubsetLowEbar}{440}
\newcommand{\FmriTaskDeltaSameM}{0.860}
\newcommand{\FmriTauARrelMedian}{2.3\%}
\newcommand{\FmriTauDiagMedian}{1.19}

\theoremstyle{plain}
\newtheorem{theorem}{Theorem}
\newtheorem{proposition}{Proposition}
\newtheorem{lemma}{Lemma}
\newtheorem{corollary}{Corollary}
\newtheorem{conjecture}{Conjecture}
\newtheorem{theoremfull}{Theorem}[section]   
\theoremstyle{definition}
\newtheorem{definition}{Definition}
\newtheorem{remark}{Remark}

\newcommand{\R}{\mathbb{R}}
\newcommand{\E}{\mathbb{E}}
\newcommand{\Prob}{\mathbb{P}}
\newcommand{\Od}{\mathrm{O}(d)}
\newcommand{\off}{\operatorname{off}}
\newcommand{\diag}{\operatorname{diag}}
\newcommand{\Ebar}{\bar{E}}
\newcommand{\Emax}{E_{\max}}
\newcommand{\lA}{\ell_{\mathrm{A}}}
\newcommand{\lM}{\ell_{\mathrm{M}}}
\newcommand{\sig}{\mathbf{p}}          
\newcommand{\Sh}{\hat\Sigma}
\newcommand{\Npool}{\bar N^{(0)}}    
\newcommand{\NpoolW}{\bar N^{(W)}}   
\DeclareMathOperator{\KL}{KL}
\DeclareMathOperator{\Var}{Var}
\DeclareMathOperator{\tr}{tr}

\icmltitlerunning{Separation Is a Riesz Energy}

\begin{document}

\twocolumn[
\icmltitle{Separation Is a Riesz Energy: \texorpdfstring{\\}{ } How Many Environments Does ICA Need?}

\icmlsetsymbol{equal}{*}
\begin{icmlauthorlist}
\icmlauthor{Yuda Bi}{equal,trends}
\icmlauthor{Yubo Guo}{equal,hust}
\icmlauthor{Jiaxiao Cai}{upenn}
\icmlauthor{Xiaonai Li}{lancaster}
\icmlauthor{Zhuolin Li}{gatech}
\icmlauthor{Vince D. Calhoun}{trends}
\end{icmlauthorlist}
\icmlaffiliation{trends}{Tri-institutional Center for Translational Research in Neuroimaging and Data Science (TReNDS), Georgia State University, Georgia Institute of Technology, Emory University, Atlanta, GA, USA}
\icmlaffiliation{hust}{Huazhong University of Science and Technology, Wuhan, China}
\icmlaffiliation{upenn}{University of Pennsylvania, Philadelphia, PA, USA}
\icmlaffiliation{lancaster}{Lancaster University, Lancaster, UK}
\icmlaffiliation{gatech}{Georgia Institute of Technology, Atlanta, GA, USA}
\icmlcorrespondingauthor{Yuda Bi}{ybi3@gsu.edu}
\icmlkeywords{blind source separation, independent component analysis, joint diagonalization, multi-environment, Riesz energy, minimax}
\vskip 0.3in
]
\printAffiliationsAndNotice{\icmlEqualContribution}

\begin{abstract}
How many environments and samples does it take to separate $d$ Gaussian sources whose variances change across environments? Each source has a signature, its vector of variances. The Riesz $2$-energy $E_j$ of source $j$ sums the inverse Fisher informations for rotating it towards other sources. With known whitening, we prove a worst-case squared-angle risk lower bound of order $E_j/(n+E_j)$ for orthogonal estimators at every $n$ and $d$. A quadratic pencil given the signatures attains $E_j/n$ up to a constant above a separation-dependent sample threshold, even with estimated whitening. For well-spread signatures at fixed $K$, the energy grows as $d^2$, $d\log d$ and $d$ for $K=2$, $3$ and $K\ge4$, respectively. A design with $K\asymp\log d$ and pairwise Fisher information bounded below by a constant makes $d\log d$ samples sufficient. The eigenvectors of a single fixed linear combination have an average first-order error coefficient of at least order $d^2$. Simulations up to $d=1000$ support the first-order predictions, and on task fMRI the energy helps rank split-half instability.
\end{abstract}

\section{Introduction}
\label{sec:intro}

\begin{figure*}[t]
\centering
\includegraphics[width=0.92\textwidth]{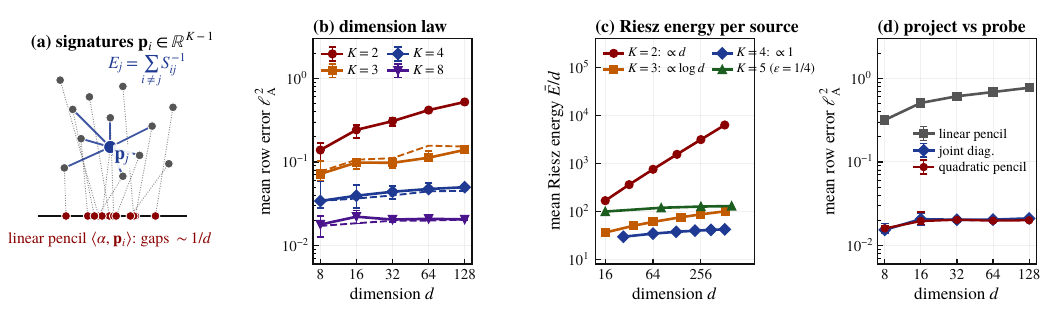}
\caption{\textbf{Separation is a Riesz energy.} \textbf{(a)}~Each source is a point $\sig_i$ in signature space. A quadratic pencil centred at $\sig_j$ uses distances to $\sig_j$ in that space (blue). A linear pencil $\sum_k\alpha_k\Sigma_k$ projects the signatures onto a line (red). \textbf{(b)}~Joint diagonalization with random signatures at $n=400d$. Solid curves show the median of the source-averaged loss $\lA^2$ over runs. Dashed curves show $\Ebar^{\rm w}/n$ (Proposition~\ref{prop:ajd}). \textbf{(c)}~For lattice signatures, $\Ebar/d$ grows like $d$ at $K=2$ and $\log d$ at $K=3$, and levels off at $K\ge4$. \textbf{(d)}~On identical data ($K=8$, $n=400d$), the error of one random linear pencil grows with $d$, while joint diagonalization and a quadratic-pencil variant stay nearly flat. The variant uses known signatures, same-sample whitening and the polar factor of the estimated directions (Appendix~\ref{app:exp}).}
\label{fig:story}
\end{figure*}

Independent component analysis (ICA; \citealp{hyvarinen2001independent}) recovers independent sources from a linear mixture $x=As$, with $A\in\R^{d\times d}$ invertible. Non-Gaussian sources are identified from i.i.d.\ samples by higher-order statistics \citep{comon1994independent}. Gaussian sources are not: i.i.d.\ samples determine $A$ only up to a rotation. \emph{Diversity} rescues them \citep{tong1991indeterminacy,adali2014diversity}. Suppose the source variances change across $K$ environments, for example across segments of a nonstationary recording \citep{matsuoka1995neural,pham2001blind} or datasets \citep{anderson2014independent}. The covariances $\Sigma_k=A\Lambda_kA^\top$ then share the mixing, and $A$ is identifiable from second-order statistics alone unless two sources share a variance profile up to scale, i.e.\ a signature. The same idea underlies nonlinear ICA with auxiliary variables and causal representation learning, where environments or interventions must be varied enough (``sufficient variability'') \citep{hyvarinen2016unsupervised,hyvarinen2019nonlinear,khemakhem2020variational,squires2023linear,jin2024learning}. Such conditions are qualitative: they say when identification is possible, not how many samples it takes or which environments to collect (finite-sample results: Appendix~\ref{app:related}).

For non-Gaussian sources, \citet{auddy2025large} established a minimax squared-angle rate of $d/n$ and a low-degree computational lower bound at $n\lesssim d^2$. For the diversity route, classical theory gives fixed-dimensional asymptotic variances \citep{pham2001blind,tichavsky2009fast,yeredor2010blind,miettinen2016separation,afsari2008sensitivity}, and \citet{chang2018principal} study a related time-series dimension reduction with growing dimension. The finite-sample cost of separating Gaussian sources is governed by how their variance signatures spread. This matters at the dimensions used in group fMRI ICA, where pipelines estimate $100$--$300$ components \citep{calhoun2001method,calhoun2012multisubject,du2020neuromark,iraji2023identifying}.

\paragraph{This paper.} Second-order statistics see source $j$ only through its \emph{signature} $\sig_j=(\sig_j(1),\dots,\sig_j(K))$, the vector of its variances across environments. We show that the diversity route is governed by the geometry of these $d$ points. Let
\begin{equation}\label{eq:S}
S_{ij}=\frac1K\sum_{k=1}^K\frac{\big(\sig_i(k)-\sig_j(k)\big)^2}{\sig_i(k)\,\sig_j(k)},
\qquad
E_j=\sum_{i\neq j}S_{ij}^{-1}.
\end{equation}
Here $S_{ij}$ is the Fisher information per sample, averaged over the environments, about the rotation that mixes sources $i$ and $j$. $E_j$ is the Riesz $2$-energy of the point $\sig_j$ in the configuration of signatures. A toy case shows why the dimension of the configuration matters. With two environments the signatures are $d$ numbers in an interval. Evenly spaced, they are $1/d$ apart, and every $E_j$ is of order $d^2$. With four, they are $d$ points in a $3$-dimensional body. Evenly spread, neighbours are $d^{-1/3}$ apart and the many far sources dominate the inverse-square sum, so $E_j$ is of order $d$.

\paragraph{Contributions.}
\begin{enumerate}[leftmargin=*,itemsep=1pt,topsep=2pt]
\item \textbf{Separation costs a Riesz energy (Section~\ref{sec:rate}).} For Gaussian sources and every $n$ and $d$, no estimator in $\Od$ recovers the $j$-th mixing direction with worst-case squared-angle risk below $c\,E_j/(n+E_j)$ (Theorem~\ref{thm:rate}). Conversely, a pencil estimator attains $E_j/n$ up to a constant for every mixing, uniformly in $d$, once $n\gtrsim K(d+\log n)/S_{\min}$, where $S_{\min}$ is the smallest divergence (Theorem~\ref{thm:upper}). It needs the signatures and can use estimated whitening. Given only a lower bound $s_0\le S_{\min}$, a grid of probes attains the average rate above the threshold with $s_0$ in the denominator (Theorem~\ref{thm:unknown}). In this regime the energy determines the minimax rate (Corollary~\ref{cor:minimax}).
\item \textbf{Dimension, not number (Section~\ref{sec:law}).} Two environments can identify the model when all signatures differ. With well-spread signatures, the energy grows as $d^2$, $d\log d$ and $d$ for $K=2$, $3$ and fixed $K\ge4$ (Theorem~\ref{thm:klaw}). These are necessary sample-size orders for every configuration (Corollary~\ref{cor:necessary}). A sufficient size of order $d\log d$ is proved for $K\asymp\log d$ environments with constant-order $S_{\min}$. Eight environments whose signatures are well-spread on a line have the same energy scaling as two (Remark~\ref{rem:dim}, Figure~\ref{fig:mech}(a)).
\item \textbf{Projecting versus probing (Section~\ref{sec:algo}).} The eigenvectors of one combination $\sum_k\alpha_k\Sh_k$ (AMUSE, generalised eigenproblems, random pencils) have a first-order error whose mean over sources is $\gtrsim d^2/n$ for every $K$ (Theorem~\ref{thm:pencil}): the combination projects the signatures onto a line. Probing one signature at a time avoids this. The first-order error of joint diagonalization is again of Riesz type (Proposition~\ref{prop:ajd}).
\end{enumerate}
Simulations up to $d=1000$ support the first-order predictions. The ratio $n/d$ needed by Jacobi to reach a target error grows slowly in our tests (Section~\ref{sec:experiments}, Conjecture~\ref{conj:linear}). On task fMRI, the energy helps rank split-half instability, while the first-order prediction underestimates its magnitude.

\paragraph{Relation to prior work.} Inverse-square error formulas are classical at fixed $d$: for eigenvectors \citep{anderson1963asymptotic} and, pair by pair, for joint diagonalization \citep{cardoso1994perturbation,miettinen2016separation}. The structure is visible in \citet[Lemma~1]{kuleshov2015tensor}, and \citet{afsari2007what,afsari2008sensitivity} showed that nearly collinear signatures make joint diagonalization ill-conditioned. What was missing is a quantitative role for the \emph{dimension} of the configuration, and a minimax theory that holds as $d$ grows. Matrices $\sum_k(X_k-\lambda_kI)^2$ are a known numerical device for approximate joint eigenvectors \citep{loring2022computing,cerjan2023quadratic}; we supply their statistics. Table~\ref{tab:landscape} summarises the rates, and Appendix~\ref{app:related} discusses further related work.

\paragraph{Why this model.} We work in the linear Gaussian model with $K$ environments, the common-principal-components model of \citet{flury1984common}. First, it is the least informative case. Its lower bounds transfer to every class of sources containing the Gaussian ones, and within a fixed non-Gaussian family the Fisher information of a rotation is at least as large (Appendix~\ref{app:FI}). Second, it is the case in which upper and lower bounds match (Corollary~\ref{cor:minimax}).

\begin{table*}[t]
\caption{Energy, sample-size thresholds and first-order error. The energy and sufficient thresholds assume well-spread signatures at fixed $K$ or constant-order $S_{\min}$ at $K\asymp\log d$. The necessary thresholds apply to every configuration. The finite-sample bounds use Gaussian sources, with equal environment sizes for the upper bounds. The last two rows give fixed-dimensional first-order scales. See the appendix reader's guide for the full conditions.}
\label{tab:landscape}
\centering\footnotesize\setlength{\tabcolsep}{4pt}
\begin{tabular}{lccccl}\toprule
 & $K=2$ & $K=3$ & fixed $K\ge4$ & $K\!\asymp\!\log d$, separated & signatures\\\midrule
energy $E_j$ (T\ref{thm:klaw}; Sec.~\ref{sec:law}) & $d^2$ & $d\log d$ & $d$ & $d$ & geometry only\\
necessary: $n\gtrsim$ (T\ref{thm:rate}, C\ref{cor:necessary}) & $d^2$ & $d\log d$ & $d$ & $d$ & may be used\\
sufficient: $\asymp\Ebar/n$ for $n\gtrsim$ (T\ref{thm:upper}, C\ref{cor:minimax}) & $d^3$ & $d^2$ & $d^{\frac{K+1}{K-1}}$ & $d\log d$ & known\\
\quad without signatures (T\ref{thm:unknown}) & $d^3$ & $d^2$ & $d^{\frac{K+1}{K-1}}$ & $d\log d$ & $s_0\asymp S_{\min}$\\
one linear pencil (T\ref{thm:pencil}, first order) & \multicolumn{4}{c}{$\gtrsim d^2/n$ for every $K$} & not used\\
joint diagonalization (P\ref{prop:ajd}, first order) & \multicolumn{4}{c}{$\asymp\Ebar/n$} & not used\\\bottomrule
\end{tabular}
\end{table*}

\section{Setting and signature geometry}
\label{sec:setting}

\paragraph{Model.} There are $K$ environments. In environment $k$ we observe $n_k=n/K$ independent copies of $x^{(k)}=A\,s^{(k)}$, with invertible $A\in\R^{d\times d}$ and independent, centred coordinates $s^{(k)}_j$ of variance $\sig_j(k)$. We normalise $\frac1K\sum_k\sig_j(k)=1$ and assume $\sig_j(k)\in[1/B,B]$, so the pooled covariance is $AA^\top$. After whitening, $A=Q\in\Od$ and $\Sigma_k=Q\Lambda_kQ^\top$ with $\Lambda_k=\diag(\sig_1(k),\dots,\sig_d(k))$; $\Sh_k$ is the sample covariance of environment $k$.

\paragraph{Scope.} Unequal sizes $n_k=\pi_kn$ replace $\frac1K\sum_k$ by $\sum_k\pi_k$ in the lower bounds, the first-order results and the dimension law; an environment with very few samples (two in the example of Appendix~\ref{app:FI}) can, however, make the estimators of the upper bounds fail.

\paragraph{Loss.} Estimators take values in the orthogonal group $\Od$, as every orthogonal joint diagonalizer does, and may be randomised. For an estimate $\hat Q\in\Od$ with columns $\hat q_a$ and the truth $Q$ with columns $q_j$, let
\begin{equation}
\rho_j(\hat Q)=1-\max_a\langle\hat q_a,q_j\rangle^2,\qquad \lA^2=\frac1d\sum_{j=1}^d\rho_j,
\end{equation}
and $\lM^2=\max_j\rho_j$. So $\rho_j$ is the squared sine of the angle between $q_j$ and the closest estimated direction, in whitened coordinates, and $\lA^2$, the average of the $\rho_j$ over sources, is permutation invariant. When the signatures are known, an estimator may also be \emph{labelled}: one unit vector $\hat q_j$ per source, with $\rho_j=1-\langle\hat q_j,q_j\rangle^2$. With estimated whitening and a general mixing, $\rho_j$ is the same quantity in source coordinates: the variance share of the other sources in the estimated source (Theorem~\ref{thm:upper}(b)). This closest-direction loss is a lower bound for the one-to-one permutation-matched loss. They agree when all $\rho_j<1/2$, since no two sources then share a closest estimate (Appendix~\ref{app:FI}). \emph{Notation:} $\mathrm{sym}(X)=\frac12(X+X^\top)$; $a\lesssim b$ means $a\le Cb$ with $C$ depending only on the stated constants, and $a\asymp b$ means both $a\lesssim b$ and $b\lesssim a$; $\Rightarrow$ is convergence in law; a polar factor is a nearest orthogonal matrix.

\paragraph{Signatures.} The signature $\sig_j\in\R^K$ lies on the hyperplane $\sum_k\sig_j(k)=K$, a $(K-1)$-dimensional space. Each summand of \eqref{eq:S} equals $r+1/r-2$ with $r=\sig_i(k)/\sig_j(k)$, the symmetrised Itakura--Saito divergence of the two variances, so under bounded ratios
\begin{equation}\label{eq:S-euclid}
\frac{\|\sig_i-\sig_j\|^2}{KB^2}\le S_{ij}\le\frac{B^2\|\sig_i-\sig_j\|^2}{K}:
\end{equation}
$S_{ij}$ is comparable to a squared Euclidean distance, and $E_j$ to a Euclidean Riesz $2$-energy. Each such term is at most $B^2+B^{-2}-2$, so $E_j\ge(d-1)/(B^2+B^{-2}-2)$ for every configuration and every $K$. Moreover, $S_{ij}=0$ iff the two signatures coincide, which is exactly the non-identifiable case of \citet{pham2001blind}. We write $\Ebar=\frac1d\sum_jE_j$, $\Emax=\max_jE_j$, $S_{\rm nn}(j)=\min_{i\neq j}S_{ij}$ for a fixed source $j$, and $S_{\min}=\min_jS_{\rm nn}(j)$. For Gaussian sources the Fisher-information reading of $S_{ij}$ is exact: the Kullback--Leibler divergence between the laws under $Q$ and $QR$, with $R$ a rotation by $\theta$ in the plane $(i,j)$, is $\frac n2\sin^2\theta\,S_{ij}$, and the information is diagonal across planes and orthogonal to the variances (Lemma~\ref{lem:FI}, Appendix~\ref{app:FI}).

\section{The Riesz rate}
\label{sec:rate}

\subsection{A lower bound for every \texorpdfstring{$n$}{n} and \texorpdfstring{$d$}{d}}

\begin{theorem}[Riesz rate]\label{thm:rate}
For Gaussian sources and $K\ge2$ there is $c>0$, depending only on $B$, such that for every $n$ and $d$, every source $j$ and every estimator $\hat Q\in\Od$, which may use the signatures,
\begin{equation}\label{eq:lb2}
\sup_{Q\in\Od}\E\,\rho_j(\hat Q)\;\ge\;c\,\frac{E_j}{n+E_j}.
\end{equation}
The same bound holds for $\lA^2$ and for $\lM^2$, with $E_j/(n+E_j)$ replaced by its average and by its maximum over $j$. If $S_{ij}=0$ for some $i\neq j$, then $E_j=\infty$ and the bound reads $c$. (Full statement: Theorem~\ref{thm:rate-full}, Appendix~\ref{app:rate}.)
\end{theorem}

The first-order error adds the contributions of pairwise rotations. The nearest signature sets the largest term, and the remaining sources also contribute. The lower bound gives risk of constant order when $n\lesssim E_j$. At fixed $d$ with distinct signatures, the Gaussian maximum-likelihood joint diagonalizer has limit law $n\rho_j\Rightarrow\sum_{i\neq j}S_{ij}^{-1}Z_i^2$, with $Z_i$ i.i.d.\ standard normal, giving a first-order coefficient of $E_j$ (Theorem~\ref{thm:rate-full}(b); \citealp{flury1986asymptotic,pham2001blind}). To obtain the finite-sample bound, we apply Assouad's lemma to rotations in the planes $(j,i)$ ordered by decreasing $S_{ji}$. The construction extends single-covariance bounds \citep{wahl2022lower,wahl2023vantrees} to information across environments and to the closest-direction loss. Haar symmetrisation yields the average bound, and all three bounds also hold for labelled estimators (Appendix~\ref{app:rate}).

\subsection{Attainment by a quadratic pencil}\label{sec:upper}

To attain the bound uniformly in $d$ we probe one signature at a time. For source $j$, let $H_j=\sum_kw_k(\Sigma_k-\sig_j(k)I)^2$. Then $H_j=Q\diag(\lambda_a)Q^\top$ with $\lambda_a=\sum_kw_k(\sig_a(k)-\sig_j(k))^2$, zero at $a=j$. With $w_k=\sig_j(k)^{-2}$, $B^{-2}KS_{aj}\le\lambda_a\le B^2KS_{aj}$: the gap at $q_j$ is the nearest-neighbour divergence, not a spacing on a line. Algorithm~\ref{alg:pencil} computes a sample version from two parts of the data and estimates the whitening on a third.

\begin{algorithm}[t]
\caption{Quadratic pencil at a probe $g\in[1/B,B]^K$ (Theorem~\ref{thm:upper}(b); $g=\sig_j$ for known signatures)}\label{alg:pencil}
\begin{algorithmic}
\STATE split each environment's sample into three equal parts $0$, $1$, $2$; $\hat C\leftarrow$ pooled covariance of part $0$
\STATE $\Sh^{(i)}_k\leftarrow$ covariance of part $i\in\{1,2\}$ in environment $k$, whitened by $\hat C^{-1/2}$
\STATE $\hat H\leftarrow\sum_kg(k)^{-2}\,\mathrm{sym}\big[(\Sh^{(1)}_k-g(k)I)(\Sh^{(2)}_k-g(k)I)\big]$
\STATE $\hat q\leftarrow$ unit eigenvector of $\hat H$ for its algebraically smallest eigenvalue (Lanczos: $O(d^2)$ per iteration)
\STATE \textbf{return} the unmixing direction $\hat b=\hat C^{-1/2}\hat q$ (with known whitening: no part $0$, return $\hat q$)
\end{algorithmic}
\end{algorithm}

\begin{theorem}[Attainment, uniformly in $d$; simplified]\label{thm:upper}
Let the sources be Gaussian and the signatures known. Run Algorithm~\ref{alg:pencil} at $g=\sig_j$ and denote the unit eigenvector it computes by $\hat q_j$. Split each environment's sample into two halves in (a) or three equal parts in (b). Let $\gamma_j=K(d+6\log n)/(nS_{\rm nn}(j))$. There are $c,C>0$, depending only on $B$, such that the following hold if $\gamma_j\le c$.
\begin{enumerate}[leftmargin=*,itemsep=1pt,topsep=2pt]
\item[(a)] (Known whitening.) For $A=Q\in\Od$, $\E\big[1-\langle\hat q_j,q_j\rangle^2\big]\le C\,E_j/n$.
\item[(b)] (Estimated whitening.) Let $A$ be invertible, let $\hat C$ be the pooled covariance of part $0$ (Algorithm~\ref{alg:pencil}), and let $\hat b_j=\hat C^{-1/2}\hat q_j$. Then $\E\rho_j\le C\,E_j/n$, where $\rho_j$ is the share of the pooled variance of the estimated source $\hat b_j^\top x$ due to the other sources.
\end{enumerate}
(Full statement, with the first-order coefficients and the cost of estimating the whitening: Theorem~\ref{thm:upper-full}, Appendix~\ref{app:upper}.)
\end{theorem}

The product of covariances from independent splits removes the bias from squaring covariance noise. Its leading noise in direction $a$ scales as $\sqrt{\lambda_a}$, allowing a relative perturbation bound controlled by the nearest-neighbour gap \citep[cf.][]{jirak2018relative}. Dividing the off-diagonal noise entry by $\lambda_a$ gives first-order variance of order $1/(nS_{aj})$. Summing over $a\ne j$ gives $E_j/n$. The rate also holds for an independent probe sufficiently close to $\sig_j$, which allows the next theorem to estimate the probes from a separate sample (Lemma~\ref{lem:probe}, Appendix~\ref{app:upper}).

\begin{theorem}[Unknown signatures; simplified]\label{thm:unknown}
Let the sources be Gaussian, and let only $B$ and a lower bound $s_0\le S_{\min}$ be known. Algorithm~\ref{alg:grid} (Appendix~\ref{app:unknown}) scores every point $g$ of a grid in signature space by $\lambda_{\min}\big(\sum_k(\Sh_k-g(k)I)^2\big)$ on one part of the sample, and keeps the points below a threshold, which, under the condition below, form one cluster per source. It then runs Algorithm~\ref{alg:pencil} on the other parts, at one probe per cluster. Let $\gamma_0=K(d+6\log n)/(ns_0)$. There are $c,C>0$, depending only on $B$, such that the following hold for every configuration with $S_{\min}\ge s_0$ if $\gamma_0\le c$.
\begin{enumerate}[leftmargin=*,itemsep=1pt,topsep=2pt]
\item[(a)] (Known whitening.) For $A=Q\in\Od$, the polar factor $\hat Q$ of the $d$ vectors of Algorithm~\ref{alg:pencil} satisfies $\sup_Q\E\lA^2(\hat Q)\le C\,\Ebar/n$.
\item[(b)] (Estimated whitening.) With the whitening estimated on a further part, $\E\lA^2\le C\,\Ebar/n$ for every invertible $A$, under the closest-direction loss $\lA^2=\frac1d\sum_j\min_a(1-z_{a,j}^2/\|z_a\|^2)$ with $z_a=A^\top\hat b_a$.
\end{enumerate}
The grid has at most $(C/\sqrt{s_0})^K$ points. (Full statement: Theorem~\ref{thm:unknown-full}.)
\end{theorem}

A grid on the signature hyperplane has $O(d)$ points for well-spread signatures at fixed $K$ when $s_0\asymp S_{\min}$ (Appendix~\ref{app:unknown}). For separated signatures, a constant $s_0$ gives $d^{O(1)}$ points when $K\asymp\log d$. Section~\ref{sec:experiments} tests a coarser grid with local refinement.

\begin{corollary}[Minimax rate]\label{cor:minimax}
Let the sources be Gaussian and $\gamma=K(d+6\log n)/(nS_{\min})\le c$. Then, for every configuration, the minimax risk is $\asymp\Ebar/n$ with constants depending only on $B$, in three settings. (i) Labelled estimators and any invertible mixing, with $\lA^2$ the average of the $\rho_j$ of Theorem~\ref{thm:upper}(b). (ii) Known whitening, with labelled estimators or estimators in $\Od$. (iii) Any invertible mixing and estimators that return $d$ unmixing directions, with the closest-direction loss of Theorem~\ref{thm:unknown}(b). The upper bounds are attained by Theorem~\ref{thm:upper}(b) and by the polar factor of the vectors of Theorem~\ref{thm:upper}(a). Without the signatures, the estimator of Theorem~\ref{thm:unknown} with $s_0=S_{\min}$ attains the bounds of (ii) in $\Od$ and of (iii) with the closest-direction loss.
\end{corollary}

The lower bound follows from Theorem~\ref{thm:rate}, using $\Emax\le d/S_{\min}\le n$ under $\gamma\le c$. Lemma~\ref{lem:matched} extends the average lower bound to the non-orthogonal estimates in (iii).

\section{The dimension law}
\label{sec:law}

\begin{figure}[t]
\centering
\includegraphics[width=0.92\columnwidth]{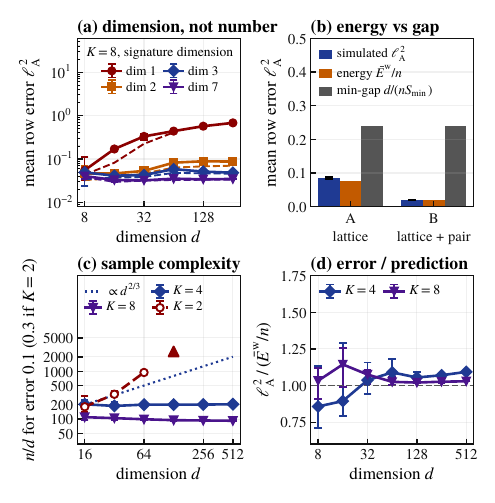}
\caption{\textbf{The mechanism.} \textbf{(a)}~\emph{Dimension, not number:} $K=8$, signatures well-spread (Halton points, fixed scale) in an affine subspace of dimension $1$, $2$, $3$ or $7$; $n=1600d$; dashed: $\Ebar^{\rm w}/n$. \textbf{(b)}~\emph{Energy beats the smallest gap:} two configurations with the same $S_{\min}$ ($K=4$, $d=27$, $n=400d$), a lattice (A) and a coarser lattice plus one close pair (B): simulated $\lA^2$ with standard errors, $\Ebar^{\rm w}/n$ and $(d/n)/S_{\min}$. \textbf{(c)}~\emph{Sample complexity:} $n/d$ at which the median $\lA^2$ reaches $0.1$ ($K=4,8$) or $0.3$ ($K=2$; triangle: not reached by $n=1600d$), from Figure~\ref{fig:collapse} (Appendix~\ref{app:exp}), with 90\% bootstrap intervals; dotted: the growth $d^{2/(K-1)}$ of the proved threshold, $K=4$. \textbf{(d)}~\emph{First order explains the curves:} error divided by $\Ebar^{\rm w}/n$ at $n=400d$, per draw (draws: Appendix~\ref{app:exp}).}
\label{fig:mech}
\end{figure}

The dimension law follows by counting signatures in shells around source $j$. In $D$ dimensions a well-spread configuration has about $d\,t^{D-1}\,dt$ signatures in a shell at distance $t$, each contributing order $t^{-2}$ at fixed $K$. Integrating from the typical nearest-neighbour scale $d^{-1/D}$ gives order $d$ for $D>2$, $d\log d$ for $D=2$ and $d^{2/D}$ for $D<2$. These are the classical Riesz energy regimes, expressed per source \citep{hardin2004discretizing,hardin2005minimal,borodachov2019discrete}. Here $D=K-1$ when signatures fill the normalisation hyperplane.

\begin{definition}[Well-spread signatures]\label{def:qu}
Let $r_d=d^{-1/(K-1)}$ and let $\Omega$ be a fixed convex body in $\{\sum_k\sig(k)=K\}\cap[1/B,B]^K$ containing the signatures. The signatures are \emph{well-spread} with constants $(c_1,C_1)$ if they are $c_1r_d$-separated and every point of $\Omega$ is within $C_1r_d$ of one of them.
\end{definition}
The covering condition rules out voids, which would make the per-source lower bound fail for $K\le3$; lattices in a box satisfy it.

\begin{theorem}[Dimension law]\label{thm:klaw}
Fix $K$. For well-spread signatures, uniformly in $j$ and with constants depending only on $(K,B,c_1,C_1,\Omega)$,
\begin{equation}
E_j\;\asymp\;\begin{cases} d^2, & K=2,\\ d\log d, & K=3,\\ d, & K\ge4.\end{cases}
\end{equation}
The upper bounds need only the separation $c_1r_d$, and the lower bounds for $\Ebar$ hold for \emph{every} configuration, by minimal-energy asymptotics \citep{hardin2005minimal}.
\end{theorem}

The constants depend on $K$ (Figure~\ref{fig:story}(c) shows the three regimes on lattices). A lattice has $m^{K-1}$ points, so the well-spread condition of Definition~\ref{def:qu} is met only once $2^{K-1}\lesssim d$. For $K\ge4$ the lower bound is the pair count $E_j\ge(d-1)/(B^2+B^{-2}-2)$; the geometry enters the upper bound, and the lower bounds only for $K=2,3$.

\begin{remark}[Dimension, not number]\label{rem:dim}
If the signatures lie in an affine subspace of dimension $D'$, Theorem~\ref{thm:klaw} holds with $D'$ in place of $K-1$ (Definition~\ref{def:qu} taken in that subspace, $r_d=d^{-1/D'}$). Corollary~\ref{cor:necessary} then holds with $D'+1$ for $K$. Appendix~\ref{app:klaw} reads off an effective dimension at each scale.
\end{remark}
Combined with Theorem~\ref{thm:rate}, the lower bounds give necessary sample sizes for every configuration.

\begin{corollary}[Necessary sample size]\label{cor:necessary}
Let $d\ge2$ and $f_K(d)=d^2$, $d\log d$, $d$ for $K=2$, $3$, $\ge4$. For Gaussian sources there is $c>0$ depending only on $(K,B)$ such that, for every configuration of signatures and every estimator $\hat Q\in\Od$,
\begin{equation*}
\sup_Q\E\,\lA^2(\hat Q)\ \ge\ c\,\min\big(1,\ f_K(d)/n\big).
\end{equation*}
No estimator recovers an average source consistently unless $n/f_K(d)\to\infty$, however the environments are chosen (under the variance bounds of Section~\ref{sec:setting}).
\end{corollary}
The proof (Appendix~\ref{app:klaw}) shows that in every configuration at least half of the sources have $E_j\gtrsim f_K(d)$. For well-spread signatures $\Ebar\asymp f_K(d)$, so the bound has the order of the maximum-likelihood limit $\Ebar/n$. For i.i.d.\ random signatures with a density bounded above and below, $\Ebar$ has the same orders in probability for $K\ge3$ ($d^3$ for $K=2$). The worst source, however, needs five environments to be of order $d$: $\Emax\asymp d^{4/3}$ in probability for $K=4$ (Proposition~\ref{prop:random}, Appendix~\ref{app:klaw}).

\paragraph{Sufficient sample sizes.} At fixed $K$, well-spread signatures have $S_{\min}\asymp d^{-2/(K-1)}$. The minimax rate $\asymp\Ebar/n$ is therefore attained once $n\gtrsim d^{(K+1)/(K-1)}$, for example $n\gtrsim d^{5/3}$ at $K=4$. A constant lower bound on $S_{\min}$ requires $K\gtrsim\log d$ by packing. It is achieved at $K\asymp\log d$ by balanced binary codes (Appendix~\ref{app:upper}). Then $E_j\asymp d$, and the threshold $n\gtrsim K(d+\log n)/s_0$ is of order $d\log d$ with constant $s_0$.

The proved sufficient threshold for fixed $K$ is polynomially larger than the necessary order $f_K(d)$. For well-spread signatures with $K\ge4$, the sufficient threshold is also the order required for a finer target, a risk below a small fraction of $S_{\rm nn}(j)/K$ (Proposition~\ref{prop:barrier}, Appendix~\ref{app:open}).

\par\smallskip\noindent\fbox{\parbox{\dimexpr\columnwidth-2\fboxsep-2\fboxrule\relax}{\textbf{Open problem.} For well-spread signatures and fixed $K$, can the Riesz rate be attained near the necessary sample size $f_K(d)$? Jacobi's performance in our simulations motivates the following conjecture (Section~\ref{sec:experiments}).
\begin{conjecture}[Linear regime]\label{conj:linear}
For well-spread signatures and $K\ge4$, the output $\hat U$ of Jacobi sweeps from the identity attains the Riesz rate, $\E\,\lA^2(\hat U)\le C\,\Ebar/n$, as soon as $n\ge C'\,d\,\mathrm{polylog}(d)$.
\end{conjecture}}}\par\smallskip

\section{Projecting versus probing}
\label{sec:algo}

A common way to use several covariances is to take the eigenvectors of one matrix $\hat M_\alpha=\sum_k\alpha_k\Sh_k$, for a fixed or random $\alpha\in\R^K$. Two-environment generalised eigenproblems, AMUSE, and the random-pencil initialisations of \citet{he2024randomized,he2025randomized} are of this form.

\begin{theorem}[A linear pencil pays $d^2$ at first order]\label{thm:pencil}
Assume known whitening and independent sources with finite fourth moments. Fix $\alpha\in\R^K$, or draw it independently of the data, with the eigenvalues $\lambda_j=\langle\alpha,\sig_j\rangle$ of $M_\alpha=Q\diag(\lambda)Q^\top$ pairwise distinct. Let $\hat q_j$ be the eigenvector of $\hat M_\alpha$ for its $j$-th eigenvalue in the order of the $\lambda_j$, and let $\rho_j(\hat q_j)=1-\langle\hat q_j,q_j\rangle^2$. As $n\to\infty$ at fixed $d$,
\begin{equation*}
n\,\rho_j(\hat q_j)\Rightarrow\sum_{i\neq j}\frac{v_{ij}(\alpha)}{(\lambda_i-\lambda_j)^2}\,Z_i^2,
\end{equation*}
with $v_{ij}(\alpha)=K\sum_{k}\alpha_k^2\,\sig_i(k)\sig_j(k)$ and $Z_i$ i.i.d.\ standard normal. For every configuration and every $K$, the means of these limit laws average over $j$ to at least $(d-1)^3/(dB^4)$, of order $d^2$ uniformly in $K$ and $\alpha$. (Appendix~\ref{app:pencil} gives a sharper bound, equation~\eqref{eq:pencil-lb}, in terms of the range of the $\lambda_j$.)
\end{theorem}

The proof (Appendix~\ref{app:pencil}) combines the eigenvector expansion of \citet{anderson1963asymptotic} with a power-mean inequality on the spacings of $d$ points in an interval. A linear pencil collapses the configuration onto the line spanned by $\alpha$, where the Riesz $2$-energy is always hypersingular ($\beta=2>D=1$). The barrier is the projection, not the use of a pencil. For $K=2$ the signatures already lie on a line; for well-spread signatures the projection loses a factor of order at least $d/\log d$ ($K=3$) and $d$ ($K\ge4$). The bound concerns the mean of the first-order limit laws for a single data-independent combination. The quadratic pencil of Theorem~\ref{thm:upper} probes instead. Its eigenvalues are comparable to the divergences of all signatures from $\sig_j$, in all $K-1$ directions, and $E_j/B^2\le K\tr H_j^+\le B^2E_j$ for its pseudo-inverse $H_j^+$.

\paragraph{Joint diagonalization.} Orthogonal joint diagonalization \citep{bunse1993numerical,cardoso1996jacobi} minimises $\hat f(U)=\sum_{k=1}^K\|\off(U^\top\Sh_kU)\|_F^2$ over $U\in\Od$, $\off(X)$ the off-diagonal part of $X$. The weighted criterion of \citet{tichavsky2009fast} and the Gaussian likelihood of \citet{pham2001blind} replace the Frobenius norm by Fisher weights. For independent sources with finite fourth moments and $S_{ij}>0$, the global minimiser $\hat U$ satisfies $n\rho_j(\hat U)\Rightarrow\sum_{i\neq j}v_{ij}Z_i^2$ at fixed $d$. The coefficients are explicit and, for known and for estimated whitening alike, satisfy $S_{ij}^{-1}\le v_{ij}\le C_BS_{ij}^{-1}$ with $C_B$ depending only on $B$. With known whitening, Fisher weights give $v_{ij}=S_{ij}^{-1}$ (Proposition~\ref{prop:ajd}, Appendix~\ref{app:ajd}). Both facts are classical at fixed $d$ \citep{cardoso1994perturbation,yeredor2000blind,pham2001blind,tichavsky2009fast}. The step taken here is to read the coefficients as a Riesz energy, so that the dimension law applies to them. We write $\Ebar^{\rm w}/n$ for the resulting first-order prediction with estimated whitening, the pipeline of Section~\ref{sec:experiments}.

\section{Experiments}
\label{sec:experiments}

\paragraph{Setup.} Unless stated otherwise, we draw $A$ from the Haar law and the sample covariances from their Wishart law with $n/K$ observations per environment. We whiten with the pooled sample covariance $\hat C$ and run Jacobi sweeps \citep{cardoso1996jacobi} from the identity (Algorithm~\ref{alg:jacobi}, at most $100$ sweeps). The pencil estimators use known whitening unless stated. We score the unit-normalised rows of $\hat U^\top\hat C^{-1/2}$ against the columns of $A$ using $\lA^2$. The estimated-whitening prediction $\Ebar^{\rm w}/n$ uses the coefficients of \eqref{eq:white-var} and tracks Jacobi's error in these simulations when $\Ebar^{\rm w}/n\lesssim0.1$. Random signatures are i.i.d.\ uniform on $\{\sig\in[\frac12,\frac32]^K:\sum_k\sig(k)=K\}$ ($B=2$). Appendix~\ref{app:exp} and the supplement give the settings, code and raw results.

\paragraph{Signature dimension governs the error.} At $n=400d$, the error grows with $d$ for $K=2$ and $3$ but stays near $0.02$ for $K=8$ (Figure~\ref{fig:story}(b)). Holding $K=8$ fixed and varying the dimension of the signatures gives the same contrast. From $d=8$ to $256$, the loss grows from $\LawDimOneFirst$ to $\LawDimOneLast$ for signatures on a line, and stays nearly flat in dimensions $3$ and $7$ (Figure~\ref{fig:mech}(a)). At $n=400d$, the errors for $K=4$ and $8$ are close to their first-order predictions (Figure~\ref{fig:mech}(d)).

\paragraph{Equal gaps can give different errors.} In two configurations with the same $S_{\min}$, the energies differ by a factor of $\LawGapEnergyRatio$ and the errors by a factor of $\LawGapErrRatio$. The errors are within $\LawGapDevA$ and $\LawGapDevB$ of their respective first-order predictions (Figure~\ref{fig:mech}(b)).

\paragraph{Jacobi's sample requirement grows slowly with dimension.} On fixed $K=4$ lattices, from $d=27$ to $\LargeDMax$, the $n/d$ needed for a mean loss of $0.1$ grows by a factor of $\MeanRiskCrossGrowth$, close to the factor of $\MeanRiskEwGrowth$ predicted by $\Ebar^{\rm w}/d$. Over the same range, the sufficient threshold proved for the quadratic pencil gives a factor of $\MeanRiskProvedGrowth$ in $n/d$ (Figure~\ref{fig:larged}). Random signatures at $K=4$ and $8$ also show slow growth in the $n/d$ needed for a median loss of $0.1$ (Figure~\ref{fig:mech}(c)). These results motivate Conjecture~\ref{conj:linear}.

\paragraph{The quadratic pencil and grid.} At $n=800d$, with $K=16$ code signatures and known whitening, the loss is $\CodeRatioMin$--$\CodeRatioMax$ times its first-order prediction for $d=\CodedMin$ to $\CodedMax$ (Figure~\ref{fig:review}(b)). Tests with sample-split estimated whitening also track the first-order prediction (Appendix~\ref{app:exp}). On $K=4$ lattices with known whitening and the true minimum Euclidean spacing supplied, a refined grid recovers one probe per source in every tested run with $n\ge\UnkRatioMin d$ and $d\le\UnkdMax$. Its loss is within $\UnkRatioOracleMax$ of the loss using the true signatures (Table~\ref{tab:unknown}).

\paragraph{One linear pencil degrades with $d$.} On the same data ($K=8$, $n=400d$), the loss of a single random linear pencil grows from $\BasePencilEightFirst$ to $\BasePencilEightLast$, while Pham's algorithm \citep{pham2001joint} stays within a factor of $\BasePhamEightMin$--$\BasePhamEightMax$ of its first-order Cram\'er--Rao prediction (Figures~\ref{fig:story}(d) and~\ref{fig:review}(a)). Using the linear pencil to initialise Jacobi gives numerically the same converged estimates as starting from the identity in the runs tested (Appendix~\ref{app:exp}).

\begin{figure}[t]
\centering
\includegraphics[width=0.84\columnwidth]{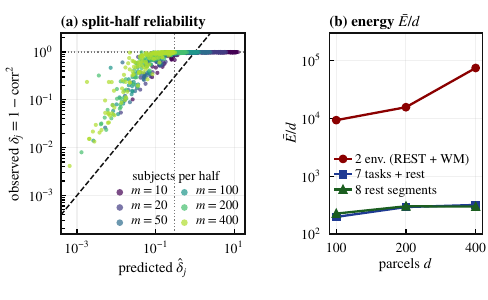}
\caption{\textbf{Real data (HCP task fMRI).} \textbf{(a)}~Observed split-half discrepancy $\delta_j=1-\mathrm{corr}^2$ versus its first-order prediction for $d=\FmriD$ components (colour indicates subjects per half, dashed line shows $y=x$, dotted lines show $\hat\delta=\FmriPredLin$ and $\delta=1$). \textbf{(b)}~Estimated energy per source $\Ebar/d$ across resolutions for two environments (rest and working memory), seven tasks plus rest, and eight rest segments.}
\label{fig:fmri}
\end{figure}

\paragraph{Real data: split-half stability in fMRI.} We fit $d=\FmriD$ components to fMRI from $\FmriN$ Human Connectome Project subjects \citep{vanessen2013wuminn,barch2013function}, using Schaefer parcels \citep{schaefer2018local} and seven tasks plus rest as environments. We compare fits from disjoint groups of subjects. At $\FmriMsub$ subjects per half, across all $\FmriD$ components, the energy $E_j$ and the inverse nearest-neighbour information $1/S_{\rm nn}(j)$ have Spearman correlations of $\FmriGapSpearmanEnergyAllMsub$ and $\FmriGapSpearmanNNAllMsub$ with the observed split-half discrepancies. Only \FmriFracLinMFifty{} of components are below the first-order cutoff $\hat\delta_j<\FmriPredLin$ at that sample size. With $\FmriMmax$ subjects per half, the mean discrepancy is $\FmriDeltaBarMFourHundred$, compared with an autocorrelation-adjusted first-order prediction of $\FmriPredBarMFourHundred$ (Figure~\ref{fig:fmri}(a)). The discrepancies level off across sample sizes, and the fitted plateau accounts for most of their ranking. The split fits did not reach the Jacobi stopping tolerance (Appendix~\ref{app:exp}).

Eight rest segments yield energies similar to the seven tasks plus rest, despite more concentrated variation across their signatures (Figure~\ref{fig:fmri}(b)). The two-environment fit yields much larger energy estimates, with considerable uncertainty because its smallest estimated signature distances are below their estimation noise (Appendix~\ref{app:exp}).

\section{Discussion}
\label{sec:discussion}
\paragraph{Diversity versus non-Gaussianity.} In the non-Gaussian route the information cost is of order $d$, while low-degree estimators need at least order $d^2$ samples \citep{auddy2025large}. With variance diversity the information cost is the Riesz energy. Section~\ref{sec:law} gives the conditions for polynomial-time attainment with the quadratic pencil. The code-signature design is related to the use of logarithmically many interventions for identification in linear causal representation learning \citep{lee2026beyond}.

\par\smallskip\noindent\fbox{\parbox{\dimexpr\columnwidth-2\fboxsep-2\fboxrule\relax}{\textbf{Practical takeaways.} (1)~A Jacobi pilot gives signatures for a plug-in sample-size diagnostic $\Ebar^{\rm w}/n$. In our simulations the prediction using the true signatures tracks the error below about $0.1$. (2)~Choose environments that spread the signatures in several directions at the scales of nearby signatures. (3)~A single linear pencil can initialise Jacobi, which retained accuracy in the tested settings.}}\par\smallskip

\paragraph{Limitations.} Theorems~\ref{thm:upper}--\ref{thm:unknown} require Gaussian sources and equal environment sizes. The grid uses a supplied separation bound $s_0$, and its size bound is polynomial in $d$ when $K\log(C/s_0)=O(\log d)$. The fixed-$K$ sample-size gap remains open. Proposition~\ref{prop:ajd} gives fixed-dimensional first-order behaviour for the global joint diagonalizer. Our model assumes independent samples and independent sources. The fMRI analysis uses a shared mixing and a Bartlett autocorrelation correction, and relatives may appear in opposite halves of a split.\label{end:body}

\section*{Impact Statement}
We foresee no negative societal consequences specific to this methodological work on source separation; the fMRI data are the open-access HCP data, used under their data use terms, and no attempt is made to identify subjects.

\ifdefined\isaccepted\def\showack{1}\fi\ifdefined\ispreprint\def\showack{1}\fi
\ifdefined\showack
\section*{Acknowledgements}
Data were provided by the Human Connectome Project, WU-Minn Consortium (Principal Investigators: David Van Essen and Kamil Ugurbil; 1U54MH091657) funded by the 16 NIH Institutes and Centers that support the NIH Blueprint for Neuroscience Research, and by the McDonnell Center for Systems Neuroscience at Washington University.
\fi

\bibliographystyle{icml2026}
\bibliography{references}

\clearpage
\appendix
\onecolumn

\section*{Notation}
\begin{center}\small\setlength{\tabcolsep}{5pt}
\begin{tabular}{ll}\toprule
$\sig_j\in\R^K$, $\sig_j(k)\in[1/B,B]$ & signature of source $j$ (variances across environments), $\frac1K\sum_k\sig_j(k)=1$\\
$S_{ij}$ \eqref{eq:S}; $S_{\min}$, $S_{\rm nn}(j)$ & Fisher information of the $(i,j)$ rotation; its minimum over pairs, over $i$ for fixed $j$\\
$E_j=\sum_{i\neq j}S_{ij}^{-1}$; $\Ebar$, $\Emax$ & Riesz $2$-energy of $\sig_j$; its average and maximum over sources\\
$\delta_a=\sig_a-\sig_j$; $r_j=\min_{i\neq j}\|\delta_i\|$ & signature differences from source $j$; distance to the nearest signature\\
$\delta_j$, $\hat\delta_j$ (Section~\ref{sec:experiments}, Appendix~\ref{app:exp} only) & split-half discrepancy of fMRI component $j$ and its prediction\\
$H_j$, $\lambda_a$, $\lambda_*$ & quadratic pencil at $\sig_j$, its eigenvalues, its gap (Section~\ref{sec:upper}, Appendix~\ref{app:upper})\\
$v^{\rm qp}_{ij}$; $E^{\rm qp}_j$, $\Ebar^{\rm qp}$; $\hat E^{\rm qp}_j$ & first-order coefficients of the estimator of Theorem~\ref{thm:upper}; with a perturbed probe (Lemma~\ref{lem:probe})\\
$v^{\rm wh}_{ij}$; $E^{\rm wh}_j$ & first-order coefficient of the estimated whitening in Theorem~\ref{thm:upper-full}(b)\\
$\gamma_j$, $\gamma$ & $K(d+6\log n)/(nS_{\rm nn}(j))$ and $K(d+6\log n)/(nS_{\min})$\\
$\tilde\gamma$ (Appendix~\ref{app:exp} only) & $Kd/(nS_{\min})$, $\gamma$ without the term $6\log n$ (Table~\ref{tab:step}, Figure~\ref{fig:extra}(b))\\
$s_0$ & known lower bound on $S_{\min}$ (Theorem~\ref{thm:unknown})\\
$v^{\rm F}_{ij}$, $v^{\rm w}_{ij}$; $\Ebar^{\rm F}$, $\Ebar^{\rm w}$ & first-order coefficients of joint diagonalization, known/estimated whitening \eqref{eq:frob-var}--\eqref{eq:white-var}\\
$v^{\rm PC}_{ij}$; $\Ebar^{\rm PC}$ & Cram\'er--Rao coefficient of the non-orthogonal problem (Appendix~\ref{app:ajd})\\
$\rho_j$, $\lA^2$, $\lM^2$ & squared sine of the angle for source $j$; average and maximal loss\\
$f_K(d)$ & $d^2$, $d\log d$, $d$ for $K=2$, $3$, $\ge4$\\\bottomrule
\end{tabular}
\end{center}

\section*{Reader's guide}
Each result of the main text with its full statement, its conditions and its proof. ``Gaussian'': Gaussian sources; ``4th mom.'': independent sources with finite fourth moments; ``equal'' and ``any'': the environment sizes $n_k$ (``any'' with the weighted divergence of Appendix~\ref{app:FI}); $\gamma_j=K(d+6\log n)/(nS_{\rm nn}(j))$, $\gamma=\max_j\gamma_j$ and $\gamma_0=K(d+6\log n)/(ns_0)$.
{\footnotesize\setlength{\tabcolsep}{2.5pt}\renewcommand{\arraystretch}{1.08}
\begin{longtable}{@{}p{1.15in}p{1.0in}p{1.15in}p{1.15in}p{0.72in}p{1.05in}@{}}
\toprule
Result (main text) & Full statement; proof & Sources; whitening; sizes & Signatures & Regime & Not covered\\\midrule
\endfirsthead
\toprule
Result (main text) & Full statement; proof & Sources; whitening; sizes & Signatures & Regime & Not covered\\\midrule
\endhead
\bottomrule
\endlastfoot
Theorem~\ref{thm:rate}, lower bound & Theorem~\ref{thm:rate-full}(a); App.~\ref{app:rate} & Gaussian; known; any & may be used; estimators in $\Od$ or labelled; $d$ directions under the closest-direction loss, average bound only (Lemma~\ref{lem:matched}) & every $n$, $d$ & the per-source and maximal losses $\rho_j$, $\lM$ for non-orthogonal estimators\\
Theorem~\ref{thm:rate}, attainment at fixed $d$ & Theorem~\ref{thm:rate-full}(b); App.~\ref{app:rate} & Gaussian; known; any & $S_{ij}>0$ & $n\to\infty$, fixed $d$ & the $d$-dependence\\
Theorem~\ref{thm:upper}(a), pencil, known whitening & Theorem~\ref{thm:upper-full}(a); App.~\ref{app:upper} & Gaussian; known; equal & known & $\gamma_j\le c$ & unequal sizes (App.~\ref{app:FI})\\
Theorem~\ref{thm:upper}(b), estimated whitening & Theorem~\ref{thm:upper-full}(b); App.~\ref{app:upper}, Lemma~\ref{lem:remainder} & Gaussian; estimated on a third; equal & known & $\gamma_j\le c$ & unequal sizes\\
Lemmas~\ref{lem:probe}, \ref{lem:probe-wh}, perturbed probes & App.~\ref{app:upper}, \ref{app:unknown} & Gaussian; known / estimated; equal & a probe within $\eta/B^4$ of the nearest-neighbour distance ($\eta\le1/8$), independent of the halves & $\gamma_j\le c$ & \\
Theorem~\ref{thm:unknown}(a), grid, known whitening & Theorem~\ref{thm:unknown-full}(a); App.~\ref{app:unknown} & Gaussian; known; equal & only $B$ and $s_0\le S_{\min}$ & $\gamma_0\le c$ & adaptivity in $s_0$; the grid is exponential in $K$\\
Theorem~\ref{thm:unknown}(b), estimated whitening & Theorem~\ref{thm:unknown-full}(b); App.~\ref{app:unknown}, Lemma~\ref{lem:probe-wh} & Gaussian; estimated on a fourth; equal & only $B$ and $s_0\le S_{\min}$ & $\gamma_0\le c$ & \\
Corollary~\ref{cor:minimax}, minimax rate & App.~\ref{app:upper}; (iii): App.~\ref{app:unknown}, Lemma~\ref{lem:matched} & Gaussian; known or estimated; equal & known; $s_0=S_{\min}$ for Theorem~\ref{thm:unknown} & $\gamma\le c$ & \\
Theorem~\ref{thm:klaw}, dimension law & App.~\ref{app:klaw} & geometry only & well-spread (Definition~\ref{def:qu}); lower bounds for $\Ebar$: every configuration & fixed $K$ & constants depend on $K$\\
Corollary~\ref{cor:necessary}, necessary size & App.~\ref{app:klaw} & Gaussian; known; any & every configuration & every $n$, $d$ & \\
Proposition~\ref{prop:random}, random signatures & App.~\ref{app:klaw} & geometry only & i.i.d., density bounded above and below on a convex body & probability $\ge1-\epsilon$, constants depend on $\epsilon$ & \\
Theorem~\ref{thm:pencil}, linear pencil & App.~\ref{app:pencil} & 4th mom.; known; any & not used; $\alpha$ independent of the data & $n\to\infty$, fixed $d$ & adaptive or iterated $\alpha$; the pencil as an initialisation (App.~\ref{app:exp})\\
Proposition~\ref{prop:ajd}, joint diagonalization & App.~\ref{app:ajd} & 4th mom.; known or estimated; any & not used; global minimiser & $n\to\infty$, fixed $d$ & Jacobi's output (Conjecture~\ref{conj:linear})\\
Proposition~\ref{prop:barrier}, barrier & App.~\ref{app:open} & Gaussian; known; any & -- & every $n$, $d$ & \\
Conjecture~\ref{conj:linear}, linear regime & Section~\ref{sec:experiments}, App.~\ref{app:exp} & simulation; estimated; equal & $K=4$ lattices and random-signature tests & tested settings, $d\le\LargeDMax$ & \\\end{longtable}\addtocounter{table}{-1}}   

\section{Fisher information of a rotation (proof of Lemma~\ref{lem:FI}), unequal sizes and non-Gaussian sources}
\label{app:FI}

\begin{lemma}[Fisher information of a rotation]\label{lem:FI}
For Gaussian sources and $R\in\Od$, with $Q'=QR$ and $P_Q$ the law of the full sample,
\begin{equation}\label{eq:KL}
\KL\big(P_{Q}\,\|\,P_{Q'}\big)=\frac{n}{2K}\sum_{k=1}^K\sum_{a\neq b}R_{ab}^2\Big(\frac{\sig_a(k)}{\sig_b(k)}-1\Big).
\end{equation}
If $R$ is a rotation by $\theta$ in the coordinate plane $(i,j)$, the divergence equals $\frac n2\sin^2\!\theta\,S_{ij}$ exactly. Hence $S_{ij}$ is the Fisher information per sample, averaged over environments, for the rotation angle in plane $(i,j)$. The information is diagonal across planes and orthogonal to the variance parameters.
\end{lemma}

For centred Gaussians, $\KL(N(0,\Sigma)\|N(0,\Sigma'))=\frac12[\tr(\Sigma'^{-1}\Sigma)-d-\log\det(\Sigma'^{-1}\Sigma)]$. With $\Sigma_k=Q\Lambda_kQ^\top$ and $\Sigma'_k=QR\Lambda_kR^\top Q^\top$ the determinants agree and $\tr(\Sigma_k'^{-1}\Sigma_k)=\tr(R\Lambda_k^{-1}R^\top \Lambda_k)=\sum_{a,b}R_{ab}^2\,\sig_a(k)/\sig_b(k)$. Since $\sum_bR_{ab}^2=1$, subtracting $d=\sum_{a,b}R_{ab}^2$ gives $\sum_{a\neq b}R_{ab}^2(\sig_a(k)/\sig_b(k)-1)$, and summing the $n_k=n/K$ samples of each environment gives \eqref{eq:KL}. With $n_k=\pi_kn$ the same computation gives \eqref{eq:KL} with $\frac1K\sum_k$ replaced by $\sum_k\pi_k$, hence $S_{ij}=\sum_k\pi_k(\sig_i(k)/\sig_j(k)+\sig_j(k)/\sig_i(k)-2)$. For a planar rotation in $(i,j)$, $R_{ij}^2=R_{ji}^2=\sin^2\theta$ and all other off-diagonal entries vanish, so the sum equals $\sin^2\theta\,(\sig_i(k)/\sig_j(k)+\sig_j(k)/\sig_i(k)-2)$, whose average over $k$ is $S_{ij}$; the Fisher information is the coefficient of $\theta^2/2$ in $\KL$ per sample, i.e.\ $S_{ij}$. For the orthogonality statements, the score for the rotation angle in plane $(i,j)$ at $\theta=0$ is $\sum_t s_{i,t}s_{j,t}(\sig_j(k_t)^{-1}-\sig_i(k_t)^{-1})$, the score for the variance $\sig_l(k)$ is a function of $s_{l,t}^2$ alone, and the cross-expectations $\E[s_is_js_{i'}s_{j'}]$, $\E[s_is_j(s_l^2-\sig_l)]$ vanish by independence unless $\{i,j\}=\{i',j'\}$. $\square$

\paragraph{Unequal sizes.} With $n_k=\pi_kn$, the proofs of Theorems~\ref{thm:rate} and~\ref{thm:klaw}, Corollary~\ref{cor:necessary} and Propositions~\ref{prop:barrier} and~\ref{prop:random} use only Lemma~\ref{lem:FI} and \eqref{eq:S-euclid} with $\|\cdot\|^2$ replaced by $\sum_k\pi_k(\cdot)_k^2$ and $K$ by $1$, so they hold with the weighted divergence. The comparison of the weighted norm with $\|\cdot\|^2/K$ makes the constants of the upper bounds in Theorem~\ref{thm:klaw} and Proposition~\ref{prop:random} depend also on $\min_kK\pi_k$ (an environment with $\pi_k\to0$ collapses one dimension); their lower bounds and Corollary~\ref{cor:necessary} hold uniformly in the sizes, since they need only $S_{ij}\le B^2\sum_k\pi_k(\sig_i(k)-\sig_j(k))^2\le B^2\|\sig_i-\sig_j\|^2$. The first-order results hold as well: Theorem~\ref{thm:rate-full}(b) through the weighted Fisher information, and Theorem~\ref{thm:pencil} and Proposition~\ref{prop:ajd}, whose proofs use only $\Var(N_{k,ab})=\sig_a(k)\sig_b(k)/n_k$, for the $\pi$-weighted pencils and criteria; Appendix~\ref{app:ajd} gives the coefficients used in Figure~\ref{fig:extra}(c). The final bound $(d-1)^3/(dB^4)$ of Theorem~\ref{thm:pencil} holds for every $\pi$, since then $v_{ij}(\alpha)=\sum_k\alpha_k^2\sig_i(k)\sig_j(k)/\pi_k\ge B^{-2}\|\alpha\|_1^2$ and $L_\alpha\le B\|\alpha\|_1$.

The non-asymptotic upper bounds (Theorems~\ref{thm:upper} and~\ref{thm:unknown}, Lemma~\ref{lem:probe}, Corollary~\ref{cor:minimax}) are proved for equal sizes only, and the $\pi$-weighted estimators fail for arbitrary sizes. Take $K_0\asymp\log d$ environments with $m$ samples each and separated code signatures (Section~\ref{sec:law}), and add one environment $k_0$ with $\sig_j(k_0)=1$ for every $j$ and two samples, so that $n=K_0m+2$ and $\pi_{k_0}=2/n$. In the split pencil of Theorem~\ref{thm:upper} with weights $\pi_kw_k$, whose halves of $k_0$ have one sample each, this environment contributes $\frac2n\big[\frac12x^\top y\,(xy^\top+yx^\top)-xx^\top-yy^\top+I\big]$ with $x,y$ independent $N(0,I_d)$, whose smallest eigenvalue is about $-|x^\top y|\,\|x\|\|y\|/n\asymp-d^{3/2}/n$ with probability bounded below, while the population pencil has its gap of order one and receives nothing from $k_0$. With $n=CK_0d$ the condition $\gamma_j\le c$ of Theorem~\ref{thm:upper}, with the weighted divergence, holds for large $C$, but this eigenvalue has size of order $\sqrt d/(C\log d)\to\infty$; so, with probability bounded below, the bottom eigenvector is within an angle $O(\log d/\sqrt d)$ of the bottom eigenvector of this term (Davis--Kahan), a direction of $\mathrm{span}\{x,y\}$ whose law is uniform on the sphere. On this event $\langle\hat q_j,q_j\rangle^2\to0$ and $\rho_j\to1$ in probability, whereas $E_j/n\asymp1/\log d$. With six samples in $k_0$ the same happens to the estimator of Theorem~\ref{thm:unknown}. The proof of Theorem~\ref{thm:upper} needs the concentration bound of Step~2 in every environment, which fails here; how far a lower bound on $\min_kn_k$ suffices we leave open. Whether the minimax rate of Corollary~\ref{cor:minimax} survives arbitrary sizes is also open; in this example it does, since discarding $k_0$ leaves an equal-size problem whose divergences differ from the weighted ones by the factor $n/(n-2)$. In practice, environments with $n_k\lesssim d+\log n$ can be dropped or merged before the pencil is formed. We expect Theorem~\ref{thm:upper} to hold whenever $\min_kK\pi_k\ge c>0$, with constants depending on $c$, and for independent sub-Gaussian sources: Step~2 of Appendix~\ref{app:upper} has a sub-Gaussian version \citep[Thm.~5.39]{vershynin2012introduction} and Step~5 and Lemma~\ref{lem:remainder} use only independence and fourth moments. We have not checked the constants and claim neither.

\paragraph{Non-Gaussian sources.} This paragraph concerns the parametric model in which the source densities are known; for the semiparametric problem with unknown densities, in the i.i.d.\ model, see \citet{amari1997blind,chen2006efficient}. Let source $j$ in environment $k$ have density $\sig_j(k)^{-1/2}g_j(s/\sqrt{\sig_j(k)})$, where $g_j$ has mean $0$ and variance $1$ and is absolutely continuous with score $\psi_j=-(\log g_j)'$, finite location information $\kappa_j=\E\psi_j(u)^2$ and finite scale information $\E[u^2\psi_j(u)^2]$, and $s\,g_j(s)\to0$ as $|s|\to\infty$. Integration by parts gives $\E[\psi_j(u)u]=1$, so $\kappa_j\ge1$ by Cauchy--Schwarz, with equality iff $g_j$ is Gaussian. The score for the rotation angle in plane $(i,j)$ at $\theta=0$ is $\psi_i(s_i)s_j-\psi_j(s_j)s_i$ (scores at the actual scales); since $\E\psi_i(s_i)^2=\kappa_i/\sig_i(k)$, its variance in environment $k$ is $\kappa_i\sig_j(k)/\sig_i(k)+\kappa_j\sig_i(k)/\sig_j(k)-2$, and it is uncorrelated with the scores of the variances and of the other planes, as in Lemma~\ref{lem:FI}. The per-sample information is therefore
\begin{equation*}
S^\kappa_{ij}=\frac1K\sum_k\Big[\kappa_i\frac{\sig_j(k)}{\sig_i(k)}+\kappa_j\frac{\sig_i(k)}{\sig_j(k)}\Big]-2\ \ge\ S_{ij},
\end{equation*}
strictly unless both sources are Gaussian, and $S^\kappa_{ij}=\kappa S_{ij}+2(\kappa-1)$ when $\kappa_i=\kappa_j=\kappa$: within this parametric model, Gaussian sources are the least informative. The term $2(\kappa-1)$ caps the contribution of each pair to the energy $E^\kappa_j=\sum_{i\neq j}(S^\kappa_{ij})^{-1}$ at $1/(2(\kappa-1))$, so $E^\kappa_j\le(d-1)/(2(\kappa-1))$ whatever the geometry, while for weakly non-Gaussian sources ($\kappa-1\ll S_{\rm nn}(j)$) the geometry of Theorem~\ref{thm:klaw} takes over. This is a statement about Fisher information; minimax lower bounds for non-Gaussian classes, and the low-degree barrier of the non-Gaussian route \citep{auddy2025large}, are separate questions.

\paragraph{Optimality among second-order estimators.} Theorem~\ref{thm:rate-full}(b) is the efficiency of maximum likelihood for Gaussian sources. For other source laws the Riesz rate remains the first-order limit of every estimator that uses the data through the sample covariances in a regular way. Let the sources be independent with finite fourth moments and let $\hat Q=T(\Sh_1,\dots,\Sh_K)$, where $T$ is differentiable at the population covariances and Fisher consistent, $T(\Sigma_1(\theta'),\dots,\Sigma_K(\theta'))=Q(\theta')$ for every parameter $\theta'=(Q',\{\Lambda'_k\})$ in a neighbourhood; every joint diagonalizer of Proposition~\ref{prop:ajd} is of this form. In the basis of $Q$ the entries $N_{k,ab}=(\Sh_k-\Sigma_k)_{ab}$, $a\le b$, are uncorrelated across pairs and environments, with variance $\sig_a(k)\sig_b(k)/n_k$ for $a<b$, since $\E[s_as_bs_{a'}s_{b'}]=0$ unless $\{a,b\}=\{a',b'\}$ and $\E[s_as_b(s_l^2-\sig_l)]=0$ for $a\neq b$; only the variances of the diagonal entries involve the fourth moments. A rotation by $\Xi$ moves the mean of $N_{k,ab}$, $a<b$, by $\Xi_{ab}(\sig_a(k)-\sig_b(k))+o(\|\Xi\|)$ and the means of the diagonal entries only at second order, while a change of $\sig_l(k)$ moves only the mean of the diagonal entry $(k,ll)$. By the delta method and the chain rule applied to the identity $T(\Sigma(\theta'))=Q(\theta')$, the rotation coordinates of $\hat Q$ have asymptotic covariance $LVL^\top$ for a matrix $L$ with $L\dot\Sigma=[\,I\ \ 0\,]$, where $V$ is the asymptotic covariance of the sample covariances and $\dot\Sigma$ the derivative of their means in $\theta$. By the two facts just stated, $\dot\Sigma$ is block diagonal: the rotation coordinates enter only the off-diagonal means at first order, the variance coordinates only the diagonal ones. So the restriction $L_{\rm off}$ of $L$ to the off-diagonal entries satisfies $L_{\rm off}\dot\Sigma_{\rm off}=I$, and since diagonal and off-diagonal entries are uncorrelated, $LVL^\top\succeq L_{\rm off}V_{\rm off}L_{\rm off}^\top\succeq(\dot\Sigma_{\rm off}^\top V_{\rm off}^{-1}\dot\Sigma_{\rm off})^{-1}$ by the Gauss--Markov inequality, where $V_{\rm off}$ is diagonal with the entries $\sig_a(k)\sig_b(k)/n_k$ above and $\dot\Sigma_{\rm off}^\top V_{\rm off}^{-1}\dot\Sigma_{\rm off}$ is diagonal with entries $\sum_kn_k(\sig_a(k)-\sig_b(k))^2/(\sig_a(k)\sig_b(k))$, which is $nS_{ab}$ for $n_k=n/K$. Hence every rotation angle has asymptotic variance at least $1/(nS_{ab})$, whatever the fourth moments (only the off-diagonal block of $V$ enters, and it is nonsingular), and the limit of $n\rho_j$ has mean at least $E_j$, which Proposition~\ref{prop:ajd}(ii) attains. Regularity cannot be dropped. For symmetric two-valued sources, $s_a=\pm\sqrt{\sig_a(k)}$, the diagonal of $Q^\top\Sh_kQ$ equals $\Lambda_k$ exactly, $V$ is singular on the diagonal entries, and a non-regular estimator can exploit them: for $d=2$, $S_{12}>0$ and known signatures (where the bound above holds for regular $T$ as well), the common solutions $\theta'$ of $\diag(R_{\theta'}^\top\Sh_kR_{\theta'})=\Lambda_k$, $k=1,2$, with $R_{\theta'}$ the rotation by $\theta'$, are the true angle and, only on an event of probability $O(n^{-1/2})$ (the off-diagonal sample moments $\nu_k=n_k^{-1}\sum_ts_{1t}s_{2t}$ are lattice variables), one further angle at distance $O(|\nu_k|)$ from it; an estimator that returns a common solution therefore has $\rho_j=0$ off this event and $\rho_j\le C\max_k\nu_k^2$ on it. Since $\E\nu_k^4=O(n^{-2})$, Cauchy--Schwarz gives $\E[n\rho_j]=O(n^{-1/4})\to0$, below the regular bound $1/S_{12}$. Estimators that use higher-order statistics are not covered; for them the information $S^\kappa_{ij}$ above applies.

\paragraph{The loss.} $\lA^2$ and $\lM^2$ coincide with the squared average and maximal losses of \citet{auddy2025large} whenever the matching is a permutation, and are never larger, so the lower bounds of Theorem~\ref{thm:rate} transfer to their losses. Like the Amari index, $\lA^2$ is permutation invariant, but it decomposes by source.

\section{Proof of Theorem~\ref{thm:rate}}
\label{app:rate}

\begin{theoremfull}[Theorem~\ref{thm:rate}, full statement]\label{thm:rate-full}
For Gaussian sources and $K\ge2$:
\begin{enumerate}[leftmargin=*,itemsep=1pt,topsep=2pt]
\item[(a)] (Lower bound.) There is a constant $c>0$ depending only on $B$ such that, for every $n$ and $d$, every source $j$ and every estimator $\hat Q\in\Od$ (which may use the signatures),
\begin{align}
\sup_{Q\in\Od}\E\,\rho_j(\hat Q)\;&\ge\;c\,\frac{E_j}{n+E_j}\quad\text{(this is \eqref{eq:lb2})},\notag\\
\sup_{Q\in\Od}\E\,\lA^2(\hat Q)\;&\ge\;\frac cd\sum_{j=1}^d\frac{E_j}{n+E_j},\label{eq:lb3}\\
\sup_{Q\in\Od}\E\,\lM^2(\hat Q)\;&\ge\;c\,\max_j\frac{E_j}{n+E_j}.\label{eq:lb4}
\end{align}
If $S_{ij}=0$ for some $i\neq j$, then $E_j=\infty$ and $E_j/(n+E_j)$ is read as $1$.
\item[(b)] (Attainment at fixed $d$.) Let $S_{ij}>0$ for all $i\neq j$. For fixed $d$ and $n\to\infty$, the Gaussian maximum-likelihood joint diagonalizer $\hat Q_{\rm ML}$ satisfies $n\,\rho_j(\hat Q_{\rm ML})\Rightarrow\sum_{i\neq j}S_{ij}^{-1}Z_i^2$ with $Z_i$ i.i.d.\ standard normal, a limit law with mean $E_j$.
\end{enumerate}
\end{theoremfull}
Part (b) is the classical efficiency of maximum likelihood \citep{flury1986asymptotic,pham2001blind}, proved at the end of this appendix with the information of Lemma~\ref{lem:FI}. All three bounds of (a) hold for labelled estimators (Appendix~\ref{app:upper}, proof of Corollary~\ref{cor:minimax}).

Fix $j$ and a reference $Q_0\in\Od$. Write $C_{\rm tri}=2B^4$. By \eqref{eq:S-euclid}, $S_{ab}\le\frac{B^2}K\|\sig_a-\sig_b\|^2\le\frac{2B^2}K\big(\|\sig_a-\sig_j\|^2+\|\sig_j-\sig_b\|^2\big)\le C_{\rm tri}(S_{ja}+S_{jb})$ for all $a,b$, with $S_{jj}=0$.

\paragraph{The hypercube.} Order the other sources as $i_1,\dots,i_m$, $m=d-1$, so that $t_l=S_{ji_l}$ is non-increasing: $t_1\ge t_2\ge\dots\ge t_m>0$. Let $E=E_j=\sum_lt_l^{-1}$, $\eta=1/(64B^4)$ and
\begin{equation*}
\theta_l^2=\frac{\eta}{(n+E)\,t_l},\qquad\Theta=\sum_l\theta_l^2=\frac{\eta E}{n+E}\le\eta\le\frac1{64}.
\end{equation*}
For $\varepsilon\in\{\pm1\}^m$ put
\begin{equation*}
Q_\varepsilon=Q_0\,R_1(\varepsilon_1\theta_1)R_2(\varepsilon_2\theta_2)\cdots R_m(\varepsilon_m\theta_m),
\end{equation*}
where $R_l(\theta)$ is the rotation by $\theta$ in the coordinate plane $(j,i_l)$, with $R_l(\theta)e_j=\cos\theta\,e_j+\sin\theta\,e_{i_l}$. Hypercubes of planar rotations are standard in minimax bounds for principal subspaces \citep{vu2013minimax,cai2013sparse}; the ordering by $S_{ji}$, which keeps the divergence between neighbours bounded without truncation, is the new ingredient.

\emph{Geometry of column $j$.} Since $R_l$ fixes every $e_{i_{l'}}$ with $l'\neq l$, applying the rotations from right to left gives
\begin{equation*}
q_j(\varepsilon)=Q_0\Big[\Big(\prod_l\cos\theta_l\Big)e_j+\sum_l\varepsilon_l\sin\theta_l\Big(\prod_{l'>l}\cos\theta_{l'}\Big)e_{i_l}\Big].
\end{equation*}
The coordinate of $q_j(\varepsilon)$ along $Q_0e_{i_l}$ is $\varepsilon_l\tau_l$ with $\tau_l=\sin\theta_l\prod_{l'>l}\cos\theta_{l'}\ge\theta_l/2$, because $\Theta\le1/64$ gives $\sin\theta_l\ge0.99\,\theta_l$ and $\prod_{l'}\cos\theta_{l'}\ge1-\Theta/2$. Also $\sin^2\!\angle(q_j(\varepsilon),Q_0e_j)=1-\prod_l\cos^2\theta_l\le\Theta$.

\emph{Reduction to a Hamming loss.} The reduction must use a function of the data alone. Let $\hat q$ be the column of $\hat Q\in\Od$ with the largest $|\langle\hat q_a,Q_0e_j\rangle|$, signed so that $\langle\hat q,Q_0e_j\rangle\ge0$, and put $\hat\varepsilon_l=\operatorname{sign}\langle\hat q,Q_0e_{i_l}\rangle$ (with $\operatorname{sign}0=1$). We claim that for every $\varepsilon$ and every $\hat Q\in\Od$,
\begin{equation}\label{eq:hamming}
\sum_l\tau_l^2\,\mathbf 1[\hat\varepsilon_l\neq\varepsilon_l]\le2\rho_j(\hat Q).
\end{equation}
If $\rho_j(\hat Q)\ge1/8$ the left side is at most $\sum_l\tau_l^2\le\Theta\le1/64\le2\rho_j$. If $\rho_j(\hat Q)<1/8$, let $q^*$ be the column closest to $q_j(\varepsilon)$, at angle $\alpha<\arcsin(1/\sqrt8)<0.362$. Since $\angle(q_j(\varepsilon),Q_0e_j)\le\arcsin\sqrt\Theta<0.13$, $q^*$ is within angle $0.5<\pi/4$ of $Q_0e_j$, while every other column, being orthogonal to $q^*$, is at angle more than $\pi/4$ from $Q_0e_j$. Hence $\hat q=\pm q^*$, and with the sign convention $\langle\hat q,q_j(\varepsilon)\rangle=\cos\alpha>0$. If $\hat\varepsilon_l\neq\varepsilon_l$ then $|\langle\hat q-q_j(\varepsilon),Q_0e_{i_l}\rangle|\ge\tau_l$, so
\begin{equation*}
\sum_l\tau_l^2\,\mathbf 1[\hat\varepsilon_l\neq\varepsilon_l]\le\|\hat q-q_j(\varepsilon)\|^2=2(1-\cos\alpha)\le2\sin^2\alpha=2\rho_j(\hat Q).
\end{equation*}

\emph{Neighbouring hypotheses.} Let $\varepsilon,\varepsilon'$ differ only in coordinate $l$. Then
\begin{equation*}
R:=Q_\varepsilon^\top Q_{\varepsilon'}=T^\top R_l(2\varepsilon'_l\theta_l)\,T,\qquad T=R_{l+1}(\varepsilon_{l+1}\theta_{l+1})\cdots R_m(\varepsilon_m\theta_m),
\end{equation*}
an exact rotation by $\delta=2\varepsilon'_l\theta_l$ in the plane spanned by the orthonormal vectors $u=T^\top e_j$ and $w=e_{i_l}$. The vector $u$ is supported on $J_l=\{j,i_{l+1},\dots,i_m\}$, and $u_{i_l}=0$. Write $R=I+(\cos\delta-1)(uu^\top+ww^\top)+\sin\delta\,(wu^\top-uw^\top)$. Two facts follow.

(1) \emph{$R$ only moves sources at least as close to $\sig_j$ as $i_l$.} $R_{ab}\neq0$ for $a\neq b$ only if $a,b\in J_l\cup\{i_l\}$. By the ordering, every $a\in J_l\cup\{i_l\}$ other than $j$ has $S_{ja}\le t_l$, so by the quasi-triangle inequality $S_{ab}\le2C_{\rm tri}t_l$ for all $a,b$ in this set.

(2) \emph{$R_{ab}^2=R_{ba}^2$ for $a\neq b$.} The antisymmetric part $\sin\delta\,(w_au_b-u_aw_b)$ of $R_{ab}$ vanishes unless $a$ or $b$ equals $i_l$; in that case the symmetric part $(\cos\delta-1)(u_au_b+w_aw_b)$ vanishes, because $u_{i_l}=0$ and $w$ is zero off $i_l$.

By (2), the two terms $(a,b)$ and $(b,a)$ of \eqref{eq:KL} combine into $R_{ab}^2\big(\sig_a(k)/\sig_b(k)+\sig_b(k)/\sig_a(k)-2\big)$, whose average over $k$ is $R_{ab}^2S_{ab}$. Hence, by Lemma~\ref{lem:FI}, (1), and $\sum_{a\neq b}R_{ab}^2\le\|R-I\|_F^2=4(1-\cos\delta)=8\sin^2\theta_l$,
\begin{equation*}
\KL\big(P_{Q_\varepsilon}\,\|\,P_{Q_{\varepsilon'}}\big)=\frac n2\sum_{a<b}R_{ab}^2S_{ab}\le\frac n2\cdot2C_{\rm tri}t_l\cdot4\sin^2\theta_l\le4C_{\rm tri}\,n\,t_l\,\theta_l^2=8B^4\eta\,\frac{n}{n+E}\le\frac18 .
\end{equation*}
By Pinsker's inequality the total variation between neighbouring hypotheses is at most $1/4$.

\emph{Assouad's lemma.} Average \eqref{eq:hamming} over $\varepsilon$ uniform on the cube. For each $l$, pairing $\varepsilon$ with its neighbour in coordinate $l$ gives $\Prob_\varepsilon(\hat\varepsilon_l\neq\varepsilon_l)+\Prob_{\varepsilon'}(\hat\varepsilon_l\neq\varepsilon'_l)\ge1-\mathrm{TV}\ge3/4$. Hence \citep[weighted form of Thm.~2.12]{tsybakov2009introduction}
\begin{equation*}
\sup_Q\E\rho_j(\hat Q)\ \ge\ \frac12\sum_l\tau_l^2\cdot\frac38\ \ge\ \frac3{64}\sum_l\theta_l^2\ =\ \frac{3\eta}{64}\cdot\frac{E_j}{n+E_j},
\end{equation*}
which is \eqref{eq:lb2} with $c=3\eta/64=3/(4096B^4)$. The argument bounds a Bayes risk under the uniform prior on the cube; since the Bayes risk of a randomised estimator is an average of Bayes risks of deterministic ones, \eqref{eq:lb2} holds for randomised estimators as well. If some $S_{ji}=0$, then $\sig_i=\sig_j$ and rotations in the plane $(j,i)$ do not change the law of the data. The truths $Q_0$ and $Q_0R(\pi/4)$ have $j$-th columns at angle $\pi/4$, and every $\hat Q\in\Od$ has $\rho_j^{(1)}+\rho_j^{(2)}\ge2\sin^2(\pi/8)$: whether one column serves both truths or two orthogonal columns do, the two angles sum to at least $\pi/4$. If both are at most $\pi/4$, convexity of $\sin^2$ on $[0,\pi/4]$ gives the claim; otherwise one term alone exceeds $\frac12>2\sin^2(\pi/8)$. Hence $\sup_Q\E\rho_j\ge\sin^2(\pi/8)>1/8$, which is \eqref{eq:lb2} with $E_j=\infty$.

\paragraph{Average and maximal losses \eqref{eq:lb3}--\eqref{eq:lb4}.} The bound for $\lM^2$ follows from $\lM^2\ge\rho_j$ for every $j$. For $\lA^2$ we symmetrise. Fix an estimator $\hat Q$, write $X$ for the full sample (a $d\times n$ matrix whose columns are the observations of all environments), and let $H$ be a Haar-distributed element of $\Od$, independent of $X$. Define the randomised estimator
\begin{equation*}
\tilde Q(X)=H^\top\hat Q(HX).
\end{equation*}
Under the truth $Q$, with known signatures, $HX$ has the law of the sample under the truth $HQ$. The loss is invariant under a common rotation: $\langle H^\top\hat q_a(HX),q_j\rangle=\langle\hat q_a(HX),Hq_j\rangle$ and $Hq_j$ is the $j$-th column of $HQ$, so $\rho_j(\tilde Q(X);Q)=\rho_j(\hat Q(HX);HQ)$. Taking expectations, with $r_j(G)=\E_{Y\sim P_G}\,\rho_j(\hat Q(Y);G)$ the risk function of $\hat Q$,
\begin{equation*}
\E_Q\,\rho_j(\tilde Q)=\E_H\,r_j(HQ)=\int_{\Od}r_j\,d\mu=:\bar r_j,
\end{equation*}
because $HQ$ is Haar-distributed for every fixed $Q$. The right side does not depend on $Q$. Applying \eqref{eq:lb2} to $\tilde Q$ gives $\bar r_j\ge cE_j/(n+E_j)$ for every $j$. Finally, for the original estimator,
\begin{equation*}
\sup_Q\E_Q\,\lA^2(\hat Q)\ \ge\ \int_{\Od}\E_G\,\lA^2(\hat Q)\,d\mu(G)\ =\ \frac1d\sum_j\bar r_j\ \ge\ \frac cd\sum_j\frac{E_j}{n+E_j}.\qquad\square
\end{equation*}

\paragraph{Attainment (b).} Here $\hat Q_{\rm ML}$ maximises the Gaussian likelihood over $Q\in\Od$ and all positive diagonal $\Lambda_k$, without the box $[1/B,B]$, so the true parameter is an interior point. The block-Gaussian model is a regular parametric family in $(Q,\{\Lambda_k\})$; by Lemma~\ref{lem:FI} the Fisher information for the rotation angle in plane $(j,i)$ is $nS_{ij}$, different planes are orthogonal at first order, and the variances are orthogonal nuisance parameters. With $S_{ij}>0$ for all pairs the model is identifiable up to signs and permutations, and the maximum-likelihood estimator is asymptotically normal with inverse-information covariance \citep{flury1986asymptotic,pham2001blind,yeredor2010blind}. In the local coordinates $\hat Q_{\rm ML}=Q\exp(\Xi)$ the angles $\sqrt n\,\Xi_{ji}$ converge jointly to independent $N(0,S_{ij}^{-1})$, and $\rho_j=\sum_{i\neq j}\Xi_{ji}^2+O(\|\Xi\|^3)$ gives the limit law, whose mean is $E_j$. $\square$

\section{Proof of Theorem~\ref{thm:upper}, Lemma~\ref{lem:probe} and Corollary~\ref{cor:minimax}}
\label{app:upper}

\begin{theoremfull}[Theorem~\ref{thm:upper}, full statement]\label{thm:upper-full}
Let the sources be Gaussian and the signatures known, and put $w_k=\sig_j(k)^{-2}$ and $\gamma_j=K(d+6\log n)/(nS_{\rm nn}(j))$. There are $c,C>0$ depending only on $B$ such that, if $\gamma_j\le c$, the following hold.
\begin{enumerate}[leftmargin=*,itemsep=1pt,topsep=2pt]
\item[(a)] (Known whitening.) Let $A=Q\in\Od$ and $2K\mid n$, split each environment's sample into halves with covariances $\Sh^{(1)}_k,\Sh^{(2)}_k$, and let $\hat q_j$ be a unit eigenvector for the smallest eigenvalue of
\begin{equation*}
\hat H_j=\sum_kw_k\,\mathrm{sym}\big[(\Sh^{(1)}_k-\sig_j(k)I)(\Sh^{(2)}_k-\sig_j(k)I)\big].
\end{equation*}
Then
\begin{equation}\label{eq:upper}
\E\big[1-\langle\hat q_j,q_j\rangle^2\big]\ \le\ \frac{E^{\rm qp}_j}{n}\big(1+C\sqrt{\gamma_j}\big)^2+\frac1{n^2},
\end{equation}
where the first-order coefficient $E^{\rm qp}_j$ (defined below) satisfies $E_j\le E^{\rm qp}_j\le\kappa_BE_j$, $\kappa_B=\frac14(B^2+B^{-2})^2$.
\item[(b)] (Estimated whitening.) Let $A$ be any invertible matrix and $3K\mid n$. Split each environment's sample into three equal parts, whiten parts $1$ and $2$ by $\hat C^{-1/2}$, $\hat C$ the pooled sample covariance of part $0$, compute $\hat q_j$ as in (a) from them, and let $\hat b_j=\hat C^{-1/2}\hat q_j$. The estimated source is $\hat b_j^\top x=z^\top s$ with $z=A^\top\hat b_j$, and $\rho_j=1-z_j^2/\|z\|^2$ satisfies, for every $A$,
\begin{equation}\label{eq:upper-wh}
\E\rho_j\ \le\ \frac3{2n}\big(E^{\rm qp}_j+2E^{\rm wh}_j\big)\big(1+C\sqrt{\gamma_j}\big)^2+\frac1{n^2},
\end{equation}
where the coefficient $E^{\rm wh}_j$ of the estimated whitening (defined in the proof of part (b)) satisfies $E^{\rm wh}_j\le B^4E_j$.
\end{enumerate}
\end{theoremfull}
The simplified statement of Section~\ref{sec:upper} follows: under $\gamma_j\le c$, $(1+C\sqrt{\gamma_j})^2\le(1+C\sqrt c)^2$, $E^{\rm qp}_j+2E^{\rm wh}_j\le(\kappa_B+2B^4)E_j$, and $n^{-2}\le(B^2+B^{-2}-2)E_j/n$ because $E_j\ge(d-1)/(B^2+B^{-2}-2)$ and $d\ge2$. In (b), $\hat b_j^\top x=z^\top s$ has pooled variance $\|z\|^2$ (the sources have pooled variance one), of which $\|z\|^2-z_j^2$ comes from the sources other than $j$, so $\rho_j$ is the share of the pooled variance of the estimated source that comes from the other sources.

Throughout, $j$ is fixed, $C$ denotes a constant depending only on $B$ that may change from line to line, and we work in the basis of $Q$ (the estimator is equivariant: $\hat H_j$ computed from $Qx$ is $Q\hat H_jQ^\top$). The sample covariances are uncentred, $\Sh^{(i)}_k=\frac1m\sum xx^\top$ over the $m=n/(2K)$ samples of half $i$ (we assume $2K\mid n$), $N^{(i)}_k=\Sh^{(i)}_k-\Lambda_k$, $\bar N_k=\frac12(N^{(1)}_k+N^{(2)}_k)$ (the noise of the full sample of environment $k$), $\tilde\Lambda_k=\Lambda_k-\sig_j(k)I$ and $\delta_a=\sig_a-\sig_j$. Expanding the product,
\begin{equation*}
\hat H_j=H+P_1+P_2,\qquad H=\sum_kw_k\tilde\Lambda_k^2=\diag(\lambda_a),\quad P_1=\sum_kw_k(\tilde\Lambda_k\bar N_k+\bar N_k\tilde\Lambda_k),\quad P_2=\sum_kw_k\,\mathrm{sym}(N^{(1)}_kN^{(2)}_k),
\end{equation*}
with $\lambda_a=\sum_kw_k\delta_a(k)^2$ and $\lambda_j=0$. With $r_k=\sig_a(k)/\sig_j(k)\in[B^{-2},B^2]$ we have $w_k\delta_a(k)^2=(r_k-1)^2$ and $KS_{aj}=\sum_k(r_k-1)^2/r_k$, so $B^{-2}KS_{aj}\le\lambda_a\le B^2KS_{aj}$ and $\lambda_*:=\min_{a\neq j}\lambda_a\ge KS_{\rm nn}(j)/B^2$. A prime denotes restriction to $e_j^\perp$: $H'=\diag(\lambda_a)_{a\neq j}$, $P'=P_1'+P_2'$, and $h=(\hat H_je_j)_{-j}=h^{(1)}+h^{(2)}$ with $h^{(1)}_a=\sum_kw_k\delta_a(k)\bar N_{k,aj}$ and $h^{(2)}=(P_2e_j)_{-j}$. The first-order error vector is $g=H'^{-1}h$, and $\E\|H'^{-1}h^{(1)}\|^2=\sum_{a\neq j}\sum_kw_k^2\delta_a(k)^2\sig_a(k)\sig_j(k)/(n_k\lambda_a^2)=E^{\rm qp}_j/n$ with $E^{\rm qp}_j=\sum_{i\neq j}v^{\rm qp}_{ij}$ and, writing $\delta=\sig_i-\sig_j$, $v^{\rm qp}_{ij}=K\sum_kw_k^2\delta_k^2\sig_i(k)\sig_j(k)\big/(\sum_kw_k\delta_k^2)^2$. The bounds $S_{ij}^{-1}\le v^{\rm qp}_{ij}\le\kappa_BS_{ij}^{-1}$ follow from Cauchy--Schwarz, $(\sum_kw_k\delta_k^2)^2\le\sum_kw_k^2\delta_k^2\sig_i\sig_j\cdot\sum_k\delta_k^2/(\sig_i\sig_j)$, and from Kantorovich's inequality applied to $v^{\rm qp}_{ij}S_{ij}=\sum(r-1)^2r\cdot\sum(r-1)^2r^{-1}/(\sum(r-1)^2)^2$ with $r_k\in[B^{-2},B^2]$, which gives $v^{\rm qp}_{ij}S_{ij}\le\frac14(B^2+B^{-2})^2=:\kappa_B$.

\paragraph{Step 1: relative boundedness.} Let $\nu=\sum_kw_k\|\bar N_k\|^2$ and $\pi_2=\|P_2\|$ (operator norms). For $x\perp e_j$,
\begin{equation*}
|x^\top P_1x|=2\Big|\sum_kw_k(\tilde\Lambda_kx)^\top(\bar N_kx)\Big|\le2\Big(\sum_kw_k\|\tilde\Lambda_kx\|^2\Big)^{1/2}\Big(\sum_kw_k\|\bar N_kx\|^2\Big)^{1/2}\le2\sqrt{x^\top Hx}\,\sqrt\nu\,\|x\|,
\end{equation*}
because $\sum_kw_k\|\tilde\Lambda_kx\|^2=\sum_a\lambda_ax_a^2$. With $x=H'^{-1/2}y$ this gives $\|E_1\|\le2\sqrt{\nu/\lambda_*}$ for $E_1=H'^{-1/2}P_1'H'^{-1/2}$: the noise is small relative to the nearest-neighbour gap $\lambda_*$, not relative to the spread of the $\lambda_a$. Similarly, $H'^{-1/2}h^{(1)}=\sum_kM_ky_k$ with $M_k=\sqrt{w_k}H'^{-1/2}\tilde\Lambda_k'$ and $y_k=\sqrt{w_k}(\bar N_ke_j)_{-j}$; since $\sum_kM_kM_k^\top=I$, $\|H'^{-1/2}h^{(1)}\|^2\le\sum_kw_k\|\bar N_ke_j\|^2\le\nu$. Finally $\|H'^{-1/2}h^{(2)}\|\le\lambda_*^{-1/2}\pi_2$.

\paragraph{Step 2: the good event.} For a $d\times m$ standard Gaussian matrix $Z$ and $s>0$, all singular values of $Z/\sqrt m$ lie in $1\pm(\sqrt d+s)/\sqrt m$ with probability at least $1-2e^{-s^2/2}$ \citep[Cor.~5.35]{vershynin2012introduction}, so $\|ZZ^\top/m-I\|\le3(\sqrt d+s)/\sqrt m$ when the latter is at most one. Since $N=\Lambda_k^{1/2}(ZZ^\top/m-I)\Lambda_k^{1/2}$, with $s=\sqrt{6\log n}$ the event $\mathcal G$ that $\|N^{(i)}_k\|\le B\zeta$ and $\|\bar N_k\|\le B\zeta$ for all $k$ and $i$, where $\zeta^2=36K(d+6\log n)/n$, has $\Prob(\mathcal G^c)\le4Kn^{-3}\le n^{-2}$ (the bound for $\bar N_k$ follows from those for the halves). The condition $(\sqrt d+s)/\sqrt m\le1$ holds because $9(\sqrt d+s)^2/m\le\zeta^2=36\gamma_jS_{\rm nn}(j)$ and $S_{\rm nn}(j)\le B^2+B^{-2}-2$, once $c\le1/(36(B^2+B^{-2}-2))$; then also $n\ge36K\cdot6\log n\ge4K$. On $\mathcal G$, since $w_k\le B^2$, $\nu\le KB^4\zeta^2$ and $\pi_2\le KB^4\zeta^2$, hence $\nu/\lambda_*\le36B^6\gamma_j$ and $\pi_2/\lambda_*\le36B^6\gamma_j$.

\paragraph{Step 3: Schur complement.} Let $\mu_0=(\hat H_j)_{jj}=(P_2)_{jj}$, so $|\mu_0|\le\pi_2$, and $A_0=(\hat H_j)_{-j,-j}=H'+P'$. By Step 1, $A_0-\mu_0\succeq H'^{1/2}(I+E_1)H'^{1/2}-2\pi_2I\succeq(1-e_0)H'$ with $e_0=2\sqrt{\nu/\lambda_*}+2\pi_2/\lambda_*$, and $e_0\le1/4$ on $\mathcal G$ once $\gamma_j\le c$. The smallest eigenvalue $\mu$ of $\hat H_j$ satisfies $\mu\le e_j^\top\hat H_je_j=\mu_0$, so $A_0-\mu\succ0$. Any eigenvector for $\mu$ then has a nonzero $j$-th coordinate (otherwise its other coordinates $v'$ would satisfy $(A_0-\mu)v'=0$), so the eigenvector is unique and equals $(1,w)/\sqrt{1+\|w\|^2}$ with $w=-(A_0-\mu)^{-1}h$. Hence
\begin{equation*}
\rho_j:=1-\langle\hat q_j,q_j\rangle^2=\frac{\|w\|^2}{1+\|w\|^2}\le\|w\|^2 .
\end{equation*}
From $\mu=\mu_0-h^\top(A_0-\mu)^{-1}h$ and $(A_0-\mu)^{-1}\preceq(A_0-\mu_0)^{-1}\preceq(1-e_0)^{-1}H'^{-1}$ we get $|\mu|\le\pi_2+2\|H'^{-1/2}h\|^2\le\pi_2+4\nu+4\pi_2^2/\lambda_*$, so $|\mu|/\lambda_*\le C\gamma_j$ on $\mathcal G$.

\paragraph{Step 4: expansion.} Write $A_0-\mu=H'^{1/2}(I+E)H'^{1/2}$ with $E=E_1+H'^{-1/2}(P_2'-\mu)H'^{-1/2}$; on $\mathcal G$, $\|E\|\le e:=2\sqrt{\nu/\lambda_*}+(\pi_2+|\mu|)/\lambda_*\le C\sqrt{\gamma_j}\le\frac12$. The identity $(I+E)^{-1}=I-E+E(I+E)^{-1}E$ gives
\begin{equation*}
w=-g+H'^{-1}(P'-\mu)H'^{-1}h-H'^{-1/2}E(I+E)^{-1}EH'^{-1/2}h .
\end{equation*}
Let $T_2=H'^{-1}P_1'H'^{-1}h^{(1)}$ and $T_3=H'^{-1/2}P_1'H'^{-1}h^{(1)}$, the two second-order terms that are polynomials of the noise alone. Splitting $h=h^{(1)}+h^{(2)}$ and $P'-\mu=P_1'+(P_2'-\mu)$ and using Step 1 for the remaining pieces, on $\mathcal G$,
\begin{equation*}
\|w\|\le\big(1+C\gamma_j\big)\|g\|+\|T_2\|+C\sqrt{\gamma_j/\lambda_*}\,\big(\|T_3\|+\|H'^{-1/2}h^{(2)}\|\big).
\end{equation*}
For instance $\|H'^{-1}P_1'H'^{-1}h^{(2)}\|=\|H'^{-1/2}E_1H'^{-1/2}h^{(2)}\|\le\lambda_*^{-1/2}\|E_1\|\,\|H'^{-1/2}h^{(2)}\|$, $\|H'^{-1}(P_2'-\mu)H'^{-1}h\|\le\lambda_*^{-1}(\pi_2+|\mu|)\|g\|$, and the last term of $w$ has norm at most $2e\lambda_*^{-1/2}\|H'^{-1/2}(P'-\mu)H'^{-1}h\|$, which is bounded in the same way.

\paragraph{Step 5: moments.} Off-diagonal entries of the noise are uncorrelated across pairs and environments, with $\Var(\bar N_{k,ab})=\sig_a(k)\sig_b(k)/n_k$ and $\Var(N^{(i)}_{k,ab})=\sig_a(k)\sig_b(k)/m$ for a half, $m=n/(2K)$. For one environment with $m'$ Gaussian samples and $a,b,b'\neq j$ with $a\notin\{b,b'\}$, expanding the four sample averages over time indices,
\begin{equation*}
\E\big[N_{ab}N_{bj}N_{ab'}N_{b'j}\big]=\mathbf 1[b=b']\Big(1-\frac1{m'}\Big)\Var(N_{ab})\Var(N_{bj})+\frac{\chi_{bb'}}{m'^3},\qquad|\chi_{bb'}|\le3B^4,
\end{equation*}
(the only pairing of time indices that survives is $\{ab,ab'\}\{bj,b'j\}$, and coincident indices contribute $m'^{-3}\E[s_a^2s_b^2s_{b'}^2s_j^2]$, i.e.\ $\chi_{bb'}=\sig_a\sig_b\sig_{b'}\sig_j$ for $b\neq b'$ and $3\sig_a\sig_b^2\sig_j$ for $b=b'$); entries of different environments are independent, and $\E[N_{ab}N_{bj}]=0$. We use $w_k\le B^2$, $\|\delta_a\|^2\le B^2\lambda_a$, $\lambda_a\ge KS_{aj}/B^2$, $\|\delta_a\|^2\le KB^2S_{aj}$ and $v^{\rm qp}_{aj}\le\kappa_BS_{aj}^{-1}$, which give, with $E^{(4)}_j=\sum_{i\neq j}S_{ij}^{-2}$,
\begin{equation}\label{eq:sums}
\sum_{a\neq j}\frac{\|\delta_a\|^2}{\lambda_a^2}\le\frac{B^4E_j}K,\qquad\sum_{a\neq j}\frac1{\lambda_a^2}\le\frac{B^4E^{(4)}_j}{K^2},\qquad\sum_{a\neq j}\frac1{\lambda_a}\le\frac{B^2E_j}K,\qquad\sum_{b\neq j}\|\delta_b\|^2v^{\rm qp}_{bj}\le\kappa_BB^2K(d-1).
\end{equation}
\emph{The term $T_2$.} $(T_2)_a=\lambda_a^{-1}\sum_{b\neq j}(P_1')_{ab}\lambda_b^{-1}h^{(1)}_b$, where $(P_1')_{ab}=\sum_kw_k(\delta_a+\delta_b)(k)\bar N_{k,ab}$ for $b\neq a$ and $h^{(1)}_b=\sum_kw_k\delta_b(k)\bar N_{k,bj}$. In $\E[(P_1')_{ab}(P_1')_{ab'}h^{(1)}_bh^{(1)}_{b'}]$ each factor is a sum over environments; by independence across environments and $\E[N_{ab}N_{bj}]=0$, the pairings of two different environments vanish unless they are $\{ab,ab'\}\{bj,b'j\}$, which forces $b=b'$, and one environment gives the display above. Hence, for the terms with $b\neq a$ and up to the $\chi$ terms,
\begin{equation*}
\E\|T_2\|^2\le\sum_{a\neq j}\frac1{\lambda_a^2}\sum_{b\neq j}\frac{\Var((P_1')_{ab})\,\E(h^{(1)}_b)^2}{\lambda_b^2}\le\frac{2B^6K}{n^2}\Big[\sum_a\frac{\|\delta_a\|^2}{\lambda_a^2}\sum_bv^{\rm qp}_{bj}+\sum_a\frac1{\lambda_a^2}\sum_b\|\delta_b\|^2v^{\rm qp}_{bj}\Big]\le\frac C{n^2}\big(E_j^2+dE^{(4)}_j\big),
\end{equation*}
by $\Var((P_1')_{ab})\le(2B^6K/n)(\|\delta_a\|^2+\|\delta_b\|^2)$, $\E(h^{(1)}_b)^2=\lambda_b^2v^{\rm qp}_{bj}/n$ and \eqref{eq:sums}: the factor $K$ of the noise variance cancels against the factors $1/K$ and $1/K^2$ of the sums. \emph{The term $T_3=H'^{-1/2}P_1'H'^{-1}h^{(1)}$.} The same pairing gives
\begin{equation*}
\E\|T_3\|^2\le\frac{2B^6K}{n^2}\Big[\sum_a\frac{\|\delta_a\|^2}{\lambda_a}\sum_bv^{\rm qp}_{bj}+\sum_a\frac1{\lambda_a}\sum_b\|\delta_b\|^2v^{\rm qp}_{bj}\Big]\le\frac{CK}{n^2}\,dE_j .
\end{equation*}
The diagonal entries $(P_1')_{aa}=2\sum_kw_k\delta_a(k)\bar N_{k,aa}$ ($b=a$) have $\E[N_{aa}^2N_{aj}^2]=(1+4/m')\Var(N_{aa})\Var(N_{aj})$ and $\E[N_{aa}N_{aj}N_{ab'}N_{b'j}]=2\sig_a^2\sig_{b'}\sig_j/m'^3$, and since $\Var((P_1')_{aa})\le8B^4K\lambda_a/n$ they contribute at most $CE^{(4)}_j/n^2$ to $\E\|T_2\|^2$ and $CKE_j/n^2$ to $\E\|T_3\|^2$. The $\chi$ terms add
\begin{equation*}
\frac{CK}{n^3}\big(d^2E^{(4)}_j+dE_j^2\big)\ \text{ to }\ \E\|T_2\|^2,\qquad \frac{CK^2}{n^3}\,d^2E_j\ \text{ to }\ \E\|T_3\|^2,
\end{equation*}
through $|w_k(\delta_a+\delta_b)(k)|\le B^2(\|\delta_a\|+\|\delta_b\|)$, $\sum_k|w_k\delta_b(k)\,w_k\delta_{b'}(k)|\le B^4\|\delta_b\|\|\delta_{b'}\|$ (so the sum over environments of the factors $1/m'^3=(K/n)^3$ adds no further $K$), $\lambda_b\ge\|\delta_b\|^2/B^2$, $\sum_b\|\delta_b\|^{-2}\le B^2E_j/K$ and $\sum_b(\|\delta_a\|+\|\delta_b\|)/\|\delta_b\|\le d+\|\delta_a\|\sqrt{d\sum_b\|\delta_b\|^{-2}}$, with $\|\delta_a\|^2$ absorbed into $\lambda_a^{-2}\le B^4\|\delta_a\|^{-4}$ (respectively $\lambda_a^{-1}$); note also $dE_j^2\le d^2E^{(4)}_j$. \emph{The term $h^{(2)}$.} $h^{(2)}_a=\sum_kw_k\sum_b\frac12\big(N^{(1)}_{k,ab}N^{(2)}_{k,bj}+N^{(2)}_{k,ab}N^{(1)}_{k,bj}\big)$ with independent halves; different $k$, and different $b$, are uncorrelated, and $\E[N_{ab}N_{bj}]=0$, so
\begin{equation*}
\E(h^{(2)}_a)^2=\sum_kw_k^2\,\frac12\sum_b\Var(N^{(1)}_{k,ab})\Var(N^{(2)}_{k,bj}),
\end{equation*}
which lies between $cK^3d/n^2$ and $CK^3d/n^2$; hence $\E\|H'^{-1}h^{(2)}\|^2\le CKdE^{(4)}_j/n^2$ and $\E\|H'^{-1/2}h^{(2)}\|^2\le CK^2dE_j/n^2$ by \eqref{eq:sums}. Since $E_j\le d/S_{\rm nn}(j)$ and $E^{(4)}_j\le E_j/S_{\rm nn}(j)$, every one of these bounds, after the factors of Step~4, is at most $C\gamma_j\,E_j/n$: for example $(E_j^2+dE^{(4)}_j)/n^2\le2(E_j/n)\,d/(nS_{\rm nn}(j))$, $(\gamma_j/\lambda_*)\,\E\|T_3\|^2\le C\gamma_j^2E_j/n$ and $KdE^{(4)}_j/n^2\le\gamma_jE_j/n$.

\paragraph{Step 6: conclusion.} On $\mathcal G^c$, $\rho_j\le1$. By Minkowski's inequality on $\mathcal G$ and $E_j\le E^{\rm qp}_j$,
\begin{equation*}
\E\rho_j\le\Big(\big(1+C\gamma_j\big)\big(\E\|g\|^2\big)^{1/2}+C\sqrt{\gamma_j}\big(E_j/n\big)^{1/2}\Big)^2+\Prob(\mathcal G^c)\le\frac{E^{\rm qp}_j}n\big(1+C\sqrt{\gamma_j}\big)^2+\frac1{n^2},
\end{equation*}
using $\E\|g\|^2\le\E\|H'^{-1}h^{(1)}\|^2+\E\|H'^{-1}h^{(2)}\|^2\le(E^{\rm qp}_j/n)(1+C\gamma_j)$ (the cross term vanishes because $h^{(2)}$ has mean zero given either half while $h^{(1)}$ is linear in the halves). $\square$

\paragraph{Computation.} $\hat H_j$ combines the $2K+1$ matrices $\mathrm{sym}(\Sh^{(1)}_k\Sh^{(2)}_k)$, $\Sh^{(1)}_k+\Sh^{(2)}_k$ and $I$, formed once in $O(Kd^3)$, so full eigendecompositions cost $O(d^4)$ for all sources. On $\mathcal G$ the gap of $\hat H_j$ above its bottom eigenvalue is at least $\frac34\lambda_*$ (Step~3) and its largest eigenvalue is $O(K)$ with a constant depending on $B$, so a Lanczos solver for the bottom eigenvector, at $O(d^2)$ per iteration, needs about $S_{\rm nn}(j)^{-1/2}\log(d/\varepsilon)$ iterations from a random start, up to factors of $B$: $O(Kd^3+d^3S_{\min}^{-1/2}\log(d/\varepsilon))$ for all sources.

\paragraph{Without splitting.} The split pencil averaged over all balanced splits of each environment is the U-statistic $\sum_kw_k\frac{n_k}{n_k-1}\big[(\hat\Sigma_k-\sig_j(k)I)^2-n_k^{-2}\sum_t(x_tx_t^\top-\sig_j(k)I)^2\big]$, the sum over the $n_k$ observations $x_t$ of environment $k$, with $\hat\Sigma_k$ their covariance: the full-sample pencil with the bias of the squared noise removed by a correction computable from the data. We analyse one split because its halves are independent; the U-statistic would need a decoupling argument.

\paragraph{Estimated whitening: proof of part (b).} \emph{Reduction.} Let $N^{(i)}_k$ ($i=0,1,2$) be the noise of the sample covariance of part $i$ in environment $k$, in source coordinates, so that part $i$ has covariances $A(\Lambda_k+N^{(i)}_k)A^\top$, and let $\Npool =\frac1K\sum_kN^{(0)}_k$. Since the parts have equal sizes and $\frac1K\sum_k\Lambda_k=I$, $\hat C=A(I+\Npool )A^\top$. Put $M=\hat C^{-1/2}A$. Then $M(I+\Npool )M^\top=I$, so $M^\top M=(I+\Npool )^{-1}$, and since $\sig_j(k)I=\sig_j(k)M(I+\Npool )M^\top$,
\begin{equation*}
\hat C^{-1/2}A(\Lambda_k+N^{(i)}_k)A^\top\hat C^{-1/2}-\sig_j(k)I=MY^{(i)}_kM^\top,\qquad Y^{(i)}_k=\tilde\Lambda_k+N^{(i)}_k-\sig_j(k)\Npool  .
\end{equation*}
The pencil of (a) on the whitened parts is therefore $MXM^\top$ with $X=\sum_kw_k\,\mathrm{sym}[Y^{(1)}_k(I+\Npool )^{-1}Y^{(2)}_k]$, and $y$ is a bottom eigenvector of $MXM^\top$ iff $z=M^\top y$ solves $Xz=\mu(I+\Npool )z$, with $\mu$ the smallest generalised eigenvalue. As $\hat b_j^\top x=y^\top Ms$, this is the $z$ of the theorem, and $\rho_j=1-z_j^2/\|z\|^2$ is a function of $X$ and $\Npool $, that is, of the noise in source coordinates: its law is the same for every invertible $A$. (For $\Npool =0$ and $A=Q$ this is part (a).) We work in the source basis, where $H=\diag(\lambda_a)$ as before, with parts $1$ and $2$ as the halves.

\emph{Expansion and first order.} With $(I+\Npool )^{-1}=I-\Npool +\Npool (I+\Npool )^{-1}\Npool $,
\begin{equation*}
X=H+P_1+P_2+W_1+W_2,\qquad W_1=-\sum_kw_k\big[\sig_j(k)(\Npool \tilde\Lambda_k+\tilde\Lambda_k\Npool )+\tilde\Lambda_k\Npool \tilde\Lambda_k\big],
\end{equation*}
where $H,P_1,P_2$ are those of (a) and $W_2$ consists of $\sum_kw_k\sig_j(k)^2(\Npool )^2=K(\Npool )^2$, $-\sum_kw_k\sig_j(k)\,\mathrm{sym}(\Npool N^{(2)}_k+N^{(1)}_k\Npool )$, $-\sum_kw_k\,\mathrm{sym}[(N^{(1)}_k-\sig_j(k)\Npool )\Npool \tilde\Lambda_k+\tilde\Lambda_k\Npool (N^{(2)}_k-\sig_j(k)\Npool )+(N^{(1)}_k-\sig_j(k)\Npool )\Npool (N^{(2)}_k-\sig_j(k)\Npool )]$ and $\sum_kw_k\,\mathrm{sym}[Y^{(1)}_k\Npool (I+\Npool )^{-1}\Npool Y^{(2)}_k]$, each of second order in the noise. Since $\tilde\Lambda_ke_j=0$, $(W_1)_{jj}=0$ and $(W_1e_j)_a=-\beta_a\Npool _{aj}$ with $\beta_a=\sum_kw_k\sig_j(k)\delta_a(k)=\sum_k\delta_a(k)/\sig_j(k)$. The entries $\Npool _{aj}$, $a\neq j$, are uncorrelated, independent of parts $1$ and $2$, with variance $\frac1{Kn_0}\sum_k\sig_a(k)\sig_j(k)$, $n_0=n/3$; so the first-order part $H'^{-1}(h^{(1)}+h^{\rm w})$ of the error vector, $h^{\rm w}=(W_1e_j)_{-j}$, has second moment $E^{\rm qp}_j/n_{12}+E^{\rm wh}_j/n_0$ with $n_{12}=2n/3$, $E^{\rm wh}_j=\sum_{a\neq j}v^{\rm wh}_{aj}$ and $v^{\rm wh}_{aj}=\beta_a^2\,\frac1K\sum_k\sig_a(k)\sig_j(k)\big/\lambda_a^2$, which is the leading term of \eqref{eq:upper-wh}. By Cauchy--Schwarz, $\beta_a^2\le\sum_kw_k\delta_a(k)^2\cdot\sum_kw_k\sig_j(k)^2=K\lambda_a$, so $v^{\rm wh}_{aj}\le B^2K/\lambda_a\le B^4S_{aj}^{-1}$; $\beta_a=0$ for every $a$ when $\sig_j$ is constant, since $\sum_k\delta_a(k)=0$.

\emph{Steps 1--4.} Let $\mathcal G$ also require $\|N^{(0)}_k\|\le B\zeta$ for every $k$, with $\zeta$ defined from parts of $n/3$ observations (this changes only constants); then $\|\Npool \|\le B\zeta\le\frac14$. For every $x$, since $\sum_kw_k\sig_j(k)^2=K$ and $\tilde\Lambda_ke_j=0$,
\begin{equation*}
|x^\top W_1x|\le2\sqrt K\,\|\Npool \|\,\sqrt{x^\top Hx}\,\|x\|+\|\Npool \|\,x^\top Hx,
\end{equation*}
so $W_1$ is relatively bounded like $P_1$, with $\nu$ replaced by $\nu+K\|\Npool \|^2$ and an additional $\|\Npool \|$ in the bound on $\|E_1\|$; also $\|H'^{-1/2}h^{\rm w}\|^2\le K\|\Npool \|^2$. Each term of $W_2$ is bounded in the same way, $|x^\top(\cdot)x|\le C\zeta^2\big(x^\top Hx+\sqrt K\sqrt{x^\top Hx}\,\|x\|\big)+CK\zeta^2\|x\|^2$; since $\|H\|\le CK$, also $\|W_2\|\le CK\zeta^2$, which plays the role of $\pi_2$ in Steps~3 and~4. Writing $z=(1,w)$, with $1$ in the $j$-th coordinate, the generalised eigenvalue equation gives
\begin{equation*}
w=-\big(A_0-\mu(I+\Npool )'\big)^{-1}(h-\mu\bar e),\qquad\mu(1+\Npool _{jj})=X_{jj}-(h-\mu\bar e)^\top\big(A_0-\mu(I+\Npool )'\big)^{-1}(h-\mu\bar e),
\end{equation*}
with $A_0=X_{-j,-j}$, $h=(Xe_j)_{-j}$ and $\bar e=(\Npool e_j)_{-j}$, and $\rho_j\le\|w\|^2$ as in Step~3. On $\mathcal G$, $X\succeq-CK\zeta^2I$ and $X_{jj}=(P_2+W_2)_{jj}\le CK\zeta^2$, so $|\mu|\le CK\zeta^2\le C\gamma_j\lambda_*$. By the relative bounds, $A_0-\mu(I+\Npool )'\succeq(1-e)H'$ with $e\le C\sqrt{\gamma_j}\le\frac12$, so as in Step~3 the bottom generalised eigenvector is unique and has $z_j\neq0$; Steps~3 and~4 go through with $A_0-\mu$ replaced by $A_0-\mu(I+\Npool )'$ and $h$ by $h-\mu\bar e$: the first change adds at most $|\mu|(1+\|\Npool \|)/\lambda_*\le C\gamma_j$ to the bounds on $\|E\|$, the second adds $\|H'^{-1}\mu\bar e\|\le C\gamma_j\|\bar e\|$ to $\|w\|$, with $\E\|\bar e\|^2\le2B^2(d-1)/n_0\le CE_j/n$.

\emph{Step 5: the remainder.} Write $Q_1=P_1'+W_1'$ for the first-order part of $X'-H'$, $h^{\rm w2}=(W_2e_j)_{-j}$, and $g=g_1+g_2$ with $g_1=H'^{-1}(h^{(1)}+h^{\rm w})$ and $g_2=H'^{-1}(h^{(2)}+h^{\rm w2})$. Step~4 bounds $\|w\|$ by $(1+C\gamma_j)\|g\|$, the analogues $H'^{-1}Q_1H'^{-1}(h^{(1)}+h^{\rm w})$ and $H'^{-1/2}Q_1H'^{-1}(h^{(1)}+h^{\rm w})$ of $T_2$ and $T_3$, and $C\sqrt{\gamma_j/\lambda_*}\,\|H'^{-1/2}(h^{(2)}+h^{\rm w2})\|$. The following lemma, the only place where part (b) needs new moment bounds, controls every term that involves $\Npool $; the terms without $\Npool $ are those of part (a).

\begin{lemma}[Remainder of part (b)]\label{lem:remainder}
In the setting of Theorem~\ref{thm:upper}(b) there is $C$ depending only on $B$ such that, uniformly in $d$ and $K$: $(W_1)_{ab}=-c_{ab}\Npool _{ab}$ with $c_{ab}^2\le CK(\|\delta_a\|^2+\|\delta_b\|^2)$ and $\E(h^{\rm w}_b)^2\le CK\lambda_b/n$, so that the four vectors $H'^{-1}Q_1H'^{-1}h$ and $H'^{-1/2}Q_1H'^{-1}h$ with $h\in\{h^{(1)},h^{\rm w}\}$ obey the bounds of $T_2$ and $T_3$ in Step~5 of part (a); and
\begin{equation}\label{eq:remainder}
\E\big[\mathbf 1_{\mathcal G}\|H'^{-1}h^{\rm w2}\|^2\big]\le C\gamma_j\,\frac{E_j}n,\qquad
\E\big[\mathbf 1_{\mathcal G}\|H'^{-1/2}h^{\rm w2}\|^2\big]\le C\gamma_j\lambda_*\,\frac{E_j}n .
\end{equation}
\end{lemma}
\begin{proof}
Each part has $m_0=n/(3K)$ observations per environment, so $\Var(N^{(i)}_{k,ab})\le2B^2/m_0$ and $\Var(\Npool _{ab})\le2B^2/n_0$, $n_0=n/3$, for all entries and $i\in\{0,1,2\}$; moreover $\E\|\Npool \|_F^2\le2B^2d^2/n_0$, $\E\|N^{(i)}_ke_j\|^2\le CKd/n$, $\E\|\Npool e_j\|^2\le Cd/n$, and on $\mathcal G$, $\|N^{(i)}_k\|\le B\zeta$, $\|\Npool \|\le B\zeta\le\frac14$ and $\zeta^2\le C\gamma_jS_{\rm nn}(j)$. Bounds that use $\|\Npool \|\le B\zeta$ are proved on the larger event $\mathcal G_{\rm wh}\supseteq\mathcal G$ that the whitening part has $\|N^{(0)}_k\|\le B\zeta$ for every $k$. It depends only on part $0$, so it is independent of parts $1$ and $2$, and $\E[\mathbf 1_{\mathcal G}X]\le\E[\mathbf 1_{\mathcal G_{\rm wh}}X]$ for $X\ge0$. We use three devices. (D1) \emph{Independence across environments:} for vectors $c_k$ that are functions of part $0$ and of the other half, $\E\big[(\sum_kw_k(N_kc_k)_a)^2\,\big|\,\text{part }0,\text{ other half}\big]=\sum_kw_k^2\sum_bc_{k,b}^2\Var(N_{k,ab})\le\frac{CK}n\sum_k\|c_k\|^2$ for every $a$, since $\E[N_{k,ab}N_{k,ab'}]=0$ for $b\neq b'$ and different environments are independent; as $\mathbf 1_{\mathcal G_{\rm wh}}$ is a function of part $0$, the same bound holds for $\E[\mathbf 1_{\mathcal G_{\rm wh}}(\cdot)^2]$ with $\E[\mathbf 1_{\mathcal G_{\rm wh}}\sum_k\|c_k\|^2]$ on the right. (D2) \emph{Weights:} if $\E v_a^2\le\sigma^2$ for every $a\neq j$, then by \eqref{eq:sums} $\E\|H'^{-1}v\|^2\le\sigma^2B^4E^{(4)}_j/K^2$ and $\E\|H'^{-1/2}v\|^2\le\sigma^2B^2E_j/K$; if instead $\E v_a^2\le\sigma^2\lambda_a$, then $\E\|H'^{-1}v\|^2\le\sigma^2B^2E_j/K$ and $\E\|H'^{-1/2}v\|^2\le\sigma^2d$; both hold verbatim with $\E$ replaced by $\E[\mathbf 1_{\mathcal G_{\rm wh}}\,\cdot\,]$. (D3) \emph{The regime:} $Kd/n\le\gamma_jS_{\rm nn}(j)$, $E^{(4)}_j\le E_j/S_{\rm nn}(j)$, $d\le CE_j$, $S_{\rm nn}(j)\le C$ and $\lambda_*\ge KS_{\rm nn}(j)/B^2$; hence $\sigma^2=CK^2d\,\zeta^{2r}/n^2$ with $r\in\{0,1,2\}$ in the first case of (D2), and $\sigma^2=CKd\,\zeta^{2r}/n^2$ in the second, give exactly \eqref{eq:remainder}. For instance, in the first case $\E\|H'^{-1}v\|^2\le CdE^{(4)}_j/n^2\le C(d/(nS_{\rm nn}(j)))E_j/n\le C\gamma_jE_j/n$ and $\E\|H'^{-1/2}v\|^2\le CKdE_j/n^2\le C\gamma_j\lambda_*E_j/n$ because $\gamma_j\lambda_*\ge K^2d/(nB^2)$.

\emph{$W_1$ and $h^{\rm w}$.} From $W_1=-\sum_kw_k[\sig_j(k)(\Npool \tilde\Lambda_k+\tilde\Lambda_k\Npool )+\tilde\Lambda_k\Npool \tilde\Lambda_k]$, $c_{ab}=\sum_kw_k[\sig_j(k)(\delta_a+\delta_b)(k)+\delta_a(k)\delta_b(k)]$, and Cauchy--Schwarz gives $c_{ab}^2\le CK(\|\delta_a\|^2+\|\delta_b\|^2)$; $h^{\rm w}_b=-\beta_b\Npool _{bj}$ with $\beta_b^2\le K\lambda_b$. So $\Var((W_1)_{ab})$ and $\E(h^{\rm w}_b)^2$ are bounded like $\Var((P_1')_{ab})$ and $\E(h^{(1)}_b)^2$ with $n_0$ in place of $n$, and the entries of $\Npool $, a sum of independent environment noises, have the pairing structure of Step~5 of part (a) (the only surviving pairing of two entries of $\Npool $ against two entries of $\Npool $, $N^{(1)}$ or $N^{(2)}$ is again $\{ab,ab'\}\{bj,b'j\}$); the computation of $\E\|T_2\|^2$ and $\E\|T_3\|^2$ therefore applies verbatim to the four mixed vectors.

\emph{$W_2$.} Expanding $X=\sum_kw_k\,\mathrm{sym}[Y^{(1)}_k(I-\Npool +R)Y^{(2)}_k]$ with $R=\Npool (I+\Npool )^{-1}\Npool $, $\|R\|\le2B^2\zeta^2$ and $\|R\|_F\le2B\zeta\|\Npool \|_F$ on $\mathcal G_{\rm wh}$, and using $\tilde\Lambda_ke_j=0$ and $Y^{(i)}_ke_j=b^{(i)}_k:=N^{(i)}_ke_j-\sig_j(k)\Npool e_j$, the vector $W_2e_j$ is the sum of five classes of terms: (i) $K(\Npool )^2e_j$; (ii) $-\frac12\sum_kw_k\sig_j(k)\big[N^{(1)}_k\Npool e_j+\Npool N^{(1)}_ke_j+\Npool N^{(2)}_ke_j+N^{(2)}_k\Npool e_j\big]$; (iii) $-\frac12\sum_kw_k\tilde\Lambda_k\Npool \big(b^{(1)}_k+b^{(2)}_k\big)$ and (iv) $-\frac12\sum_kw_k\big[(N^{(1)}_k-\sig_j(k)\Npool )\Npool b^{(2)}_k+(N^{(2)}_k-\sig_j(k)\Npool )\Npool b^{(1)}_k\big]$, which together are $-\sum_kw_k\,\mathrm{sym}[Y^{(1)}_k\Npool Y^{(2)}_k]e_j$; and (v) $\frac12\sum_kw_k\big[Y^{(1)}_kRb^{(2)}_k+Y^{(2)}_kRb^{(1)}_k\big]$, whose $\tilde\Lambda_k$-terms are (iii) with $\Npool $ replaced by $R$ and whose other terms are (iv) with the inner $\Npool $ replaced by $R$. We bound each class, always restricted to the coordinates $a\neq j$.

(i) $(K(\Npool )^2)_{aj}=K\sum_b\Npool _{ab}\Npool _{bj}$, and the pairing identity of Step~5 gives $\E(K(\Npool )^2)_{aj}^2\le K^2\big[\sum_b\Var(\Npool _{ab})\Var(\Npool _{bj})+Cd^2/n_0^3\big]\le CK^2d/n^2$ for every $a$ (the two terms with $b\in\{a,j\}$ involve diagonal entries, whose moments are bounded as at the end of Step~5 of part (a)): (D2) with $r=0$.

(ii) By (D1), conditionally on $\Npool $, $\E(\sum_kw_k\sig_j(k)(N^{(i)}_k\Npool e_j)_a)^2\le\frac{CK}n\cdot KB^2\|\Npool e_j\|^2$ and $\E(\sum_kw_k\sig_j(k)(\Npool N^{(i)}_ke_j)_a)^2=\sum_kw_k^2\sig_j(k)^2\sum_b(\Npool _{ab})^2\Var(N^{(i)}_{k,bj})\le\frac{CK^2}n\|\Npool _{a\cdot}\|^2$; taking expectations, $\E v_a^2\le CK^2d/n^2$ for every $a$: (D2) with $r=0$.

(iii) The $a$-th coordinate is $-\frac12\sum_kw_k\delta_a(k)u_{k,a}$ with $u_{k,a}=(\Npool N^{(2)}_k)_{aj}+(N^{(1)}_k\Npool )_{ja}-2\sig_j(k)((\Npool )^2)_{aj}$. For the first two terms (D1), applied with the factor $\delta_a(k)$ inside the sum over $k$, gives $\E\big[(\sum_kw_k\delta_a(k)(\Npool N^{(2)}_k)_{aj})^2\,\big|\,\text{part }0\big]=\sum_kw_k^2\delta_a(k)^2\sum_b(\Npool _{ab})^2\Var(N^{(2)}_{k,bj})\le B^2\lambda_a\cdot\frac{CK}n\|\Npool _{a\cdot}\|^2$, and the same for $(N^{(1)}_k\Npool )_{ja}$, with $\E\|\Npool _{a\cdot}\|^2\le2B^2d/n_0$; the third term is $\beta_a((\Npool )^2)_{aj}$ with $\beta_a^2\le K\lambda_a$ and $\E((\Npool )^2)_{aj}^2\le Cd/n_0^2$. Hence $\E v_a^2\le CKd\lambda_a/n^2$: the second case of (D2) with $r=0$, which gives $\E\|H'^{-1}v\|^2\le CdE_j/n^2\le C\gamma_jE_j/n$ and $\E\|H'^{-1/2}v\|^2\le CKd^2/n^2\le C\gamma_j\lambda_*E_j/n$.

(iv) Split $(N^{(1)}_k-\sig_j(k)\Npool )\Npool b^{(2)}_k=N^{(1)}_k(\Npool b^{(2)}_k)-\sig_j(k)(\Npool )^2b^{(2)}_k$. For the first part, (D1) conditionally on parts $0$ and $2$ gives $\E[\mathbf 1_{\mathcal G_{\rm wh}}v_a^2]\le\frac{CK}n\E\big[\mathbf 1_{\mathcal G_{\rm wh}}\sum_k\|\Npool b^{(2)}_k\|^2\big]\le\frac{CK}n\,B^2\zeta^2\,\E\sum_k\|b^{(2)}_k\|^2\le CK^3d\zeta^2/n^2$, since $\E\sum_k\|b^{(2)}_k\|^2\le CK^2d/n$; and (D2) with $\sigma^2=CK^3d\zeta^2/n^2$ gives $\E[\mathbf 1_{\mathcal G_{\rm wh}}\|H'^{-1}v\|^2]\le CKd\zeta^2E^{(4)}_j/n^2\le C\gamma_j(Kd/n)E_j/n\le C\gamma_jE_j/n$ and $\E[\mathbf 1_{\mathcal G_{\rm wh}}\|H'^{-1/2}v\|^2]\le CK^2d\zeta^2E_j/n^2\le C\gamma_j\lambda_*E_j/n$, by (D3). For the second part write $(\Npool )^2\sum_kw_k\sig_j(k)b^{(2)}_k=(\Npool )^2z_0-K(\Npool )^3e_j$ with $z_0=\sum_kw_k\sig_j(k)N^{(2)}_ke_j$, whose coordinates are centred and uncorrelated given $\Npool $ with $\E z_{0,b}^2\le CK^2/n$; then $\E[\mathbf 1_{\mathcal G_{\rm wh}}\|(\Npool )^2z_0\|^2]\le\frac{CK^2}n\E[\mathbf 1_{\mathcal G_{\rm wh}}\|(\Npool )^2\|_F^2]\le\frac{CK^2}n\,B^2\zeta^2\,\frac{Cd^2}{n}$, and $\|H'^{-1}(\Npool )^2z_0\|\le\lambda_*^{-1}\|(\Npool )^2z_0\|$ gives $\E[\mathbf 1_{\mathcal G_{\rm wh}}\|H'^{-1}(\Npool )^2z_0\|^2]\le Cd^2\zeta^2/(n^2S_{\rm nn}(j)^2)\le C\gamma_j^2E_j/n$ and $\E[\mathbf 1_{\mathcal G_{\rm wh}}\|H'^{-1/2}(\Npool )^2z_0\|^2]\le CKd^2\zeta^2/(n^2S_{\rm nn}(j))\le C\gamma_j\lambda_*E_j/n$, again by (D3). The remaining term is a function of $\Npool $ alone: $\E[\mathbf 1_{\mathcal G_{\rm wh}}K^2\|(\Npool )^3e_j\|^2]\le K^2B^4\zeta^4\,\E\|\Npool e_j\|^2\le CK^2\zeta^4d/n$, which after the factor $\lambda_*^{-2}$ is at most $C\zeta^4d/(nS_{\rm nn}(j)^2)\le C\gamma_j^2d/n\le C\gamma_j^2E_j/n$, and after $\lambda_*^{-1}$ at most $CK\zeta^4d/(nS_{\rm nn}(j))\le C\gamma_j\lambda_*E_j/n$. The terms with $1$ and $2$ exchanged are identical.

(v) The $\tilde\Lambda_k$-terms are (iii) with $\Npool $ replaced by $R$. For the first two parts of (iii) the per-coordinate bound becomes $\E[\mathbf 1_{\mathcal G_{\rm wh}}v_a^2]\le B^2\lambda_a\frac{CK}n\E[\mathbf 1_{\mathcal G_{\rm wh}}\|R_{a\cdot}\|^2]\le CK\zeta^4\lambda_a/n$, and the second case of (D2) with $\sigma^2=CK\zeta^4/n$ gives $\E[\mathbf 1_{\mathcal G_{\rm wh}}\|H'^{-1}v\|^2]\le C\zeta^4E_j/n\le C\gamma_j^2E_j/n$ and $\E[\mathbf 1_{\mathcal G_{\rm wh}}\|H'^{-1/2}v\|^2]\le CKd\zeta^4/n\le C\gamma_j\lambda_*E_j/n$. The third part, $\beta_a(R\Npool )_{aj}$, is summed over $a$ instead: $\sum_a\beta_a^2(R\Npool )_{aj}^2/\lambda_a^2\le(K/\lambda_*)\|R\Npool e_j\|^2$ and $\sum_a\beta_a^2(R\Npool )_{aj}^2/\lambda_a\le K\|R\Npool e_j\|^2$, with $\E[\mathbf 1_{\mathcal G_{\rm wh}}\|R\Npool e_j\|^2]\le4B^4\zeta^4\,Cd/n$, which gives $C\gamma_j^2S_{\rm nn}(j)d/n\le C\gamma_j^2E_j/n$ and $CK\zeta^4d/n\le C\gamma_j\lambda_*E_j/n$. The other terms are (iv) with the inner $\Npool $ replaced by $R$, bounded in the same way with $\|R\|\le2B^2\zeta^2$ in place of $\|\Npool \|\le B\zeta$, $\|\Npool R\|_F\le\|\Npool \|\,\|R\|_F$ in place of $\|(\Npool )^2\|_F$ and $K\Npool R\Npool e_j$ in place of $K(\Npool )^3e_j$: the resulting bounds are those of (iv) multiplied by $\zeta^2\le C$. Every class therefore satisfies \eqref{eq:remainder}, classes (i)--(iii) even without the indicator and (iv)--(v) with $\mathbf 1_{\mathcal G_{\rm wh}}\ge\mathbf 1_{\mathcal G}$; summing the five classes proves the lemma.
\end{proof}

\emph{Step 6.} By Lemma~\ref{lem:remainder} and Step~5 of part (a), $\E[\mathbf 1_{\mathcal G}\|g_2\|^2]\le C\gamma_jE_j/n$, the second moment of $g_1$ is exactly $\frac3{2n}(E^{\rm qp}_j+2E^{\rm wh}_j)$, and the $T_2$-, $T_3$-type terms of Step~4 are at most $C\gamma_jE_j/n$ in mean square after the factors of Step~4. Unlike in (a), some cross terms between $g_1$ and $g_2$ need not vanish (for instance $h^{\rm w}$ against $K((\Npool )^2)_{\cdot j}$), so we do not expand $\E\|g\|^2$: Minkowski's inequality, as in Step~6 with $E_j\le E^{\rm qp}_j$, bounds $(\E\rho_j\mathbf 1_{\mathcal G})^{1/2}$ by $(1+C\gamma_j)(\E\|g_1\|^2)^{1/2}+C(\gamma_jE_j/n)^{1/2}$, which gives \eqref{eq:upper-wh}. $\square$

\paragraph{Perturbed probes.} The estimator uses $\sig_j$ only through the shifts $\sig_j(k)I$, which commute with every $\Lambda_k$, and through the weights. The next lemma replaces both by a probe $\hat\sig$.

\begin{lemma}[Perturbed probes]\label{lem:probe}
In the setting of Theorem~\ref{thm:upper}(a), let $\hat\sig\in[1/B,B]^K$ be independent of the two halves, and let $\hat q_j$ be a unit bottom eigenvector of $\hat H_j$ with $\sig_j$ replaced by $\hat\sig$ in the shifts and in the weights, $\hat w_k=\hat\sig(k)^{-2}$. Write $\hat\lambda_a=\sum_k\hat w_k(\sig_a(k)-\hat\sig(k))^2$ and $\hat\delta_a=\delta_a-2(\hat\sig-\sig_j)$, and suppose that
\begin{equation}\label{eq:probe-acc}
\|\hat\sig-\sig_j\|^2\ \le\ \eta^2K\,S_{\rm nn}(j)/B^6 .
\end{equation}
There are $c,C>0$ depending only on $B$ such that, if $\eta\le1/8$ and $\gamma_j\le c$,
\begin{equation*}
\E\big[1-\langle\hat q_j,q_j\rangle^2\,\big|\,\hat\sig\big]\le\frac{\hat E^{\rm qp}_j}{n}\big(1+C\sqrt{\gamma_j}\big)^2+\frac1{n^2},\qquad
\hat E^{\rm qp}_j=\sum_{a\neq j}\frac{K\sum_k\hat w_k^2\hat\delta_a(k)^2\sig_a(k)\sig_j(k)}{(\hat\lambda_a-\hat\lambda_j)^2},
\end{equation*}
where $E_j\le\hat E^{\rm qp}_j\le B^8\big(\frac{1+2\eta}{1-2\eta}\big)^2E_j\le3B^8E_j$, and $\hat E^{\rm qp}_j\le(1+C\eta)E^{\rm qp}_j$ if moreover $\max_k|\hat\sig(k)/\sig_j(k)-1|\le\eta$.
\end{lemma}
\begin{proof}
Condition on $\hat\sig$. Put $\varpi_k=\sqrt{\hat w_k}\,(\hat\sig(k)-\sig_j(k))$ and, for $a\neq j$, $u_a(k)=\sqrt{\hat w_k}\,\delta_a(k)$. Since $\hat\sig\in[1/B,B]^K$, $\hat w_k\in[B^{-2},B^2]$, so \eqref{eq:S-euclid} gives $\|u_a\|^2\ge B^{-2}\|\delta_a\|^2\ge KS_{aj}/B^4$, while $\|\varpi\|^2\le B^2\|\hat\sig-\sig_j\|^2\le\eta^2KS_{\rm nn}(j)/B^4\le\eta^2\|u_a\|^2$. Also $\|\hat\sig-\sig_j\|\le\eta\|\delta_a\|/B^2$ for every $a\neq j$.

\emph{(i) The population pencil.} $\sum_k\hat w_k(\Lambda_k-\hat\sig(k)I)^2=\diag(\hat\lambda_a)$ with $\hat\lambda_a=\|u_a-\varpi\|^2$ for $a\neq j$ and $\hat\lambda_j=\|\varpi\|^2$. Subtracting $\hat\lambda_jI$, which does not change eigenvectors, the estimator is the bottom eigenvector of $\tilde H+\hat P_1+\hat P_2$ with $\tilde H=\diag(\tilde\lambda_a)$, $\tilde\lambda_a=\hat\lambda_a-\hat\lambda_j=\|u_a\|^2-2\langle u_a,\varpi\rangle\in[(1-2\eta)\|u_a\|^2,(1+2\eta)\|u_a\|^2]$, $\tilde\lambda_j=0$, $\hat P_1=\sum_k\hat w_k(\hat\Lambda_k\bar N_k+\bar N_k\hat\Lambda_k)$ with $\hat\Lambda_k=\Lambda_k-\hat\sig(k)I$, and $\hat P_2=\sum_k\hat w_k\,\mathrm{sym}(N^{(1)}_kN^{(2)}_k)$. Hence $\tilde\lambda_*:=\min_{a\neq j}\tilde\lambda_a\ge(1-2\eta)KS_{\rm nn}(j)/B^4$ and $\hat\lambda_j\le\eta^2\tilde\lambda_*/(1-2\eta)$.

\emph{(ii) Steps~1--4.} Let $\hat\nu=\sum_k\hat w_k\|\bar N_k\|^2$ and $\hat\pi_2=\|\hat P_2\|$; on $\mathcal G$, $\hat\nu\le KB^4\zeta^2$ and $\hat\pi_2\le KB^4\zeta^2$ as before, so $\hat\nu/\tilde\lambda_*$ and $\hat\pi_2/\tilde\lambda_*$ are at most $C\gamma_j$. For $x\perp e_j$, the computation of Step~1 gives $|x^\top\hat P_1x|\le2(\sum_k\hat w_k\|\hat\Lambda_kx\|^2)^{1/2}\sqrt{\hat\nu}\,\|x\|$ with $\sum_k\hat w_k\|\hat\Lambda_kx\|^2=x^\top\tilde Hx+\hat\lambda_j\|x\|^2\le(1+2\eta^2)\,x^\top\tilde Hx$, so $\|\tilde H'^{-1/2}\hat P_1'\tilde H'^{-1/2}\|\le3\sqrt{\hat\nu/\tilde\lambda_*}$. The entries of the first-order vector are $h^{(1)}_a=\sum_k\hat w_k\hat\delta_a(k)\bar N_{k,aj}$, because $(\sig_a-\hat\sig)+(\sig_j-\hat\sig)=\hat\delta_a$, and $\tilde H'^{-1/2}h^{(1)}=\sum_k\hat M_ky_k$ with $\hat M_k=\sqrt{\hat w_k}\,\tilde H'^{-1/2}\diag(\hat\delta_a(k))_{a\neq j}$; since $\sqrt{\hat w}\,\hat\delta_a=u_a-2\varpi$, $\sum_k\hat M_k\hat M_k^\top=\diag(\|u_a-2\varpi\|^2/\tilde\lambda_a)\preceq(1+2\eta)^2(1-2\eta)^{-1}I$, and $\|\tilde H'^{-1/2}h^{(1)}\|^2\le(1+C\eta)\hat\nu$. The diagonal entry is now $\mu_0=(\hat P_1)_{jj}+(\hat P_2)_{jj}$ with $(\hat P_1)_{jj}=2\sum_k\hat w_k(\sig_j(k)-\hat\sig(k))\bar N_{k,jj}$, so $|\mu_0|\le\hat\pi_2+2\sqrt{\hat\lambda_j\hat\nu}\le\hat\pi_2+3\eta\sqrt{\hat\nu/\tilde\lambda_*}\,\tilde\lambda_*$. With $\lambda_*,\nu,\pi_2$ replaced by $\tilde\lambda_*$, $(1+C\eta)\hat\nu$ and $\hat\pi_2+2\sqrt{\hat\lambda_j\hat\nu}$, Steps~3 and~4 go through with two changes, both because $\mu_0$ now contains the linear term $(\hat P_1)_{jj}$: Step~3 gives $|\mu|/\tilde\lambda_*\le C(\gamma_j+\eta\sqrt{\gamma_j})$ instead of $C\gamma_j$, and the factor $1+C\gamma_j$ in front of $\|g\|$ in Step~4 becomes $1+C\sqrt{\gamma_j}$. Step~6 needs only the latter, and on $\mathcal G$ every ratio that Steps~3 and~4 bound is still at most $C\sqrt{\gamma_j}$.

\emph{(iii) Steps~5--6.} The moment computations of Step~5 have the same form, with $\delta_a$ replaced by $\hat\delta_a$ in $h^{(1)}$ and by $\sig_a-\hat\sig$ in the entries of $\hat P_1$, $w$ by $\hat w$, and $\lambda_a$ by $\tilde\lambda_a$. They used only $w_k\le B^2$, $\lambda_a\ge B^{-2}\|\delta_a\|^2$, $KS_{aj}/B^2\le\|\delta_a\|^2\le KB^2S_{aj}$, $|w_k(\delta_a+\delta_b)(k)|\le B^2(\|\delta_a\|+\|\delta_b\|)$, and $v^{\rm qp}_{bj}\le\kappa_BS_{bj}^{-1}$ through $\E(h^{(1)}_b)^2=\lambda_b^2v^{\rm qp}_{bj}/n$; the last is replaced by $\hat v_{bj}\le3B^8S_{bj}^{-1}$, the bound on each term of $\hat E^{\rm qp}_j$ proved below. Since $\|\hat\sig-\sig_j\|\le\eta\|\delta_a\|/B^2$, the vectors $\hat\delta_a$ and $\sig_a-\hat\sig$ have norms within factors $1\pm2\eta$ of $\|\delta_a\|$, and $\tilde\lambda_a\ge(1-2\eta)B^{-2}\|\delta_a\|^2$, so every bound of Step~5 holds with a constant depending only on $B$. Step~6 gives the claim, with $\E\|\tilde H'^{-1}h^{(1)}\|^2=\hat E^{\rm qp}_j/n$.

\emph{The coefficient.} Write $\hat v_{aj}$ for the $a$-th term of $\hat E^{\rm qp}_j$. Since $\tilde\lambda_a=\sum_k\hat w_k\delta_a(k)\hat\delta_a(k)$, Cauchy--Schwarz gives $\tilde\lambda_a^2\le\sum_k\hat w_k^2\hat\delta_a(k)^2\sig_a(k)\sig_j(k)\cdot\sum_k\delta_a(k)^2/(\sig_a(k)\sig_j(k))$, that is $\hat v_{aj}\ge S_{aj}^{-1}$, so $\hat E^{\rm qp}_j\ge E_j$ exactly as in Theorem~\ref{thm:upper}; this is what Step~6 uses. Since $\hat w_k^2\sig_a(k)\sig_j(k)\le B^4\hat w_k$ and $\sqrt{\hat w}\,\hat\delta_a=u_a-2\varpi$, the numerator of $\hat v_{aj}$ is at most $KB^4\|u_a-2\varpi\|^2\le KB^4(1+2\eta)^2\|u_a\|^2$, and the denominator $\tilde\lambda_a^2$ is at least $(1-2\eta)^2\|u_a\|^4$; with $\|u_a\|^2\ge KS_{aj}/B^4$ this gives $\hat v_{aj}\le B^8\big(\frac{1+2\eta}{1-2\eta}\big)^2S_{aj}^{-1}$, which is at most $3B^8S_{aj}^{-1}$ for $\eta\le1/8$. If also $|\hat\sig(k)/\sig_j(k)-1|\le\eta$ for every $k$, then $\hat w_k/w_k\in[(1+\eta)^{-2},(1-\eta)^{-2}]$, and since $\|\hat\delta_a-\delta_a\|\le2\eta\|\delta_a\|/B^2$, each term of $\hat E^{\rm qp}_j$ is within a factor $1+C\eta$ of the corresponding term of $E^{\rm qp}_j$.
\end{proof}

Signatures known to relative accuracy $\eta\sqrt{S_{\rm nn}(j)}$ and clipped to $[1/B,B]^K$ (clipping does not increase their distance to $\sig_j$, which lies in the box) satisfy \eqref{eq:probe-acc} with $\eta$ replaced by $B^4\eta$. In general \eqref{eq:probe-acc} holds for a probe within $\eta/B^4$ of the distance from $\sig_j$ to its nearest neighbour, since $KS_{\rm nn}(j)\ge\min_{i\neq j}\|\sig_i-\sig_j\|^2/B^2$ by \eqref{eq:S-euclid}; Theorem~\ref{thm:unknown} provides such probes. The lemma is conservative: with $K=16$ codes ($B=2$, $d\in\{\PluginDs\}$, $n\in\{100d,400d\}$), signatures read off a Jacobi pilot computed on an independent batch change the error by at most $\PluginPilotMax$, although their relative error reaches $\PluginPilotRel\sqrt{S_{\rm nn}(j)}$, far beyond \eqref{eq:probe-acc}.

\paragraph{Corollary~\ref{cor:minimax}.} Theorem~\ref{thm:rate-full}(a) holds for labelled estimators with the same proof: the reduction \eqref{eq:hamming} only uses the vector assigned to $q_j$ (if its angle to $q_j(\varepsilon)$ is below $0.362$, the sign convention makes $\langle\hat q,q_j(\varepsilon)\rangle=\cos\alpha$), and the Haar symmetrisation preserves labels. Under $\gamma\le c$ every source has $\gamma_j\le c$ and $E_{\max}\le d/S_{\min}\le n$, so the lower bound \eqref{eq:lb3} is at least $\frac c2\Ebar/n$, and averaging \eqref{eq:upper} over $j$ gives the upper bound with $C'=2\kappa_B$, after decreasing $c$ if necessary so that $(1+C\sqrt c)^2\le\frac32$ (the term $n^{-2}$ is absorbed because $\gamma\le c$ forces $n\Ebar\ge d$). For (i) the lower bound holds a fortiori, since the orthogonal mixings $A=Q$ are invertible and for them $\rho_j=1-\langle\hat b_j,q_j\rangle^2/\|\hat b_j\|^2$ is the labelled loss; averaging \eqref{eq:upper-wh} over $j$, with $E^{\rm qp}_j\le\kappa_BE_j$ and $E^{\rm wh}_j\le B^4E_j$, gives the upper bound, uniformly in $A$. \emph{Orthogonal estimators.} Let $U$ be the orthogonal polar factor of $\hat V=[\hat q_1,\dots,\hat q_d]$, its nearest orthogonal matrix in Frobenius norm. For the sign matrix $D$ with $\langle D_{jj}\hat q_j,q_j\rangle\ge0$, $UD$ is the polar factor of $\hat VD$, so $\|UD-Q\|_F\le\|UD-\hat VD\|_F+\|\hat VD-Q\|_F\le2\|\hat VD-Q\|_F$. Since $1-\langle u,q\rangle^2\le\|u-q\|^2$ and $\|D_{jj}\hat q_j-q_j\|^2=2(1-|\langle\hat q_j,q_j\rangle|)\le2\rho_j$, $\lA^2(U)\le\frac1d\|UD-Q\|_F^2\le8\,\lA^2(\hat q)$ deterministically: the same rate holds over $\Od$, with $8C'$. Theorem~\ref{thm:unknown} gives the same upper bound over $\Od$ (not over labelled estimators) for an estimator that uses only $B$ and a lower bound $s_0$ on $S_{\min}$; with $s_0=S_{\min}$ its condition is $\gamma\le c$ for a smaller $c$. For well-spread signatures $S_{\min}\asymp d^{-2/(K-1)}$. For separated signatures $E_j\le(d-1)/s_0$, while every configuration has $E_j\ge(d-1)/(B^2+B^{-2}-2)$, because each summand of \eqref{eq:S} is at most $B^2+B^{-2}-2$; hence $\Ebar\asymp d$ with constants depending only on $(B,s_0)$ and not on $K$. Setting (iii), the closest-direction loss of Theorem~\ref{thm:unknown}(b) for estimators that return $d$ directions, is proved in Appendix~\ref{app:unknown} after Lemma~\ref{lem:matched}.

\emph{Separated configurations exist with $K\asymp\log d$.} Let $K$ be even and let $y_1,\dots,y_d\in\{\pm1\}^K$ have $K/2$ entries equal to $+1$ each and pairwise Hamming distances at least $K/4$; for independent uniform balanced words the Hamming distance is twice a hypergeometric variable with mean $K/4$, so by Hoeffding's inequality for sampling without replacement \citep{hoeffding1963probability} a pair violates the condition with probability at most $e^{-K/16}$, and $d$ such words exist once $K\ge32\log d$. Put $\sig_j=\mathbf 1+\epsilon y_j$ with $\epsilon\in(0,1)$. Then $\sum_k\sig_j(k)=K$, $\sig_j(k)\in[1-\epsilon,1+\epsilon]$, and each coordinate where $y_i$ and $y_j$ differ contributes $(2\epsilon)^2/(1-\epsilon^2)$ to $KS_{ij}$, so $S_{ij}=\frac{4\epsilon^2}{1-\epsilon^2}\cdot\frac{\#\{k:y_i(k)\neq y_j(k)\}}K\ge\frac{\epsilon^2}{1-\epsilon^2}$. Conversely, $d$ signatures with $S_{\min}\ge s_0$ are, by \eqref{eq:S-euclid}, $\sqrt{Ks_0}/B$-separated in a $(K-1)$-dimensional body of diameter at most $B\sqrt K$, so a volume count gives $d\le(1+2B^2/\sqrt{s_0})^{K-1}$, i.e.\ $K\gtrsim\log d$. $\square$

\section{Unknown signatures: proof of Theorem~\ref{thm:unknown}}
\label{app:unknown}

\begin{theoremfull}[Theorem~\ref{thm:unknown}, full statement]\label{thm:unknown-full}
Let the sources be Gaussian and let only $B$ and a lower bound $s_0\le S_{\min}$ be known. There are $c,C>0$ depending only on $B$ such that, if $K(d+6\log n)/(ns_0)\le c$, the following hold for every configuration with $S_{\min}\ge s_0$.
\begin{enumerate}[leftmargin=*,itemsep=1pt,topsep=2pt]
\item[(a)] (Known whitening.) Let $A=Q\in\Od$ and $3K\mid n$. Algorithm~\ref{alg:grid} selects probes on a third of the sample by thresholding $\lambda_{\min}\big(\sum_k(\Sh_k-g(k)I)^2\big)$ over a grid of points $g$, runs the pencil of Theorem~\ref{thm:upper}(a) on the other two thirds at each kept probe, and returns the polar factor $\hat Q$ of the $d$ vectors. Then $\sup_Q\E\lA^2(\hat Q)\le C\,\Ebar/n$; the grid has at most $(C/\sqrt{s_0})^K$ points.
\item[(b)] (Estimated whitening.) Let $A$ be any invertible matrix and $4K\mid n$. Split the sample into four equal parts. The second, whitened by its own pooled covariance, selects the probes as in (a), with the mesh and threshold given below; at each kept probe the pencil of Theorem~\ref{thm:upper}(b) is formed from the last two, whitened by the pooled covariance $\hat C$ of the first, giving $\hat b=\hat C^{-1/2}\hat q$. Then $\E\lA^2\le C\,\Ebar/n$ for every $A$, where $\lA^2=\frac1d\sum_j\min_a\big(1-z_{a,j}^2/\|z_a\|^2\big)$ with $z_a=A^\top\hat b_a$ is the loss $\rho_j$ of Theorem~\ref{thm:upper}(b) with each source matched to its closest estimate.
\end{enumerate}
\end{theoremfull}
The proof takes $c=\min\{B^{-8}/27648,\,2c_L/3\}$ in part (a), with $c_L$ the constant of Lemma~\ref{lem:probe}, and a smaller qualitative constant in part (b).

Throughout, $\Omega_B=\{\sig\in\R^K:\sum_k\sig(k)=K\}\cap[1/B,B]^K$, which contains every signature, $c_L$ and $C_L$ are the constants $c,C$ of Lemma~\ref{lem:probe} (which holds for every $\eta\le1/8$), and $\Pi$ is the Euclidean projection onto the convex set $\Omega_B$. We assume $3K\mid n$.

\begin{algorithm}[h]
\caption{Grid probes (the estimator of Theorem~\ref{thm:unknown}(a))}\label{alg:grid}
\begin{algorithmic}
\STATE \textbf{input:} $B$ and $s_0\le S_{\min}$; the sample, split into parts $0$, $1$, $2$ of $n/3$ observations, $n/(3K)$ per environment, with covariances $\Sh^{(0)}_k,\Sh^{(1)}_k,\Sh^{(2)}_k$
\STATE $\eta\leftarrow1/8$, $t\leftarrow\eta\sqrt{Ks_0}/B^3$, $h\leftarrow t/4$, $\bar\varepsilon\leftarrow54K^2B^2(d+6\log n)/n$, $\tau\leftarrow2h^2+2\bar\varepsilon$, $\underline r\leftarrow\sqrt{Ks_0}/B$
\STATE $\mathcal N\leftarrow\{\Pi(x):x\in\{1/B+2hm/\sqrt K:m=0,1,\dots\}^K\cap[1/B,B+2h/\sqrt K]^K\}$
\STATE accept $g\in\mathcal N$ if $\lambda_{\min}\big(\hat H^{(0)}(g)\big)\le\tau$, where $\hat H^{(0)}(g)=\sum_k(\Sh^{(0)}_k-g(k)I)^2$
\STATE cluster the accepted points by single linkage at distance $\underline r/2$; if there are not $d$ clusters, return $I_d$
\STATE from each cluster keep one point $\hat\sig$ (say, with the smallest $\lambda_{\min}$)
\STATE for each kept $\hat\sig$: $\hat q\leftarrow$ unit bottom eigenvector of $\sum_k\hat w_k\,\mathrm{sym}\big[(\Sh^{(1)}_k-\hat\sig(k)I)(\Sh^{(2)}_k-\hat\sig(k)I)\big]$, $\hat w_k=\hat\sig(k)^{-2}$
\STATE \textbf{return} a polar factor $\hat Q$ of the $d\times d$ matrix of these vectors
\end{algorithmic}
\end{algorithm}

\paragraph{The grid.} Every $\sig\in\Omega_B$ has a point $x$ of the cubic grid with $\|x-\sig\|\le\frac{2h}{\sqrt K}\cdot\frac{\sqrt K}2=h$, and $\|\Pi(x)-\sig\|=\|\Pi(x)-\Pi(\sig)\|\le h$ because $\Pi$ is non-expansive. So $\mathcal N\subset\Omega_B$ is an $h$-net of $\Omega_B$. Its size is at most $\big((B-1/B)\sqrt K/(2h)+2\big)^K\le\big(16B^4/\sqrt{s_0}+2\big)^K$. Computing $\sum_k(\Sh^{(0)}_k)^2$ once, each $\hat H^{(0)}(g)$ costs $O(Kd^2)$ and its smallest eigenvalue $O(d^3)$.

\paragraph{Two facts for all probes at once.} Work in the basis of $Q$ and let $N_k=\Sh^{(0)}_k-\Lambda_k$ and $\varepsilon=\sum_k\|N_k\|^2$. For every $g\in\R^K$ and every source $j$,
\begin{align}
\lambda_{\min}\big(\hat H^{(0)}(g)\big)&\le e_j^\top\hat H^{(0)}(g)e_j=\sum_k\big\|(\sig_j(k)-g(k))e_j+N_ke_j\big\|^2\le2\|\sig_j-g\|^2+2\varepsilon,\label{eq:fact1}\\
\lambda_{\min}\big(\hat H^{(0)}(g)\big)&\ge\tfrac12\min_a\|\sig_a-g\|^2-\varepsilon .\label{eq:fact2}
\end{align}
For \eqref{eq:fact2}, write $\hat H^{(0)}(g)=\sum_k(D_k+N_k)^2$ with $D_k=\Lambda_k-g(k)I$. For symmetric $X,Y$, $(X+Y)^2\succeq\frac12X^2-Y^2$, because $(X+Y)^2-\frac12X^2+Y^2=(X/\sqrt2+\sqrt2Y)^2\succeq0$. Hence $\hat H^{(0)}(g)\succeq\frac12\sum_kD_k^2-\sum_kN_k^2\succeq\frac12\diag(\|\sig_a-g\|^2)-\varepsilon I$.

\paragraph{The noise event.} Let $\mathcal G_0$ be the event that $\|N_k\|\le B\omega$ for all $k$, with $\omega=3(\sqrt d+s)/\sqrt{n/(3K)}$ and $s=\sqrt{6\log n}$. As in Step~2 of Appendix~\ref{app:upper}, $\Prob(\mathcal G_0^c)\le2Ke^{-s^2/2}\le n^{-2}$, provided $(\sqrt d+s)^2\le n/(3K)$, which the assumption below implies. On $\mathcal G_0$, $\varepsilon\le KB^2\omega^2\le54K^2B^2(d+6\log n)/n=\bar\varepsilon$.

\paragraph{Selection.} Assume
\begin{equation}\label{eq:grid-cond}
\frac{K(d+6\log n)}{ns_0}\ \le\ \frac{1}{27648\,B^8}=\frac{\eta^2}{432B^8},\qquad\text{equivalently}\qquad\bar\varepsilon\le t^2/8 .
\end{equation}
On $\mathcal G_0$: every signature $\sig_j$ has a point $g\in\mathcal N$ with $\|g-\sig_j\|\le h$, and by \eqref{eq:fact1} it is accepted. By \eqref{eq:fact2}, every accepted $g$ is within $\sqrt{2(\tau+\varepsilon)}\le\sqrt{4h^2+6\bar\varepsilon}\le\sqrt{t^2/4+3t^2/4}=t$ of some signature. Since $S_{ij}\le B^2\|\sig_i-\sig_j\|^2/K$, signatures are at least $\underline r$ apart, and $t=\eta\underline r/B^2\le\underline r/8$. Accepted points near the same signature are therefore within $2t\le\underline r/4$ of each other, and points near different signatures are more than $\underline r-2t\ge\frac34\underline r$ apart. Single linkage at $\underline r/2$ returns exactly one cluster per source, and every kept $\hat\sig$ lies in $\Omega_B$ within $t$ of the signature of its cluster.

\paragraph{Estimation.} Let $j$ be the source of a cluster and $\hat\sig$ its kept point. Since $S_{\rm nn}(j)\ge S_{\min}\ge s_0$, $\|\hat\sig-\sig_j\|^2\le t^2=\eta^2Ks_0/B^6\le\eta^2KS_{\rm nn}(j)/B^6$, which is \eqref{eq:probe-acc}. Parts $1$ and $2$ are independent of part $0$, and their $2n/3$ observations give $\gamma'_j=K(d+6\log(2n/3))/((2n/3)S_{\rm nn}(j))\le\frac32K(d+6\log n)/(ns_0)\le c_L$ under the condition of the theorem with $c\le\frac23c_L$. Lemma~\ref{lem:probe}, applied conditionally on part $0$ with parts $1$ and $2$ as the halves, gives on $\mathcal G_0$
\begin{equation*}
\E\big[1-\langle\hat q,q_j\rangle^2\,\big|\,\text{part }0\big]\le\frac{3\hat E^{\rm qp}_j}{2n}\big(1+C_L\sqrt{c_L}\big)^2+\frac9{4n^2}\le\frac{CE_j}n,
\end{equation*}
using $\hat E^{\rm qp}_j\le3B^8E_j$ and $E_j\ge(d-1)/(B^2+B^{-2}-2)$.

\paragraph{Conclusion.} On $\mathcal G_0$ the clusters label the vectors by sources, and the polar argument of Appendix~\ref{app:upper} (after permuting the columns, a polar factor of the permuted matrix is the permuted polar factor) gives $\lA^2(\hat Q)\le8\cdot\frac1d\sum_j(1-\langle\hat q_{(j)},q_j\rangle^2)$, where $\hat q_{(j)}$ is the vector of the cluster of $\sig_j$. Off $\mathcal G_0$ the loss is at most one. Hence $\E\lA^2(\hat Q)\le8C\Ebar/n+n^{-2}\le C'\Ebar/n$. The law of $N_k$, the event $\mathcal G_0$ and Lemma~\ref{lem:probe} do not depend on $Q$, so the bound holds for every $Q$, which gives the supremum. Part (a) holds with $c=\min\{1/(27648B^8),\,2c_L/3\}$. $\square$

\paragraph{Estimated whitening: proof of part (b).} The four parts $W,0,1,2$ have $n/4$ observations each and $n/(4K)$ per environment; $\zeta$ (the radius called $\omega$ in part (a)), $\bar\varepsilon$ and the event $\mathcal G_0$ are defined as above for parts of this size, $\mathcal G_0$ now the event that $\|N^{(0)}_k\|\le B\zeta$ for all $k$, so that the pooled noise $\Npool =\frac1K\sum_kN^{(0)}_k$ of part $0$ satisfies $\|\Npool \|\le B\zeta\le\frac14$ on $\mathcal G_0$ (the noise of parts $W$, $1$, $2$ is handled inside the lemma below); constants $C_0,C$ depend only on $B$.

\emph{Selection with self-whitening.} Part $0$ is whitened by its own pooled covariance $\hat C_0=A(I+\Npool )A^\top$. With $M_0=\hat C_0^{-1/2}A$, $M_0^\top M_0=(I+\Npool )^{-1}$, and $g(k)I=g(k)M_0(I+\Npool )M_0^\top$,
\begin{equation*}
\hat H^{(0)}(g)=\sum_k\big(\hat C_0^{-1/2}\Sh^{(0)}_k\hat C_0^{-1/2}-g(k)I\big)^2=M_0Z(g)M_0^\top,\qquad Z(g)=\sum_kY_k(I+\Npool )^{-1}Y_k,
\end{equation*}
with $Y_k=\Lambda_k+N^{(0)}_k-g(k)(I+\Npool )$.
Since $\lambda_{\min}(M_0ZM_0^\top)=\min_zz^\top Zz/z^\top(I+\Npool )z$ and $Z\succeq0$, it lies between $\lambda_{\min}(Z)/(1+\|\Npool \|)$ and $\lambda_{\min}(Z)/(1-\|\Npool \|)$. With $\varepsilon'=\sum_k\|N^{(0)}_k\|^2+KB^2\|\Npool \|^2$, the two facts \eqref{eq:fact1}--\eqref{eq:fact2} become $\lambda_{\min}(Z)\le e_j^\top Ze_j\le(1-\|\Npool \|)^{-1}\sum_k\|Y_ke_j\|^2\le(3\|\sig_j-g\|^2+3\varepsilon')/(1-\|\Npool \|)$, from $Y_ke_j=(\sig_j(k)-g(k))e_j+N^{(0)}_ke_j-g(k)\Npool e_j$, and $\lambda_{\min}(Z)\ge(\frac12\min_a\|\sig_a-g\|^2-2\varepsilon')/(1+\|\Npool \|)$, from $(X+Y)^2\succeq\frac12X^2-Y^2$ with $X=\Lambda_k-g(k)I$ and $Y=N^{(0)}_k-g(k)\Npool $. On $\mathcal G_0$ therefore $\lambda_{\min}(\hat H^{(0)}(g))\le6\|\sig_j-g\|^2+6\varepsilon'$ for every $j$ and $\lambda_{\min}(\hat H^{(0)}(g))\ge\frac8{25}\min_a\|\sig_a-g\|^2-\frac{32}{25}\varepsilon'$ (trivially if the right-hand side is negative, since $\hat H^{(0)}(g)\succeq0$), with $\varepsilon'\le\bar\varepsilon':=C_0K^2B^2(d+6\log n)/n$. Algorithm~\ref{alg:grid} is run with mesh $h=t/8$ and threshold $\tau=6h^2+6\bar\varepsilon'$: every signature has an accepted grid point within $h$, and every accepted point is within $\big(\frac{75}4h^2+\frac{91}4\bar\varepsilon'\big)^{1/2}\le t$ of a signature once $\bar\varepsilon'\le t^2/64$, that is, once $K(d+6\log n)/(ns_0)\le\eta^2/(64C_0B^8)$. The clustering step is unchanged, so on $\mathcal G_0$ the kept points $\hat\sig$ are within $t$ of their signatures, one per source, and they are functions of part $0$ alone. The grid has at most $(32B^4/\sqrt{s_0}+2)^K$ points.

\emph{Estimation.} Parts $1$ and $2$ are whitened by $\hat C^{-1/2}$, $\hat C$ the pooled covariance of part $W$, and the pencil of Theorem~\ref{thm:upper}(a) is formed at the probe $\hat\sig$. The following lemma combines Lemma~\ref{lem:probe} with the proof of Theorem~\ref{thm:upper}(b).

\begin{lemma}[Perturbed probes with estimated whitening]\label{lem:probe-wh}
In the setting of Theorem~\ref{thm:upper}(b), with parts $W,1,2$ of $n/4$ observations each, let $\hat\sig\in[1/B,B]^K$ be independent of parts $W$, $1$, $2$ and satisfy \eqref{eq:probe-acc} with $\eta\le1/8$. Let $\hat q$ be a unit bottom eigenvector of the pencil of Theorem~\ref{thm:upper}(a) at the probe $\hat\sig$ (shifts $\hat\sig(k)I$, weights $\hat w_k=\hat\sig(k)^{-2}$) computed from the whitened parts $1$ and $2$, and let $z=A^\top\hat C^{-1/2}\hat q$. There are $c,C>0$ depending only on $B$ such that, if $\gamma_j\le c$, then $\E\big[1-z_j^2/\|z\|^2\,\big|\,\hat\sig\big]\le CE_j/n$ for every invertible $A$.
\end{lemma}
\begin{proof}
Condition on $\hat\sig$ and write $\epsilon_j=\sig_j-\hat\sig$, so that $\hat\Lambda_k:=\Lambda_k-\hat\sig(k)I$ has $\hat\Lambda_ke_j=\epsilon_j(k)e_j$ and, by the proof of Lemma~\ref{lem:probe}, $|\epsilon_j(k)|\le\eta\|\delta_a\|/B^2$ for every $a\neq j$. The reduction of Theorem~\ref{thm:upper}(b) gives, with $M=\hat C^{-1/2}A$, $\Npool$ replaced by $\NpoolW$ and $M^\top M=(I+\NpoolW )^{-1}$, that the pencil is $M\hat XM^\top$ with $\hat X=\sum_k\hat w_k\,\mathrm{sym}[\hat Y^{(1)}_k(I+\NpoolW )^{-1}\hat Y^{(2)}_k]$, $\hat Y^{(i)}_k=\hat\Lambda_k+N^{(i)}_k-\hat\sig(k)\NpoolW $, that $z$ solves $\hat Xz=\mu(I+\NpoolW )z$ for the smallest generalised eigenvalue $\mu$, and that the law of $1-z_j^2/\|z\|^2$ does not depend on $A$. Subtracting $\hat\lambda_j(I+\NpoolW )$, $\hat\lambda_j=\sum_k\hat w_k\epsilon_j(k)^2$, does not change the generalised eigenvectors, and $\tilde X=\hat X-\hat\lambda_j(I+\NpoolW )=\tilde H+\hat P_1+\hat P_2+\hat W_1+\hat W_2$ with $\tilde H=\diag(\tilde\lambda_a)$, $\tilde\lambda_j=0$, as in Lemma~\ref{lem:probe}, $\hat P_1,\hat P_2$ as there, and $\hat W_1,\hat W_2$ the terms of $W_1,W_2$ with $(\tilde\Lambda_k,\sig_j(k),w_k)$ replaced by $(\hat\Lambda_k,\hat\sig(k),\hat w_k)$, together with $-\hat\lambda_j\NpoolW $. Three changes against the two proofs being combined. (1) The first-order whitening term is $(\hat W_1e_j)_a=-\hat\beta_a\NpoolW _{aj}$ with $\hat\beta_a=\sum_k\hat w_k\big(\sig_a(k)\sig_j(k)-\hat\sig(k)^2\big)+\hat\lambda_j$, and $\hat\beta_a^2\le CK\tilde\lambda_a$ by Cauchy--Schwarz, since $\sig_a\sig_j-\hat\sig^2=\sig_j(\sig_a-\hat\sig)+\hat\sig(\sig_j-\hat\sig)$, $\|\sqrt{\hat w}(\sig_a-\hat\sig)\|^2\le(1+\eta)^2\tilde\lambda_a/(1-2\eta)$ and $\|\sqrt{\hat w}\,\epsilon_j\|^2\le\eta^2\tilde\lambda_a/(1-2\eta)$; hence $\E\|\tilde H'^{-1}\hat W_1'e_j\|^2\le CE_j/n$. (2) The relative bounds of Step~1 hold for $\hat W_1$ with $\sum_k\hat w_k\|\hat\Lambda_kx\|^2\le(1+2\eta^2)x^\top\tilde Hx$ for $x\perp e_j$, as in Lemma~\ref{lem:probe}(ii), and $\hat\lambda_j\|\NpoolW \|\,\|x\|^2\le\eta^2\|\NpoolW \|\,x^\top\tilde Hx/(1-2\eta)$; $\|\hat W_2\|\le CK\zeta^2$ as before. The bound on $\mu$ needs the Schur identity. The diagonal entry $(\tilde X)_{jj}=2\sum_k\hat w_k\epsilon_j(k)\bar N_{k,jj}-2\NpoolW _{jj}\big(\sum_k\hat w_k\hat\sig(k)\epsilon_j(k)+\hat\lambda_j\big)+(\hat P_2+\hat W_2)_{jj}$ now contains first-order terms; by Cauchy--Schwarz and $\sum_k\hat w_k\epsilon_j(k)^2=\hat\lambda_j\le\eta^2\tilde\lambda_*/(1-2\eta)$, on $\mathcal G$ it satisfies $|(\tilde X)_{jj}|\le C(\eta\sqrt{\gamma_j}+\gamma_j)\tilde\lambda_*$. The relative bounds of $\hat P_1$ and of $\hat W_1$ without its term $-\hat\lambda_j\NpoolW $ hold against $\diag(\hat\lambda_a)$ for every $x$, so, as in part (b), $\hat X\succeq-CK\zeta^2I$ and $\tilde X\succeq-\big(CK\zeta^2+\hat\lambda_j(1+\|\NpoolW \|)\big)I$; with $\mu\le(\tilde X)_{jj}/(1+\NpoolW _{jj})$ this gives only the a priori bound $|\mu|\le C(\gamma_j+\eta^2)\tilde\lambda_*\le C\tilde\lambda_*$. The identity $\mu(1+\NpoolW _{jj})=(\tilde X)_{jj}-(h-\mu\bar e)^\top(A_0-\mu(I+\NpoolW )')^{-1}(h-\mu\bar e)$ of part (b), with $A_0=\tilde X_{-j,-j}$ and $h=(\tilde Xe_j)_{-j}$, sharpens it. Since $-\mu(I+\NpoolW )'\succeq-2|(\tilde X)_{jj}|\,I$, the relative bounds give $A_0-\mu(I+\NpoolW )'\succeq(1-e_0)\tilde H'$ with $e_0\le C\sqrt{\gamma_j}\le\frac14$; moreover $\|\tilde H'^{-1/2}h\|^2\le CK\zeta^2$, from $\|\tilde H'^{-1/2}h^{(1)}\|^2\le(1+C\eta)\hat\nu$, $\|\tilde H'^{-1/2}(\hat W_1e_j)_{-j}\|^2\le CK\|\NpoolW \|^2$ and $\|\hat P_2+\hat W_2\|\le CK\zeta^2$, and $\|\tilde H'^{-1/2}\bar e\|^2\le B^2\zeta^2/\tilde\lambda_*$. Hence $|\mu|(1-\|\NpoolW \|)\le|(\tilde X)_{jj}|+CK\zeta^2+C\zeta^2\mu^2/\tilde\lambda_*$, and by the a priori bound the last term is at most $C\zeta^2|\mu|\le|\mu|/4$ for $\gamma_j\le c$. So $|\mu|/\tilde\lambda_*\le C(\gamma_j+\eta\sqrt{\gamma_j})$ in place of $C\gamma_j$, as in Lemma~\ref{lem:probe}. (3) Lemma~\ref{lem:remainder} applies to the second-order terms with $(\hat\delta_a,\hat\Lambda_k,\hat w_k,\tilde\lambda_a)$ in place of $(\delta_a,\tilde\Lambda_k,w_k,\lambda_a)$, since its proof uses only $w_k\le B^2$, the pairing structure of the noise, $\lambda_a\asymp\|\delta_a\|^2$ up to factors of $B$, which holds for $\tilde\lambda_a$ and $\|\hat\delta_a\|$ within factors $1\pm2\eta$, and the sums \eqref{eq:sums}; the terms created by $\hat\Lambda_ke_j=\epsilon_j(k)e_j$, which vanished for $\tilde\Lambda_k$, obey the same bounds: $\sum_k\hat w_k\epsilon_j(k)N^{(i)}_k\NpoolW e_j$ and $\big(\sum_k\hat w_k\epsilon_j(k)\hat\sig(k)\big)(\NpoolW )^2e_j$ are the classes (ii) and (i) with $\epsilon_j(k)$ in place of $\sig_j(k)$ and $\sum_k\hat w_k\epsilon_j(k)\hat\sig(k)$ in place of $K$, and $\sum_k\hat w_k\epsilon_j(k)^2=\hat\lambda_j\le\eta^2\tilde\lambda_a/(1-2\eta)$ for every $a\neq j$ keeps their bounds; the same terms with $R$ in place of $\NpoolW $ obey smaller bounds, since $\|Re_j\|\le2\|\NpoolW \|\,\|\NpoolW e_j\|$ on the event $\mathcal G_{\rm wh}$ of Lemma~\ref{lem:remainder} (here for part $W$), whose indicator these bounds carry as there; and the term $\sum_k\hat w_k\epsilon_j(k)\hat\Lambda_kRe_j$, which has one factor $\NpoolW $ fewer than class (v), has $a$-th coordinate $R_{aj}\sum_k\hat w_k\epsilon_j(k)(\sig_a(k)-\hat\sig(k))$ with $|\sum_k\hat w_k\epsilon_j(k)(\sig_a(k)-\hat\sig(k))|\le C\eta\tilde\lambda_a$ by Cauchy--Schwarz, so that $\|\tilde H'^{-1}v\|^2\le C\eta^2\|Re_j\|^2$ and $\|\tilde H'^{-1/2}v\|^2\le C\eta^2K\|Re_j\|^2$, and $\E[\mathbf 1_{\mathcal G_{\rm wh}}\|Re_j\|^2]\le C\zeta^2d/n$ gives the two bounds of \eqref{eq:remainder}. Steps~3--6 of Theorem~\ref{thm:upper}(b) then give $\E[1-z_j^2/\|z\|^2\mid\hat\sig]\le(2\hat E^{\rm qp}_j+CE_j)(1+C\sqrt{\gamma_j})^2/n+n^{-2}\le CE_j/n$ (parts $1$ and $2$ hold $n/2$ observations and part $W$ holds $n/4$), with $\hat E^{\rm qp}_j\le3B^8E_j$ from Lemma~\ref{lem:probe}.
\end{proof}

\emph{Conclusion of (b).} The kept probes are functions of part $0$ and are, on $\mathcal G_0$, within $t$ of their signatures, which is \eqref{eq:probe-acc} with $\eta=1/8$ as in part (a); parts $W$, $1$, $2$ are independent of part $0$, and their sizes give $\gamma'_j\le2K(d+6\log n)/(ns_0)\le c$ under the condition of the theorem. By Lemma~\ref{lem:probe-wh}, conditionally on part $0$ and on $\mathcal G_0$, the vector of the cluster of source $j$ has $\E[1-z_j^2/\|z\|^2]\le CE_j/n$, and the closest-direction loss of the theorem is at most this labelled loss. Off $\mathcal G_0$ the loss is at most one. Hence $\E\lA^2\le C\Ebar/n+n^{-2}\le C'\Ebar/n$ for every invertible $A$, since the laws of the noises do not depend on $A$ or $Q$. $\square$

\paragraph{The closest-direction loss is minimax too: Corollary~\ref{cor:minimax}(iii).} Theorem~\ref{thm:unknown}(b) returns $d$ unmixing directions that need not be orthogonal in the true whitened coordinates, and its loss matches every source to its closest estimate, so two sources may be matched to the same estimate. The lower bound of Theorem~\ref{thm:rate-full}(a) is for estimators in $\Od$. The next lemma transfers it.

\begin{lemma}[Closest-direction loss and the polar factor]\label{lem:matched}
Let $Q\in\Od$ have columns $q_j$ and let $V=[v_1,\dots,v_d]$ have unit columns, not necessarily orthogonal. Put $\varrho_j=1-\max_a\langle q_j,v_a\rangle^2$, and let $U\in\Od$ be an orthogonal matrix nearest to $V$ in Frobenius norm (for nonsingular $V$, its polar factor $V(V^\top V)^{-1/2}$). Then
\begin{equation*}
\sum_{j=1}^d\rho_j(U)\le16\sum_{j=1}^d\varrho_j,\qquad\rho_j(U)=1-\max_a\langle q_j,u_a\rangle^2 .
\end{equation*}
\end{lemma}
\begin{proof}
Let $J=\{j:\varrho_j<\frac12\}$ and, for $j\in J$, let $a(j)$ be a column with $\langle q_j,v_{a(j)}\rangle^2=1-\varrho_j$. The map $a$ is injective on $J$: if $a(i)=a(j)=a$ with $i\neq j$, then $\langle q_i,v_a\rangle^2+\langle q_j,v_a\rangle^2\le\|v_a\|^2=1$, because $q_i\perp q_j$, so $\varrho_i+\varrho_j\ge1$, which contradicts $\varrho_i,\varrho_j<\frac12$. Extend $a$ to a bijection $\pi$ of $\{1,\dots,d\}$ by assigning the sources outside $J$ to the unused columns in any order. Let $s_j\in\{\pm1\}$ be the sign of $\langle q_j,v_{\pi(j)}\rangle$ ($s_j=1$ if it vanishes), and let $P=\sum_js_je_{\pi(j)}e_j^\top$, a signed permutation matrix, so that $(VP)e_j=s_jv_{\pi(j)}$ and $(UP)e_j=s_ju_{\pi(j)}$. For $j\in J$, with $x=\langle q_j,v_{\pi(j)}\rangle$,
\begin{equation*}
\|s_jv_{\pi(j)}-q_j\|^2=2-2|x|\le2(1-x^2)=2\varrho_j,
\end{equation*}
since $1-|x|\le1-x^2$ for $|x|\le1$; for $j\notin J$, $\|s_jv_{\pi(j)}-q_j\|^2\le2\le4\varrho_j$. Hence $\|V-QP^\top\|_F^2=\|VP-Q\|_F^2\le4\sum_j\varrho_j$. The matrix $QP^\top$ is orthogonal, so the choice of $U$ gives $\|U-V\|_F\le\|QP^\top-V\|_F$, and by the triangle inequality $\|U-QP^\top\|_F\le2\|V-QP^\top\|_F$. Finally, for $x=\langle q_j,s_ju_{\pi(j)}\rangle\in[-1,1]$, $\rho_j(U)\le1-x^2\le2-2x=\|s_ju_{\pi(j)}-q_j\|^2$. Summing over $j$,
\begin{equation*}
\sum_j\rho_j(U)\le\|UP-Q\|_F^2=\|U-QP^\top\|_F^2\le4\,\|V-QP^\top\|_F^2\le16\sum_j\varrho_j .\qedhere
\end{equation*}
\end{proof}

\emph{Proof of Corollary~\ref{cor:minimax}(iii).} Let an estimator return nonzero directions $\hat b_1,\dots,\hat b_d$ (a zero vector may be replaced by any unit vector, which does not increase the closest-direction loss); it may use the signatures and be randomised. Restrict the supremum over invertible mixings to $A=Q\in\Od$. Then $z_a=Q^\top\hat b_a$, $z_{a,j}^2/\|z_a\|^2=\langle q_j,v_a\rangle^2$ with $v_a=\hat b_a/\|\hat b_a\|$, and the closest-direction loss is $\lA^2=\frac1d\sum_j\varrho_j$ in the notation of Lemma~\ref{lem:matched}. Let $U$ be a nearest orthogonal matrix to $V=[v_1,\dots,v_d]$, chosen as a measurable function of $V$ (a measurable choice exists, as $\Od$ is compact and the distance is continuous). Then $U$ is an estimator in $\Od$ that may use the signatures, and Theorem~\ref{thm:rate-full}(a) applies to it. By \eqref{eq:lb3} and the lemma,
\begin{equation*}
\sup_{Q\in\Od}\E\,\lA^2(\hat b)\ \ge\ \frac1{16}\sup_{Q\in\Od}\E\,\lA^2(U)\ \ge\ \frac c{16d}\sum_{j=1}^d\frac{E_j}{n+E_j}\ \ge\ \frac c{32}\cdot\frac{\Ebar}n,
\end{equation*}
the last step because $\Emax\le d/S_{\min}\le n$ under $\gamma\le c$. The supremum over all invertible $A$ is at least the supremum over $\Od$. The upper bound is Theorem~\ref{thm:unknown-full}(b) with $s_0=S_{\min}$, whose condition is $\gamma\le c$ for a smaller $c$; with the signatures, Theorem~\ref{thm:upper-full}(b) gives it as well, since the closest-direction loss is at most the labelled loss $\frac1d\sum_j\rho_j$. For the same reason the lower bound of (iii) implies the lower bound of (i) again, with the constant $c/32$ instead of $c/2$. $\square$

\paragraph{Cost and practice.} For well-spread signatures and fixed $K$, with $s_0\asymp S_{\min}\asymp d^{-2/(K-1)}$, $|\mathcal N|=O(d^{K/(K-1)})$ and the total cost is $O(d^{3+K/(K-1)})$. A grid of the hyperplane itself, with mesh $2h/\sqrt{K-1}$ in an orthonormal basis of $\{\sum_kx(k)=0\}$ around $\mathbf 1$ and projected onto $\Omega_B$, is also an $h$-net of $\Omega_B$, with at most $(32B^4\sqrt K/\sqrt{s_0}+2)^{K-1}=O(d)$ points for fixed $K$, and the proof is unchanged. For separated signatures with $K\le a\log d$, $|\mathcal N|\le(16B^4/\sqrt{s_0}+2)^{a\log d}=d^{\,a\log(16B^4/\sqrt{s_0}+2)}$, a polynomial whose degree grows with $a$, $B$ and $1/s_0$; for the codes of Section~\ref{sec:law} ($a\ge32$) it cannot be computed in practice, and the grid is a device for the proof rather than a practical algorithm. The proof uses two facts only: the acceptance rule finds a probe near every signature, and every accepted probe is near some signature. Any procedure with these two properties can replace the grid. Appendix~\ref{app:exp} reports a coarse grid followed by alternating refinement, and Appendix~\ref{app:upper} (after Lemma~\ref{lem:probe}) a Jacobi pilot.

\section{Proof of Theorem~\ref{thm:klaw}, Corollary~\ref{cor:necessary} and Proposition~\ref{prop:random}}
\label{app:klaw}

By \eqref{eq:S-euclid} it suffices to consider the Euclidean energy $\tilde E_j=K\sum_{i\neq j}\|\sig_i-\sig_j\|^{-2}$ in the $(K-1)$-dimensional normalised signature space, with $r_{\min}=c_1d^{-1/(K-1)}$. \emph{Upper bound.} The balls of radius $r_{\min}/2$ around the signatures are disjoint, so the number of signatures at distance in $[mr_{\min},(m+1)r_{\min})$ from $\sig_j$ is at most $Cm^{K-2}$ for $m\ge1$ (the volume of the shell divided by that of a ball of radius $r_{\min}/2$), and
\begin{equation}
\tilde E_j\le\frac{CK}{r_{\min}^2}\sum_{m=1}^{M}m^{K-4},\qquad M\le\frac{\mathrm{diam}}{r_{\min}}\asymp d^{1/(K-1)} .
\end{equation}
For $K=2$ the series converges, $\sum m^{-2}=\pi^2/6$, and $\tilde E_j\le C'/r_{\min}^2\asymp d^2$; for $K=3$ it is $\sum m^{-1}\le\log M+1\asymp\log d$, giving $d\log d$; for $K\ge4$ it is $\asymp M^{K-3}$, giving $M^{K-3}/r_{\min}^2\asymp d^{(K-3)/(K-1)}d^{2/(K-1)}=d$. \emph{Lower bound.} Since $\Omega$ is a convex body containing $\sig_j$, the shell $\{x\in\Omega:\ mr_d\le\|x-\sig_j\|<(m+1)r_d\}$ has volume at least $c\,m^{K-2}r_d^{K-1}$ for $1\le m\le cM$, with $c$ depending on $\Omega$ (a cone with apex $\sig_j$ of fixed aperture lies in $\Omega$ up to distance $\asymp\mathrm{inradius}(\Omega)$). Every point of this shell is within $C_1r_d$ of a signature, so the signatures at distance in $[(m-C_1)r_d,(m+1+C_1)r_d)$ from $\sig_j$ cover the shell with balls of radius $C_1r_d$ and number at least $c'm^{K-2}$. Summing over $m$ with the shells grouped so that the counts do not overlap, the same sums bound $\tilde E_j$ from below, uniformly in $j$. For arbitrary configurations, all of which lie in the compact convex set $\Omega_B=\{\sum_k\sig(k)=K\}\cap[1/B,B]^K$ of positive $(K-1)$-dimensional volume, the lower bound $\Ebar\gtrsim d$ follows from $S_{ij}\le B^2\mathrm{diam}^2$ termwise, and the lower bounds for $K\in\{2,3\}$ are the minimal-energy asymptotics on $\Omega_B$, $N^{1+\beta/D}$ for $\beta>D$ and $N^2\log N$ for $\beta=D$ \citep{kuijlaars1998asymptotics,hardin2005minimal,borodachov2019discrete}, applied to $\beta=2$, $D=K-1$, $N=d$ and divided by $N$.

\emph{Proof of Corollary~\ref{cor:necessary}.} Put $f(N)=N^2$, $N\log(N+1)$ and $N$ for $K=2$, $3$ and $K\ge4$; since $d\log d\le d\log(d+1)$, it suffices to prove the bound with this $f$. Every configuration lies in $\Omega_B$, and the minimal-energy bounds just used give $c_*>0$, depending only on $(K,B)$, such that any $N\ge2$ points of $\Omega_B$ satisfy $\sum_{a\neq b}\|\sig_a-\sig_b\|^{-2}\ge c_*Nf(N)$. For $K\ge4$ this holds because all $N(N-1)$ distances are at most $\mathrm{diam}\,\Omega_B$. For $K=2$ the $N-k$ distances between points $k$ apart in the sorted order sum to at most $k\,\mathrm{diam}\,\Omega_B$, so Jensen's inequality for $k=1$ already gives order $N^3$. For $K=3$ it is the case $\beta=D$ of the asymptotics, together with positivity of the minimal energy for each fixed $N$. The largest energy within a subset is at least the average, so by \eqref{eq:S-euclid} any $N\ge2$ of the signatures contain one, $j$, with $E_j\ge(K/B^2)c_*f(N)=:c_3f(N)$. Let $m=\max(2,\lceil d/2\rceil)\le d$ and $\mathcal A=\{j:E_j<c_3f(m)\}$. If $|\mathcal A|\ge m$, then any $m$ members of $\mathcal A$ contain one with $E_j\ge c_3f(m)$, which is impossible. Hence at least $d-m+1\ge d/2$ sources have $E_j\ge c_3f(m)\ge c_3f(d)/4$. Since $x\mapsto x/(n+x)$ is increasing, \eqref{eq:lb3} gives $\sup_Q\E\lA^2\ge\frac c2\,x/(n+x)$ with $x=c_3f(d)/4$, and $x/(n+x)\ge\frac12\min(1,c_3/4)\min(1,f(d)/n)$. $\square$

\begin{proposition}[Random signatures]\label{prop:random}
Let the normalised signatures be i.i.d.\ from a density bounded above and below on a convex body in the hyperplane. For every $\epsilon>0$ there are constants such that, with probability at least $1-\epsilon$,
\begin{align*}
\Ebar&\asymp d^3,\ d\log d,\ d \quad (K=2,\ 3,\ \ge4),\\
\Emax&\asymp d^4,\ d^2,\ d^{4/3},\ d \quad (K=2,\ 3,\ 4,\ \ge5).
\end{align*}
\end{proposition}
\emph{Random signatures: proof of Proposition~\ref{prop:random}.} Write $D=K-1$. By \eqref{eq:S-euclid} it suffices to prove the statements for the Euclidean energies, now without the fixed factor $K$: $\tilde E_j=\sum_{i\neq j}\|\sig_i-\sig_j\|^{-2}$, $\tilde{\bar E}=\frac1d\sum_j\tilde E_j$ and $\tilde E_{\max}$. Let the signatures be i.i.d.\ with density $g$ on a convex body $\Omega$ of the hyperplane, $0<g_{\min}\le g\le g_{\max}$. For $x\in\Omega$ and $t\le\mathrm{diam}\,\Omega$, convexity gives $c\,t^D\le\mathrm{vol}(B(x,t)\cap\Omega)\le\omega_Dt^D$, so $\pi(x,t)=\Prob(\|\sig-x\|\le t)\in[c_\pi t^D,C_\pi t^D]$ with $c_\pi=cg_{\min}$, $C_\pi=\omega_Dg_{\max}$. Constants $C,C_\epsilon$ below depend on $(D,g_{\min},g_{\max},\Omega)$ and on $\epsilon$, and change from line to line. Let $N_j(t)$ be the number of signatures $i\neq j$ within distance $t$ of $\sig_j$, and $N_{\rm P}(t)$ the number of unordered pairs within distance $t$. The layer-cake formula gives
\begin{equation}\label{eq:layercake}
\tilde E_j=\int_0^\infty2t^{-3}N_j(t)\,dt,\qquad \sum_{i<j}\|\sig_i-\sig_j\|^{-2}=\int_0^\infty2t^{-3}N_{\rm P}(t)\,dt .
\end{equation}

\emph{Step 1: bulk of every centre.} Conditionally on $\sig_j$, $N_j(t)\sim\mathrm{Bin}(d-1,\pi(\sig_j,t))$. Let $t_1=(A\log d/d)^{1/D}$ with $A$ large. For $t\ge t_1$, $\E[N_j(t)\mid\sig_j]\ge c_\pi dt^D/2\ge(c_\pi A/2)\log d$, and the Chernoff bound gives $\Prob(N_j(t)\ge2\E[N_j(t)\mid\sig_j])\le d^{-c_\pi A/6}$. On the dyadic grid $t_m=2^mt_1$, $m\le\log_2d$, a union bound over $j$ and $m$ shows that with probability at least $1-d^{-2}$ (for $A$ large), for all $j$ and all $t\in[t_1,\mathrm{diam}\,\Omega]$, $N_j(t)\le N_j(t_{m+1})\le2^{D+1}C_\pi dt^D$. On this event the part of \eqref{eq:layercake} above $t_1$ is at most $2^{D+2}C_\pi d\int_{t_1}^{\mathrm{diam}}t^{D-3}dt+(d-1)(\mathrm{diam}\,\Omega)^{-2}$ (above the diameter, $N_j(t)=d-1$), which is at most $Cd$ for $D\ge3$, $Cd\log d$ for $D=2$ and $Cd/t_1\le Cd^2/\log d$ for $D=1$, uniformly in $j$; and every centre has at most $M_1=2^{D+1}C_\pi A\log d$ neighbours below $t_1$.

\emph{Step 2: near neighbours of every centre.} Let $r_{(k)}(j)$ be the distance from $\sig_j$ to its $k$-th nearest neighbour. Since $N_j(t)\ge k$ implies $N_j(t)(N_j(t)-1)\cdots(N_j(t)-k+1)\ge k!$, and the falling factorial moments of a binomial are $(d-1)^{(k)}\pi^k\le(C_\pi dt^D)^k$,
\begin{equation*}
\Prob\big(\exists j:\ r_{(k)}(j)\le t\big)\le d\,(C_\pi dt^D)^k/k! .
\end{equation*}
Choose $t_k$ by $d(C_\pi dt_k^D)^k/k!=\epsilon2^{-k-2}$. Then $\sum_k\Prob(\exists j: r_{(k)}(j)\le t_k)\le\epsilon/4$, and $(k!)^{1/k}\ge k/e$ gives $t_k^{-2}\le(2eC_\pi d/k)^{2/D}(4d/\epsilon)^{2/(kD)}$. On the complementary event, for every $j$ the part of \eqref{eq:layercake} below $t_1$ is at most
\begin{equation*}
\sum_{k\le M_1}t_k^{-2}\le C_\epsilon\Big(d^{4/D}+d^{3/D}+d^{8/(3D)}\sum_{3\le k\le M_1}k^{-2/D}\Big)\le C_\epsilon\big(d^{4/D}+d^{8/(3D)}\log d\big)\le C_\epsilon d^{4/D}.
\end{equation*}
With Step 1, with probability at least $1-\epsilon/2$,
\begin{equation}\label{eq:Emax-upper}
\tilde E_{\max}\le C_\epsilon\big(d^{4/D}+d\,[1,\ \log d,\ d/\log d]_{D\ge3,\,D=2,\,D=1}\big)\le C_\epsilon\,(d^4,\ d^2,\ d^{4/3},\ d,\ d)\quad(D=1,2,3,4,\ge5).
\end{equation}

\emph{Step 3: the average.} By \eqref{eq:layercake}, $\tilde{\bar E}=\frac2d\int_0^\infty2t^{-3}N_{\rm P}(t)\,dt$. Here $\E N_{\rm P}(t)=\binom d2\E\pi(\sig,t)\in[c_\pi d^2t^D/4,\ C_\pi d^2t^D/2]$ for $t\le\mathrm{diam}\,\Omega$, and, since indicators of disjoint pairs are independent and at most $d^3$ ordered pairs of pairs share a point, $\Var N_{\rm P}(t)\le\E N_{\rm P}(t)+d^3C_\pi^2t^{2D}\le\E N_{\rm P}(t)\,(1+Cdt^D)$. Let $t_2=(M/(c_\pi d^2))^{1/D}$ with $M=64/\epsilon$. On the dyadic grid $t_m=2^mt_2$, Chebyshev's inequality gives $\Prob(N_{\rm P}(t_m)\ge2\E N_{\rm P}(t_m))\le(1+Cdt_m^D)/\E N_{\rm P}(t_m)\le4\cdot2^{-mD}/M+C/d$; summed over the grid this is at most $8/M+C\log d/d\le\epsilon/4$ for $d\ge d_0(\epsilon)$. Hence, with probability at least $1-\epsilon/4$, $N_{\rm P}(t)\le2^{D+1}C_\pi d^2t^D$ for all $t\ge t_2$, and the part of the pair energy above $t_2$ is at most $Cd^2\int_{t_2}^{\mathrm{diam}}t^{D-3}dt+\binom d2(\mathrm{diam}\,\Omega)^{-2}\le Cd^2\,[1,\ \log d,\ 1/t_2]$, i.e.\ at most $C_\epsilon(d^2,\ d^2\log d,\ d^4)$ for $D\ge3$, $D=2$, $D=1$. Below $t_2$ there are at most $N_{\rm P}(t_2)\le2\E N_{\rm P}(t_2)\le M_2:=(C_\pi/c_\pi)M$ pairs. For the $k$-th smallest pair distance $r_{[k]}$ and fixed $k\le M_2$, $\Prob(r_{[k]}\le t)\le\E[N_{\rm P}(t)^{(k)}]/k!$, and the falling factorial moment is a sum over ordered $k$-tuples of distinct pairs. A $k$-tuple spans a graph with $v$ vertices and $c$ components, has probability at most $(C_\pi t^D)^{v-c}$ (place the vertices along a spanning forest), and there are at most $d^v(2k)^{2k}$ tuples with given $(v,c)$. The disjoint tuples ($v=2k$, $c=k$) contribute at most $(C_\pi d^2t^D)^k$; every other tuple has $v-c-k\le0$ and $2c-v\le-1$, so at a scale $t$ with $C_\pi d^2t^D=\kappa$ its contribution is at most $(2k)^{2k}\max(1,\kappa)^kd^{-1}$. Choosing $t_k$ by $2(C_\pi d^2t_k^D)^k/k!=\epsilon2^{-k-2}$ (so that $\kappa_k=C_\pi d^2t_k^D$ depends only on $k$ and $\epsilon$), $\sum_{k\le M_2}\Prob(r_{[k]}\le t_k)\le\epsilon/4$ for $d\ge d_0(\epsilon)$, and on the complementary event the pair energy below $t_2$ is at most $\sum_{k\le M_2}t_k^{-2}\le C_\epsilon d^{4/D}$. Dividing by $d/2$, with probability at least $1-\epsilon$,
\begin{equation}\label{eq:Emean-upper}
\tilde{\bar E}\le C_\epsilon\big(d^{4/D-1}+[d,\ d\log d,\ d^3]_{D\ge3,\,D=2,\,D=1}\big)\le C_\epsilon\,(d^3,\ d\log d,\ d)\quad(D=1,\ 2,\ \ge3).
\end{equation}

\emph{Lower bounds.} The minimal-energy asymptotics of Theorem~\ref{thm:klaw} hold for every configuration, so $\tilde{\bar E}\ge c\,(d^2,\ d\log d,\ d)$ for $D=1,2,\ge3$ surely, and $\tilde E_{\max}\ge\tilde{\bar E}$. For the closest pair, with $t=(M'/(c_\pi d^2))^{1/D}$ and $M'=64/\epsilon$, Chebyshev gives $\Prob(N_{\rm P}(t)=0)\le\Var N_{\rm P}(t)/(\E N_{\rm P}(t))^2\le4(1+CM'/d)/M'\le\epsilon/4$ for $d\ge d_0(\epsilon)$. Hence, with probability at least $1-\epsilon/4$, some pair is within distance $t$, its two members have energy at least $t^{-2}\ge c_\epsilon d^{4/D}$, and $\tilde{\bar E}\ge\frac2dt^{-2}\ge c_\epsilon d^{4/D-1}$. For $D=1$ this gives $d^3$ and $d^4$; for $D=2,3$ it gives $\tilde E_{\max}\ge c_\epsilon d^2,\ c_\epsilon d^{4/3}$; for $D\ge4$, $\tilde E_{\max}\ge\tilde{\bar E}\ge cd$. Together with \eqref{eq:Emax-upper}--\eqref{eq:Emean-upper} this proves the proposition: the exceptional events have total probability at most $\epsilon+d^{-2}$, and replacing $\epsilon$ by $\epsilon/2$ gives the statement. $\square$

\paragraph{Effective dimension at a scale.} The identity $\sum_{i\neq j}\|\sig_i-\sig_j\|^{-2}=\int_0^\infty2t^{-3}N_j(t)\,dt$ of \eqref{eq:layercake} holds for every configuration, with $N_j(t)$ the number of other signatures within distance $t$ of $\sig_j$. The energy therefore depends on the configuration only through the growth of $N_j$ between the nearest-neighbour distance and the diameter, and an effective dimension can be read at each scale as the local growth exponent of $N_j$. For instance, signatures within distance $\tau$ of a $D'$-dimensional affine subspace and well spread there at scales above $\tau$ have $N_j(t)\asymp d\,(t/\mathrm{diam})^{D'}$ for $\max(\tau,r_{\rm nn})\lesssim t\le\mathrm{diam}$, $r_{\rm nn}$ the nearest-neighbour distance; these scales contribute to the energy as in dimension $D'$, and Remark~\ref{rem:dim} applies approximately when $\tau$ is below the nearest-neighbour distance.

\section{Proof of Theorem~\ref{thm:pencil}}
\label{app:pencil}

Theorem~\ref{thm:pencil} states that the means of the limit laws average over $j$ to at least $(d-1)^3/(dB^4)$. We prove the sharper bound
\begin{equation}\label{eq:pencil-lb}
\frac1d\sum_j\sum_{i\neq j}\frac{v_{ij}(\alpha)}{(\lambda_i-\lambda_j)^2}\ \ge\ \frac{K\|\alpha\|^2}{B^2}\cdot\frac{(d-1)^3}{d\,L_\alpha^2},
\end{equation}
where $L_\alpha\le\|\alpha\|\,\mathrm{diam}\{\sig_j\}$ is the length of the range of the $\lambda_j$; since $\mathrm{diam}^2\{\sig_j\}\le KB^2$, the right-hand side is at least $(d-1)^3/(dB^4)$.

Write $\hat M_\alpha=M_\alpha+N$ with $N=\sum_k\alpha_k(\hat\Sigma_k-\Sigma_k)$. In the basis of $Q$, $N'=Q^\top NQ$ has entries that are uncorrelated across pairs, with $\Var(N'_{ij})=\sum_k\alpha_k^2\Var((\hat\Sigma_k)_{ij})=\sum_k\alpha_k^2\sig_i(k)\sig_j(k)/n_k=v_{ij}(\alpha)/n$. With distinct eigenvalues, first-order eigenvector perturbation gives $\hat q_j=q_j+\sum_{i\neq j}\frac{N'_{ij}}{\lambda_j-\lambda_i}q_i+O(\|N\|^2)$, and $\sqrt n\,N'_{ij}$ are asymptotically independent $N(0,v_{ij}(\alpha))$ by the central limit theorem, hence the limit law of $n\rho_j$ \citep{anderson1963asymptotic}. For \eqref{eq:pencil-lb}, $v_{ij}(\alpha)\ge K\|\alpha\|^2/B^2$. Let $g_1,\dots,g_{d-1}$ be the consecutive spacings of the sorted $\lambda$'s, $\sum_mg_m\le L_\alpha$. The nearest-neighbour distance of a point is the smaller of its two adjacent spacings, so $\sum_j\sum_{i\neq j}(\lambda_i-\lambda_j)^{-2}\ge\sum_j\max(g_{\rm left}(j)^{-2},g_{\rm right}(j)^{-2})\ge\frac12\sum_j\big(g_{\rm left}(j)^{-2}+g_{\rm right}(j)^{-2}\big)=\sum_mg_m^{-2}$, and by the power-mean inequality $\frac1{d-1}\sum_mg_m^{-2}\ge\big(\frac1{d-1}\sum_mg_m\big)^{-2}\ge(d-1)^2/L_\alpha^2$. Finally $\mathrm{diam}^2\{\sig_j\}\le\sum_k(B-1/B)^2\le KB^2$. With estimated whitening the noise $N$ is replaced by $\sum_k\alpha_kN^{\rm w}_k$ with $N^{\rm w}_{k,ij}=N_{k,ij}-\frac12(\sig_i(k)+\sig_j(k))\bar N_{ij}$, whose variance is $K\sum_k(\alpha_k-\bar\alpha_{ij})^2\sig_i(k)\sig_j(k)/n$ with $\bar\alpha_{ij}=\frac1{2K}\sum_k\alpha_k(\sig_i(k)+\sig_j(k))$; since $\sum_k(\alpha_k-\bar\alpha_{ij})^2\sig_i\sig_j\ge\|\alpha-\bar\alpha_{ij}\mathbf 1\|^2/B^2\ge\mathrm{dist}(\alpha,\R\mathbf 1)^2/B^2$, the same argument gives \eqref{eq:pencil-lb} with $\|\alpha\|$ replaced by $\mathrm{dist}(\alpha,\R\mathbf 1)$. $\square$

\section{Joint diagonalization: proof of Proposition~\ref{prop:ajd}}
\label{app:ajd}

\begin{proposition}[Joint diagonalization]\label{prop:ajd}
Let $\hat U$ be a global minimiser of $\hat f$ over $\Od$, let the sources be independent with finite fourth moments, and let $S_{ij}>0$ for all $i\neq j$.
\begin{enumerate}[leftmargin=*,itemsep=1pt,topsep=2pt]
\item[(i)] (First-order variance, distribution-free.) As $n\to\infty$ at fixed $d$, $n\,\rho_j(\hat U)\Rightarrow\sum_{i\neq j}v_{ij}Z_i^2$, $Z_i$ i.i.d.\ standard normal, with explicit coefficients $v_{ij}$, one formula for known whitening and one for the unmixing rows with estimated whitening (\eqref{eq:frob-var} and \eqref{eq:white-var} below), both satisfying $S_{ij}^{-1}\le v_{ij}\le C_BS_{ij}^{-1}$ with $C_B=2B^4(1+B^6)$.
\item[(ii)] (Efficiency of weighting.) With known whitening, weights $1/(\hat{\sig}_a(k)\hat{\sig}_b(k))$ on the $(a,b)$ entries, with consistent $\hat\sig$, give $v_{ij}=S_{ij}^{-1}$, attaining Theorem~\ref{thm:rate-full}(b).
\end{enumerate}
\end{proposition}

For a pair $(i,j)$ write $\delta=\sig_i-\sig_j\in\R^K$ and $\bar\delta_j=\frac1K\sum_k\sig_j(k)\delta_k$. The coefficients of part (i) are
\begin{equation}\label{eq:frob-var}
v^{\rm F}_{ij}=\frac{K\sum_k\delta_k^2\,\sig_i(k)\sig_j(k)}{\|\delta\|^4},
\end{equation}
\begin{equation}\label{eq:white-var}
v^{\rm w}_{ij}=\frac{K\sum_k(\delta_k-\bar\delta_j)^2\,\sig_i(k)\sig_j(k)}{\|\delta\|^4},
\end{equation}
the second for the unmixing row $\hat b_j$ of source $j$ with the loss $1-z_j^2/\|z\|^2$, $z=A^\top\hat b_j$, of Theorem~\ref{thm:upper}(b). The joint diagonalizer is equivariant, so this loss has the same law for every invertible $A$; for orthogonal $A$ it is the squared sine of the angle between $\hat b_j$ and the $j$-th column of $A$; $\bar\delta_j$ uses the source whose row is estimated, so $v^{\rm w}_{ij}\neq v^{\rm w}_{ji}$ in general. We write $E^{\rm F}_j=\sum_{i\neq j}v^{\rm F}_{ij}$, $E^{\rm w}_j=\sum_{i\neq j}v^{\rm w}_{ij}$, and $\Ebar^{\rm F}$, $\Ebar^{\rm w}$ for their averages over $j$. With environment proportions $\pi_k$ and the $\pi$-weighted criterion and pooled covariance, the same derivation gives the variance of $\sum_k\beta_kN_{k,ab}$ with $\beta_k=\omega_k\delta_k/\Omega_{ab}-\pi_kc_{ab}$, $\Omega_{ab}=\sum_k\omega_k\delta_k^2$, $c_{ab}=\frac12\big(\sum_k\omega_k\delta_k(\sig_a(k)+\sig_b(k))/\Omega_{ab}+1\big)$, for weights $\omega_k=\pi_k$ (Frobenius) or $\omega_k=\pi_k/(\sig_a(k)\sig_b(k))$ (Fisher); for the non-orthogonal problem the efficient coefficient is the Cram\'er--Rao bound $v^{\rm PC}_{ij}=\omega_{ji}/(\omega_{ij}\omega_{ji}-1)$ with $\omega_{ij}=\sum_k\pi_k\sig_i(k)/\sig_j(k)$, which satisfies $v^{\rm PC}_{ij}\ge S_{ij}^{-1}$ because $(\omega_{ji}-1)^2\ge0$.

\paragraph{(i) First-order variance, known whitening.} At fixed $d$ the global minimiser is consistent: the population criterion has the true $Q$, up to signs and permutations, as its unique minimiser when all $S_{ij}>0$, and $\hat f\to f$ uniformly on $\Od$. So we may write $V=Q^\top\hat U=\exp(\Xi)$ with $\Xi$ antisymmetric and small. In the basis of $Q$, $\Sh_k$ becomes $\Lambda_k+N_k$ with $N_k=Q^\top(\Sh_k-\Sigma_k)Q$, and for $a\neq b$,
\begin{equation*}
(V^\top(\Lambda_k+N_k)V)_{ab}=N_{k,ab}+\Xi_{ab}\big(\sig_a(k)-\sig_b(k)\big)+O(\|\Xi\|^2+\|\Xi\|\|N_k\|).
\end{equation*}
The criterion decouples at first order into independent pair problems $\min_{\Xi_{ab}}\sum_k(N_{k,ab}+\Xi_{ab}\delta_k)^2$ with $\delta=\sig_a-\sig_b$, solved by $\Xi_{ab}=-\sum_k\delta_kN_{k,ab}/\|\delta\|^2$, i.e.\ $\Xi_{ba}=\sum_k\delta_kN_{k,ab}/\|\delta\|^2$. For independent centred sources, $\E[s_as_bs_{a'}s_{b'}]$ vanishes unless $\{a,b\}=\{a',b'\}$ and $\E[s_a^2s_b^2]=\sig_a\sig_b$, so the $N_{k,ab}$ are uncorrelated across pairs and environments with $\Var(N_{k,ab})=\sig_a(k)\sig_b(k)/n_k$, whatever the law of the sources. The central limit theorem for the $N_{k,ab}$ and $\rho_j=\sum_{i\neq j}\Xi_{ji}^2+O(\|\Xi\|^3)$ give the limit law, whose mean is \eqref{eq:frob-var}. The lower bound $v^{\rm F}_{ij}\ge S_{ij}^{-1}$ is Cauchy--Schwarz, $\|\delta\|^4\le\sum_k\delta_k^2\sig_i\sig_j\cdot\sum_k\delta_k^2/(\sig_i\sig_j)$, with equality iff $\sig_i(k)\sig_j(k)$ is constant over the $k$ with $\sig_i(k)\neq\sig_j(k)$; the upper bound $v^{\rm F}_{ij}\le B^4S_{ij}^{-1}$ uses $\sig_i\sig_j\in[B^{-2},B^2]$.

\paragraph{(i) Estimated whitening.} By equivariance we take $A=Q\in\Od$. Let $\bar N=\frac1K\sum_kN_k$ be the noise of the pooled covariance in the basis of $Q$; the pooled population covariance is the identity by the normalisation. The estimated whitening matrix is $(I+\bar N)^{-1/2}=I-\frac12\bar N+O(\|\bar N\|^2)$, and the whitened environment covariances are
\begin{equation*}
(I-\tfrac12\bar N)(\Lambda_k+N_k)(I-\tfrac12\bar N)=\Lambda_k+N_k-\tfrac12(\bar N\Lambda_k+\Lambda_k\bar N)+O(\|N\|^2),
\end{equation*}
so the pair problem is solved by $\Xi_{ba}=\sum_k\delta_kN^{\rm w}_{k,ab}/\|\delta\|^2$ with $N^{\rm w}_{k,ab}=N_{k,ab}-\frac12(\sig_a(k)+\sig_b(k))\bar N_{ab}$, as in the complete formula of \citet[Thm.~2]{miettinen2016separation}. The $a$-th unmixing row is the $a$-th column of $(I-\frac12\bar N)V$, whose coefficient along the $b$-th column of $A$ is $\Xi_{ba}-\frac12\bar N_{ab}+O(\|N\|^2)$. Since $\sum_k\delta_k\,\frac12(\sig_a(k)+\sig_b(k))=\frac12(\|\sig_a\|^2-\|\sig_b\|^2)$ and $\bar N=\frac1K\sum_kN_k$, a short computation gives
\begin{equation*}
\Xi_{ba}-\tfrac12\bar N_{ab}=\frac{1}{\|\delta\|^2}\sum_k\Big(\delta_k-\frac1K\sum_{k'}\sig_a(k')\delta_{k'}\Big)N_{k,ab},
\end{equation*}
i.e.\ the weights $\delta_k$ are replaced by $\delta_k-\bar\delta_a$ with $\bar\delta_a$ the $\sig_a$-weighted mean of $\delta$. Its variance is \eqref{eq:white-var} (with $j=a$, $i=b$). The loss of the whitened-space rotation $V$ against $Q$, and of the estimated mixing columns $(I+\frac12\bar N)V$ against $A$, have the same form with $\bar\delta_a$ replaced by the $\frac12(\sig_a+\sig_b)$- and $\sig_b$-weighted means. The two-sided comparison: $v^{\rm w}_{ij}\ge S_{ij}^{-1}$ by Cauchy--Schwarz, $\|\delta\|^4=\big(\sum_k(\delta_k-\bar\delta_j)\delta_k\big)^2\le\sum_k(\delta_k-\bar\delta_j)^2\sig_i\sig_j\cdot\sum_k\delta_k^2/(\sig_i\sig_j)$, where $\sum_k\delta_k=0$ by the normalisation; and $\sum_k(\delta_k-\bar\delta_j)^2\sig_i\sig_j\le2\sum_k\delta_k^2\sig_i\sig_j+2\bar\delta_j^2\sum_k\sig_i\sig_j\le(2+2B^6)\sum_k\delta_k^2\sig_i\sig_j$, using $\bar\delta_j^2\le B^2\|\delta\|^2/K$, $\sum_k\sig_i\sig_j\le KB^2$ and $\sum_k\delta_k^2\sig_i\sig_j\ge\|\delta\|^2/B^2$; hence $v^{\rm w}_{ij}\le(2+2B^6)v^{\rm F}_{ij}\le C_BS_{ij}^{-1}$ with $C_B=2B^4(1+B^6)$. The constants can differ a lot: for $K=2$, $\sig_i=(1.90,0.10)$ and $\sig_j=(1.89,0.11)$ the coefficients are $v^{\rm F}_{ij}=18010$, $v^{\rm w}_{ij}=414$ for the row of source $j$ and $v^{\rm w}_{ji}=378$ for that of source $i$, all above the information bound $S_{ij}^{-1}=219$; simulations with $n=2\cdot10^5$ give $n\,\E\rho=421\pm11$ and $385\pm10$. For the random signatures of Section~\ref{sec:experiments} the two formulas differ by a few percent.

\paragraph{(ii)} With weights $1/(\sig_a(k)\sig_b(k))$ the pair problem becomes $\min\sum_k(N_{k,ab}+\Xi_{ab}\delta_k)^2/(\sig_a\sig_b)$, whose solution has variance $K/(n\sum_k\delta_k^2/(\sig_a\sig_b))=1/(nS_{ab})$; consistently estimated weights change the solution by $o_P(n^{-1/2})$, so $n\rho_j$ has the same limit law as with the true weights, $\sum_{i\neq j}S_{ij}^{-1}Z_i^2$ (Slutsky's lemma).

\section{Below the threshold}
\label{app:open}

\begin{proposition}[Accuracy at the nearest-neighbour scale]\label{prop:barrier}
Let $c$ be the constant of Theorem~\ref{thm:rate}. If an estimator satisfies $\sup_Q\E\,\rho_j(\hat Q)\le\eta\,S_{\rm nn}(j)$ with $\eta S_{\rm nn}(j)\le c/2$, then $n\ge c\,E_j/(2\eta S_{\rm nn}(j))$. If $\sup_Q\E\,\lM^2(\hat Q)\le\eta\,S_{\min}\le c/2$, then $n\ge c\,\Emax/(2\eta S_{\min})$.
\end{proposition}
Indeed, \eqref{eq:lb2} gives $cE_j/(n+E_j)\le\eta S_{\rm nn}(j)$, so $n\ge E_j(c/(\eta S_{\rm nn}(j))-1)\ge cE_j/(2\eta S_{\rm nn}(j))$, and \eqref{eq:lb4} gives the second claim. $\square$

\paragraph{The threshold of Theorem~\ref{thm:upper}.} Steps~1--4 of Appendix~\ref{app:upper} need $\nu/\lambda_*\le c$, with $\nu\asymp K^2(d+\log n)/n$ and $\lambda_*\asymp KS_{\rm nn}(j)$. When $E_j\asymp d$, as for well-spread signatures with $K\ge4$ and for separated ones, this says that the first-order error $E_j/n$ is below a fraction of $S_{\rm nn}(j)/K$, and by Proposition~\ref{prop:barrier} no estimator is that accurate unless $n\gtrsim KE_j/S_{\rm nn}(j)$. This suggests, but does not prove, that arguments comparing the noise with the nearest-neighbour gap in operator norm cannot go further. Below the threshold the worst-case error of a source is $\gg S_{\rm nn}(j)/K$ (Proposition~\ref{prop:barrier}) but spread over $d-1$ directions, each pair contributing about $1/(nS_{ij})$; we expect that a proof there must average these contributions like independent noise, as a local law for the joint-diagonalization resolvent would.

\paragraph{The factor $K$ in $\gamma_j$.} Part of it comes from the operator-norm bounds of Step~1, which use Cauchy--Schwarz over environments and could be sharpened, but not all of it. By Step~5 the error vector $g=H'^{-1}h$ of the estimator contains $H'^{-1}h^{(2)}$, the product of the two halves' noise, uncorrelated with $H'^{-1}h^{(1)}$ and with second moment of exact order $KdE^{(4)}_j/n^2$ (up to factors of $B$): each environment's halves have $n/(2K)$ observations, $d$ terms add up in each entry, and so do the $K$ environments. For separated signatures $E^{(4)}_j\asymp d$ and $E_j\asymp d$, so this term is $(Kd/n)\,E_j/n$, and $\E\|g\|^2$ stays within a constant of $E_j/n$ only once $n\gtrsim Kd$; for $K\asymp\log d$ this is the $d\log d$ of Section~\ref{sec:law}. A simulation agrees (balanced codes, $d=32$, known whitening, $150$ draws): for $K=8$, $16$, $32$, $n\,\E\lA^2/\Ebar^{\rm qp}$ of the labelled vectors is $\FKSimTwo$ at $n=200d$ and $\FKSimEight$ at $n=800d$, against $1+n\,\E\|H'^{-1}h^{(2)}\|^2/\Ebar^{\rm qp}$ (averaged over sources) $=\FKPredTwo$ and $\FKPredEight$, and it is about the same at equal $Kd/n$ ($\FKEqualA$ for $K=8$ at $n=200d$, $\FKEqualB$ for $K=32$ at $n=800d$). We therefore expect this estimator to need the factor $K$; reaching $n\asymp d$ with $\log d$ separated environments would need another estimator, for instance one that does not multiply the per-environment covariances.

\paragraph{A second-order drift of the quadratic pencil.} A second-order expansion suggests a systematic shift in the gap that can become important below the threshold of Theorem~\ref{thm:upper}. In the basis of $Q$ at the exact probe $\sig_j$, let $i$ be a near source and $F$ a set of far sources, $\lambda_b\gg\lambda_i$ for $b\in F$. Eliminating $F$ by a Schur complement, the expected diagonal entries of the effective matrix on $\{i,j\}$ are, to second order in the noise, $\lambda_a-\sum_{b\in F}\E(P_1)_{ab}^2/\lambda_b$, since $\E P_1=\E P_2=0$. With $\E(P_1)_{ab}^2=\frac Kn\sum_kw_k^2(\delta_a(k)+\delta_b(k))^2\sig_a(k)\sig_b(k)$ and $\delta_j=0$, expanding to first order in $\delta_i$ gives the effective gap
\begin{equation}\label{eq:drift}
\lambda_i-\frac Kn\langle g_j,\delta_i\rangle,\qquad g_j(k)=\sum_{b\in F}\frac{w_k^2\,\sig_b(k)\big(\sig_b(k)^2-\sig_j(k)^2\big)}{\lambda_b}.
\end{equation}
The far sources repel the two levels by different amounts. Moving the probe to $\sig_j+\eta_j$ changes the population gap by $-2\sum_kw_k\delta_i(k)\eta_j(k)$, so $\eta_j(k)=-\frac K{2n}g_j(k)/w_k$ undoes the shift at first order. Since $|\sig_b^2-\sig_j^2|\le2B|\delta_b|$ and $\lambda_b\ge B^{-2}\|\delta_b\|^2$, $|g_j(k)|\le2B^8\sum_b\|\delta_b\|^{-1}$, of order $d$ for well-spread signatures with $K\ge3$, so $\|\eta_j\|=O(K^{3/2}d/n)$. As $|\langle g_j,\delta_i\rangle|\le\sqrt K\max_k|g_j(k)|\,\|\delta_i\|$, the shift is $O(K^{3/2}d\,r_j/n)$ against a gap $\lambda_i\asymp r_j^2$, so the gap can close unless $n\gtrsim K^{3/2}d/r_j$, which is $d^{K/(K-1)}$ for well-spread signatures and fixed $K$. Under $\gamma_j\le c$ the relative shift is $O(r_j)$. The expansion \eqref{eq:drift} is a second-order computation, not a theorem.

Two simulations illustrate this shift. Both use $K=5$, $d=300$, one near source at $\sig_j\pm0.05\,z$ for a unit vector $z$ in the hyperplane, and $d-2$ sources around $\sig_j+Rz$; the effective gap is measured on the Schur complement of the far block. With $\sig_j=\mathbf 1$ ($w_k=1$), $R=0.5$, $n=1000d$ and $\DriftReps$ draws, the population gap is $\DriftGapPop$ for both signs, the predicted gaps are $\DriftPredPlus$ and $\DriftPredMinus$, the measured ones $\DriftGapPlus$ and $\DriftGapMinus$ (standard errors at most $\DriftSE$), with $\E\rho_j=\DriftRhoPlus$ and $\DriftRhoMinus$; at the corrected probe (weights unchanged) they are $\DriftGapPlusC$ and $\DriftGapMinusC$, with $\E\rho_j=\DriftRhoPlusC$ and $\DriftRhoMinusC$. A tilted $\sig_j\propto(1.3,0.8,1.1,0.9,0.9)$ ($R=0.4$, $\DriftTReps$ draws, population gap $\DriftTGapPop$) gives the same picture: at $n=1000d$ the measured gaps are $\DriftTGapPlus$ and $\DriftTGapMinus$ against predicted $\DriftTPredPlus$ and $\DriftTPredMinus$, and the corrected ones $\DriftTGapPlusC$ and $\DriftTGapMinusC$ ($\DriftTResid$ standard errors apart), and at $n=4000d$ the corrected ones are $\DriftTGapPlusCB$ and $\DriftTGapMinusCB$ (standard errors at most $\DriftTSEB$): the correction removes the asymmetry at first order.

\paragraph{Towards the linear regime.} A proof of Conjecture~\ref{conj:linear}, or of its analogue for the quadratic pencil at the corrected probe, would need, we expect, a multiscale replacement of Step~1 that compares the noise with the gap shell by shell, $\{i:\|\sig_i-\sig_j\|\asymp2^{-l}\}$; a shell at distance $r$ holds about $d\,r^{K-1}$ sources and contributes about $d\,r^{K-3}$ to the energy. In simulations below the proved threshold, Jacobi sweeps follow the first-order prediction wherever $\Ebar^{\rm w}/n\lesssim0.1$ (Section~\ref{sec:experiments}, Figure~\ref{fig:larged}), and below $100d$ they are stable under initialisation (Appendix~\ref{app:exp}, runs below $n=100d$).

\section{Further related work}
\label{app:related}
The Gaussian model with a common $Q$ and diagonal $\Lambda_k$ is the common-principal-components model of \citet{flury1984common}, whose maximum-likelihood asymptotics \citep{flury1986asymptotic} give the information $S_{ij}$. Fixed-dimension theory of second-order BSS goes back to \citet{cardoso1994perturbation,belouchrani1997blind,pham2001blind}, with optimal weightings and Cram\'er--Rao bounds \citep{yeredor2000blind,tichavsky2009fast,yeredor2010blind}, sensitivity analyses \citep{afsari2007what,afsari2008sensitivity} and complete variance formulas for SOBI and AMUSE \citep{miettinen2012statistical,miettinen2016separation}; see \citet{nordhausen2022usage}. Pham's algorithm \citep{pham2001joint} computes the Gaussian likelihood estimator and AC-DC \citep{yeredor2002nonorthogonal} and FFDIAG \citep{ziehe2004fast} are non-orthogonal least-squares methods.

\citet{kuleshov2015tensor} bound joint diagonalization by Euclidean signature gaps and remove the smallest gap by random projections, without counting the dimension of the configuration or proving lower bounds; \citet{he2024randomized,he2025randomized} reduce it to a random pencil plus refinement; their guarantees bound the joint-diagonalization residual, and when signatures are close a small residual does not imply a small direction error, which is what Theorem~\ref{thm:pencil} quantifies at first order. Random combinations of slices are also the first step of simultaneous diagonalization for tensors; smoothed and Fourier analyses lower-bound the eigenvalue gaps of such a combination in probability \citep{bhaskara2014smoothed,goyal2014fourier}, while \eqref{eq:pencil-lb} bounds the first-order risk of its eigenvectors for every combination. In growing dimension \citet{chang2018principal} analyse a matrix-based dimension reduction for stationary vector time series; we found no sample-size results for full-rank common principal components. Step~1 of Appendix~\ref{app:upper} bounds the noise relative to $H_j$, so that it must beat the nearest-neighbour gap rather than the spread of the spectrum, in the spirit of relative perturbation bounds for eigenvalues and eigenvectors \citep{ipsen1998relative} and for empirical covariance operators \citep{jirak2018relative}; Davis--Kahan bounds \citep{davis1970rotation,yu2015useful} compare the noise in operator norm with the eigengap. For a single covariance matrix, \citet{wahl2022lower,wahl2023vantrees} proves non-asymptotic lower bounds for principal subspaces of the same form as Theorem~\ref{thm:rate}, a sum over pairs of inverse Fisher informations $\lambda_i\lambda_j/(\lambda_i-\lambda_j)^2$ each truncated at a constant, from a van Trees inequality for orthogonally equivariant experiments and Haar priors; the general inequality of the second paper covers the Haar symmetrisation of Appendix~\ref{app:rate}. Three things differ here. The $K$ covariances share their eigenvectors and the information is averaged over environments, so the quantity is the signature divergence $S_{ij}$ rather than an eigengap of one matrix; the sources have no natural ordering, so the loss matches each source to the closest estimated direction, and a bound for one specified direction does not transfer to it (Appendix~\ref{app:rate} uses the orthogonality of the other columns to control the matching); and ordering the hypercube by $S_{ji}$ removes the truncation, so that \eqref{eq:lb2} holds for every $n$.

The matrix $\sum_k(X_k-\lambda_kI)^2$ of Theorem~\ref{thm:upper} is the quadratic composite operator of numerical linear algebra and mathematical physics: \citet{cerjan2023quadratic} use its smallest singular value, as a function of the probe $\lambda$, to locate approximate joint eigenvectors of non-commuting observables (the quadratic pseudospectrum), and \citet{loring2022computing} take its bottom eigenvectors before a Jacobi joint diagonalization. Both are deterministic; Theorems~\ref{thm:upper} and~\ref{thm:unknown} add the statistics, a sample-split form that removes the bias of the squared noise, and the grid of probes. Landscape and convergence are studied by \citet{usevich2020approximate,troedsson2024joint}.

For the non-Gaussian route see \citet{auddy2025large,ricci2025feature,genc2026learnability} and hardness results \citep{diakonikolas2017statistical,diakonikolas2024sos,damian2024computational}. Identifiability from several environments is qualitative in nonlinear ICA \citep{hyvarinen2016unsupervised,hyvarinen2019nonlinear,khemakhem2020variational,hyvarinen2023nonlinear}, also when the segmentation into environments is itself learned \citep{halva2020hidden}, and in causal representation learning, where environments are interventions on a latent causal model, linearly mixed \citep{squires2023linear,jin2024learning} or not \citep{buchholz2023learning,vonkugelgen2023nonparametric}; our results give sample sizes for independent latents and a common linear mixing. \citet{jiang2026finite} bounds nonlinear ICA by an environment-level diversity, the smallest KL divergence between two environments, which does not resolve which pairs of sources are hard. \citet{acarturk2024sample} give finite-sample upper bounds for interventional causal representation learning, without lower bounds, and \citet{lee2026beyond} show that $K\asymp\log d$ unknown multi-node interventions forming a separating system identify a linear causal model; our codewords play an analogous role, under different assumptions, for the sample size of second-order separation. \citet{rathod2026sample} observe eigenvalue crowding in a pencil-based causal discovery method, which Theorem~\ref{thm:pencil} may explain. In multi-task linear representation learning the benefit of many tasks is governed by a task-diversity condition number \citep{tripuraneni2021provable}; here the analogous quantity is the dimension of the signature configuration, not the number of environments.

\section{Experimental details}
\label{app:exp}

\begin{figure}[tb]
\centering
\includegraphics[width=\textwidth]{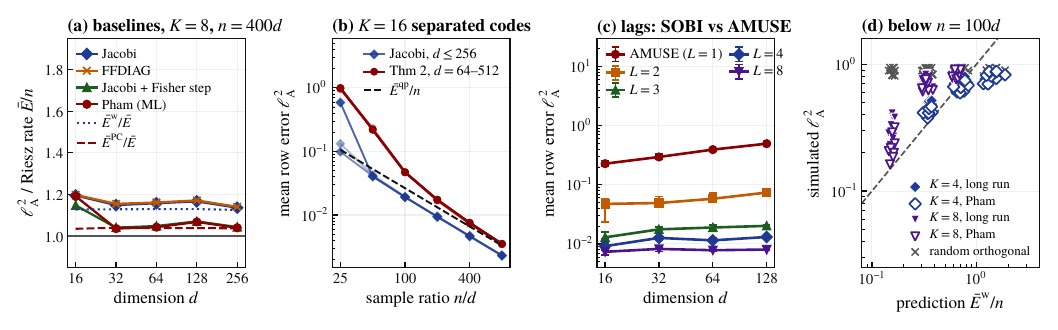}
\caption{\textbf{Algorithms, attainment and the lag route.} \textbf{(a)}~$K=8$, $n=400d$, identical data for all methods (medians over three to six draws), error divided by the Riesz rate $\Ebar/n$ (solid line at $1$); dashed: $\Ebar^{\rm PC}/\Ebar$, the Cram\'er--Rao coefficient with estimated whitening; dotted: $\Ebar^{\rm w}/\Ebar$. \textbf{(b)}~Separated signatures ($K=16$ balanced codes, $S_{\min}\ge\CodeSmin$): the estimator of Theorem~\ref{thm:upper} with known whitening (red) and Jacobi with estimated whitening (blue, $d\le256$) against $n/d$ for $d=\CodedMin$ to $\CodedMax$ (light to dark); dashed: $\Ebar^{\rm qp}/n$. \textbf{(c)}~SOBI with $L$ lags on MA(8) sources, $n=200d$ (medians and quartiles); AMUSE is $L=1$. \textbf{(d)}~Runs below $n=100d$ ($K\in\{4,8\}$, $d\in\{\LowDs\}$): Jacobi continued to $1000$ sweeps (filled) and Pham's algorithm (open) against $\Ebar^{\rm w}/n$; crosses: a random orthogonal matrix.}
\label{fig:review}
\end{figure}

\begin{figure}[tb]
\centering
\includegraphics[width=\textwidth]{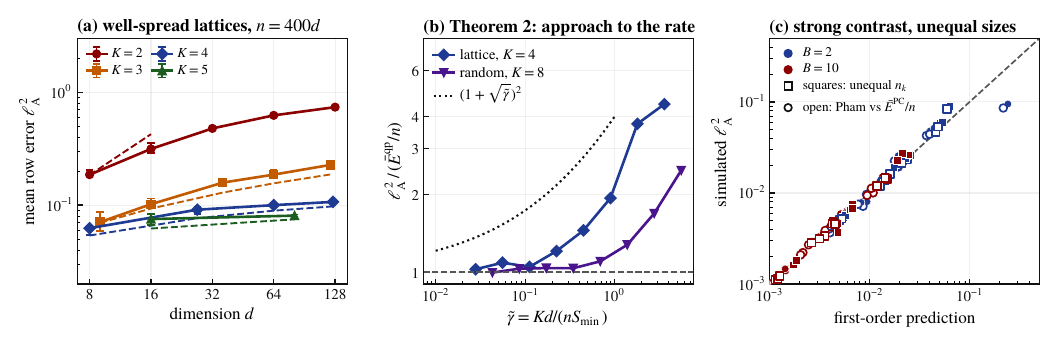}
\caption{\textbf{(a)}~Lattice signatures (Definition~\ref{def:qu}), $n=400d$, eight draws: the dimension law with the assumptions of Theorems~\ref{thm:rate} and~\ref{thm:klaw}. \textbf{(b)}~The estimator of Theorem~\ref{thm:upper} (known whitening) approaches its prediction $\Ebar^{\rm qp}/n$ as $\tilde\gamma=Kd/(nS_{\min})$ decreases ($d=64$, four draws per point; diamonds: $K=4$ lattice, with ratio $\QPconvLat$ to $\Ebar^{\rm qp}/n$ at $\tilde\gamma=\QPconvLatgamma$; triangles: $K=8$ random signatures); dotted: $(1+\sqrt{\tilde\gamma})^2$, the factor of \eqref{eq:upper} with $C=1$ and $\tilde\gamma$ in place of $\gamma_j$. \textbf{(c)}~Strong contrast ($B=10$, red) and unequal sizes (squares): Jacobi against $\Ebar^{\rm w}/n$ (filled) and Pham against $\Ebar^{\rm PC}/n$ (open).}
\label{fig:extra}
\end{figure}

\begin{figure}[tb]
\centering
\includegraphics[width=\textwidth]{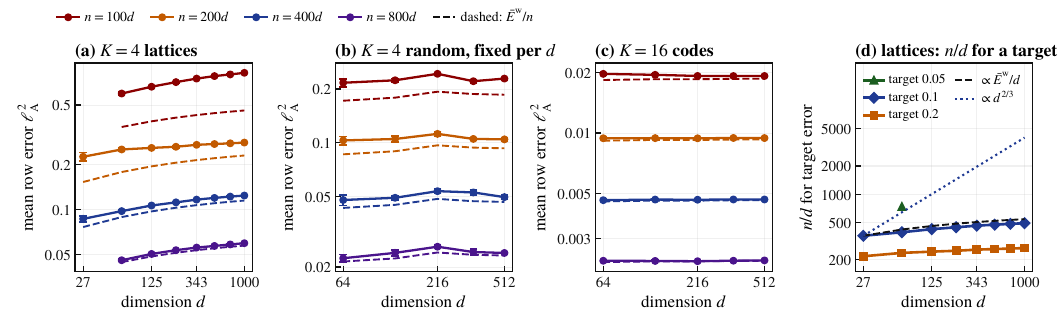}
\caption{\textbf{Mean risk on fixed configurations, larger $d$.} Mean $\lA^2$ with $95\%$ intervals against $d$ at $n/d\in\{100,200,400,800\}$; dashed: $\Ebar^{\rm w}/n$. \textbf{(a)}~$K=4$ lattices, $d=27$ to $1000$. \textbf{(b)}~One fixed draw of $K=4$ random signatures per $d$. \textbf{(c)}~$K=16$ balanced codes. \textbf{(d)}~$n/d$ at which the mean error of the lattices reaches $0.1$ and $0.2$ (and $0.05$, reached within $n\le800d$ only at $d=\LargeCrossFivedMin$); dashed: the growth of $\Ebar^{\rm w}/d$; dotted: the growth $d^{2/3}$ of the proved threshold.}
\label{fig:larged}
\end{figure}

The supplementary material contains the code, the raw results and three tables: the runs below $n=100d$ (means per setting; Figure~\ref{fig:review}(d) plots the individual runs), the mean risk on fixed lattices (Figure~\ref{fig:larged}(a), and $n=50d$) and the values of Figure~\ref{fig:extra}(c). Its README maps every figure, table and number of this appendix to the script that produced it.

\paragraph{Algorithms.} Algorithm~\ref{alg:jacobi} is implemented in C and agrees with a NumPy implementation to machine precision. Each sweep visits the pairs in lexicographic order with the Cardoso--Souloumiac angle, and a run stops when the largest $|\sin\theta|$ of a sweep is below $10^{-10}$, or after $100$ sweeps; in the sample-complexity runs with $n\ge100d$, $\ConvBig$ of $\NBig$ converged, with a median of $\SweepsMedian$ sweeps. \emph{Pham's algorithm} \citep{pham2001joint} minimises $\sum_k\pi_k[\log\det\diag(T\hat\Sigma_kT^\top)-\log\det(T\hat\Sigma_kT^\top)]$ over unmixing matrices $T$, the Gaussian likelihood of the non-orthogonal model, by pairwise updates (the update of the reference implementation in \texttt{pyriemann}, implemented in C; raw covariances, at most $300$ sweeps, rows of $T$ normalised). \emph{FFDIAG} \citep{ziehe2004fast} updates $V\leftarrow(I+W)V$ with $W$ solving the $2\times2$ systems of the Frobenius criterion on the whitened covariances, step capped at $\|W\|_F\le0.9$. The \emph{Fisher step} is the weighted least-squares step defined under ``Attainment'' below, applied to the covariances with estimated whitening. Appendix~\ref{app:ajd} gives the predictions for unequal sizes and for Fisher weights with estimated whitening, and the Cram\'er--Rao coefficient $v^{\rm PC}$ of the non-orthogonal problem.

\begin{algorithm}[htb]
\caption{Orthogonal joint diagonalization (Jacobi sweeps)}\label{alg:jacobi}
\begin{algorithmic}
\STATE whiten by the pooled covariance; form $\Sh_1,\dots,\Sh_K$; $V\leftarrow I_d$
\REPEAT
\FOR{each pair $a<b$}
\STATE $g_k\leftarrow\big((\Sh_k)_{aa}-(\Sh_k)_{bb},\ 2(\Sh_k)_{ab}\big)$ for $k\le K$
\STATE $G\leftarrow\sum_kg_kg_k^\top$
\STATE $\theta\leftarrow\frac14\operatorname{atan2}\big(2G_{12},\ G_{11}-G_{22}\big)$\hfill($\operatorname{atan2}(0,0)=0$: $\theta=0$ iff $G_{12}=0\le G_{11}-G_{22}$)
\STATE rotate rows/columns $(a,b)$ of every $\Sh_k$ and columns $(a,b)$ of $V$ by $\theta$
\ENDFOR
\UNTIL{all angles are below tolerance}
\end{algorithmic}
\end{algorithm}

\begin{figure}[tb]
\centering
\includegraphics[width=\textwidth]{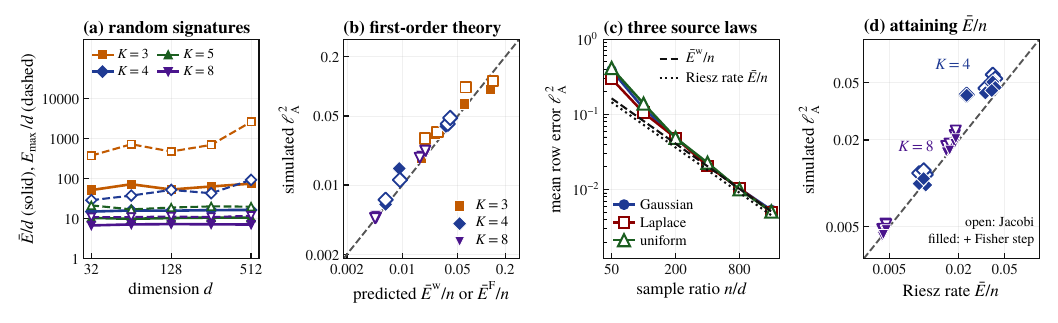}
\caption{\textbf{Verification.} \textbf{(a)}~Energy per source for random signatures (medians over eleven draws): the average $\Ebar/d$ (solid) saturates from $K=4$ on, the worst source $\Emax/d$ (dashed) only from $K=5$ on (Proposition~\ref{prop:random}); $K=2$ is off scale. \textbf{(b)}~Simulated mean loss $\lA^2$ against the first-order prediction for $K\in\{3,4,8\}$, $d\in\{16,32\}$ and $n\in\{400d,1600d\}$, one signature draw and six mixings each: filled markers use estimated whitening and $\Ebar^{\rm w}/n$, open markers known whitening and $\Ebar^{\rm F}/n$; dashed: identity. Points off the line are draws with a pair outside the first-order regime. \textbf{(c)}~Gaussian, Laplace and uniform sources ($d=32$, $K=8$, one signature draw, eight mixings each) give the same error from $n=200d$ on, on top of $\Ebar^{\rm w}/n$ (dashed; ratios $1.03$, $1.01$ and $1.11$ at $n=400d$) and above the Riesz rate $\Ebar/n$ (dotted). \textbf{(d)}~Attaining the Riesz rate (known whitening, $K\in\{4,8\}$, $d\in\{16,\dots,128\}$, $n\in\{400d,1600d\}$): Algorithm~\ref{alg:jacobi} (open) sits at $\Ebar^{\rm F}/n$, above the Riesz rate; one Fisher-weighted step (filled) lands on it.}
\label{fig:verify}
\end{figure}

\begin{figure}[tb]
\centering
\includegraphics[width=\textwidth]{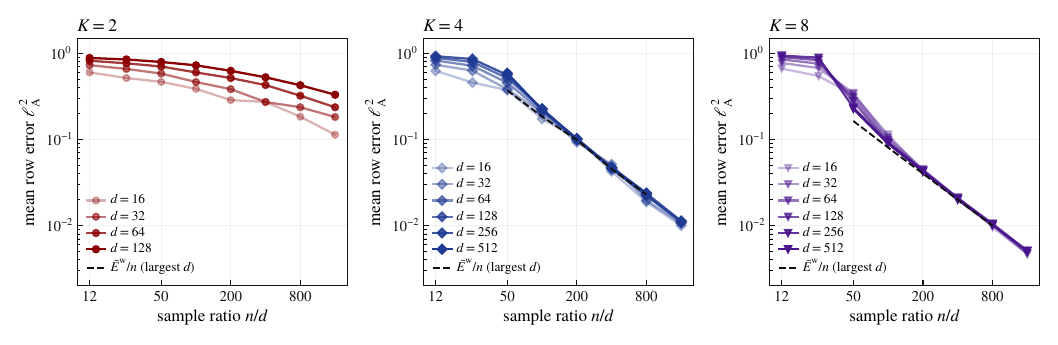}
\caption{\textbf{Collapse in $n/d$.} Median error against $n/d$ for random signatures, $K\in\{2,4,8\}$ and $d$ from $16$ (light) to $512$ (dark; $K=2$ up to $128$), ten draws per point ($d\le128$), six ($d=256$) or three ($d=512$); dashed: $\Ebar^{\rm w}/n$ for the largest $d$. For $K=4$ and $K=8$ the curves coincide; for $K=2$ they move up with $d$.}
\label{fig:collapse}
\end{figure}

\paragraph{Simulation and draws.} For Gaussian sources the environment covariances are sufficient statistics: $n_k\hat\Sigma_k=A\Lambda_k^{1/2}GG^\top\Lambda_k^{1/2}A^\top$ with $G$ a $d\times n_k$ standard Gaussian matrix, and $GG^\top$ is drawn through the Bartlett decomposition when $n_k\ge d$ (exact in law, $O(d^3)$) and formed from $G$ otherwise. The halves of Theorem~\ref{thm:upper} are two independent such draws with $n_k/2$ samples; the other methods use their average, the full-sample covariance. Each draw uses new signatures (when random), a Haar-distributed $A$ and new data. Curves show medians and quartiles over draws; Figures~\ref{fig:verify}(c,d) and~\ref{fig:larged} show means over draws of $A$ and the data for fixed signature configurations. Values of $\lA^2$ above $1/2$ (for example in Figures~\ref{fig:mech}(a), \ref{fig:review}(d), \ref{fig:larged} and~\ref{fig:collapse}) are lower bounds for a permutation-matched loss (Section~\ref{sec:setting}).

\paragraph{Signatures.} \emph{Random}: i.i.d.\ uniform on $\{\sig\in[\frac12,\frac32]^K:\sum_k\sig(k)=K\}$, drawn by sampling the first $K-1$ coordinates uniformly and accepting if the last one lies in $[\frac12,\frac32]$ (a linear map, so the accepted law is uniform; $B=2$). \emph{Lattice}: $\sig(k)=1+\epsilon u_k$ ($k<K$), $u$ on a regular grid of $m^{K-1}$ points in $(-\frac12,\frac12)^{K-1}$, and $\sig(K)=1-\epsilon\sum_{k<K}u_k$. With $\epsilon=\frac12$, used for $K\le4$, every variance lies in $(\frac14,\frac74)$, so $B<4$ and Definition~\ref{def:qu} holds with fixed constants; for $K=5$, where $\sig(K)$ would approach $0$, Figure~\ref{fig:story}(c) uses $\epsilon=\frac14$ (variances in $(\frac12,\frac32)$), and the $K=5$ lattices of Figure~\ref{fig:extra}(a), with $\epsilon=\frac12$, have $B=2$ and $3$ at $d=16$ and $81$. \emph{Halton}: $\sig_j=\mathbf 1+sVz_j$ with $z_j$ the first $d$ points of the Halton sequence in $[-\frac12,\frac12]^{D'}$, $V$ a random orthonormal $K\times D'$ basis orthogonal to $\mathbf 1$, and a fixed scale $s$ that keeps every coordinate in $[\frac12,\frac32]$; the separation $\min_{i\neq j}\|z_i-z_j\|$ and the covering radius in the cube, multiplied by $d^{1/D'}$, stay in $[0.22,0.81]$ and $[0.99,1.91]$ for $d=8$ to $256$, so the configurations are well spread in the range simulated. \emph{Codes}: $\sig_j=\mathbf 1+\frac12y_j$ with balanced $y_j\in\{\pm1\}^{16}$ at pairwise Hamming distance at least $4$, built greedily from random balanced words ($S_{\min}\ge\CodeSmin$). \emph{Equal gap, different energy} (Figure~\ref{fig:mech}(b)): configuration A is the $K=4$ lattice with $d=27$; configuration B takes $26$ points of a coarser $K=4$ lattice ($\epsilon=1$, without the point whose last variance is $0$; variances in $[\frac13,2]$, $B=3$) and adds one point at Fisher divergence $S_{\min}(A)$ from one of them.

\paragraph{Figure details.} Figure~\ref{fig:story}(b): medians and quartiles over sixteen draws (eight at $d=128$); the well-spread case, lattices, is Figure~\ref{fig:extra}(a). Figure~\ref{fig:story}(c): see ``Signatures''. Figure~\ref{fig:story}(d): known signatures, whitening by the pooled covariance of the same data, polar factor; six draws per $d$. Figure~\ref{fig:mech}(a): medians over eight draws, four for $d\ge128$; (b): twelve draws, with standard errors; (c): see Figure~\ref{fig:collapse}; (d): medians and quartiles over draws.

\paragraph{Lags and runs below $n=100d$.} The lag-signature of an MA(8) source $s_t=\sum_{i=0}^8b_i\epsilon_{t-i}$ ($b\sim N(0,I_9)$, normalised to unit variance) is $(\sum_ib_ib_{i+\tau})_{\tau\le L}$; AMUSE takes the eigenvectors of the whitened, symmetrised lag-1 covariance, and SOBI applies Algorithm~\ref{alg:jacobi} to the whitened, symmetrised covariances at lags $1,\dots,L$. Lag-signatures have no sum constraint, so $L$ lags give a configuration of dimension up to $L$, and the first-order variance of SOBI \citep[Thm.~2]{miettinen2016separation} has their Euclidean gap in its denominator: the dimension count suggests $n\gtrsim d^2$ for AMUSE and $n\gtrsim d$ for SOBI with $L\ge3$ lags whose lag-signatures are well spread, a heuristic outside our theorems, since lags are not independent environments. At $n=200d$, AMUSE degrades from $\AmuseFirst$ to $\AmuseLast$ as $d$ goes from $\SobidMin$ to $\SobidMax$, SOBI with two lags from $\SobiTwoFirst$ to $\SobiTwoLast$, and SOBI with $L\ge3$ lags stays between $\SobiThreeMin$ and $\SobiThreeMax$ (Figure~\ref{fig:review}(c)). Below $n=100d$ (Figure~\ref{fig:review}(d); $\LowN$ runs, $n/d\in\{12,25,50\}$, $\LowReps$ draws per setting), where \LowCappedPhrase{} stop at the $100$-sweep cap, we keep the data and the whitening, continue Jacobi to $1000$ sweeps, restart it from four Haar-random rotations (at most $300$ sweeps each) and run Pham's algorithm. Continuing changes the error by at most $\LowLongChangeMax$ (median $\LowLongChangeMed$), the restarts end within $\LowRestartSpreadMed$ of each other (median; at most $\LowRestartSpreadMax$), their criterion values are within $\LowCritGainMax$ of the continued run, and $\LowConvLong$ of the $\LowN$ continued runs converge: Jacobi is stable under initialisation, which does not show that the landscape is benign. Pham's algorithm is $\LowPhamGainMin$--$\LowPhamGainMax$ more accurate at $K=8$, $n=50d$, $\LowPhamGainOtherMin$--$\LowPhamGainOtherMax$ in the other settings with $n\ge25d$, and about as accurate at $n=12d$, where $\Ebar^{\rm w}/n$ is about $\LowTwelvePredMed$ (the first-order prediction is not meaningful above about $0.3$).

\begin{table}[tb]
\caption{Baselines on identical data, $n=400d$ (means over draws, $\times10^3$); the estimator of Theorem~\ref{thm:upper} uses the known signatures and whitening, and the column gives the loss of its labelled vectors (Table~\ref{tab:step} gives their polar factor).}
\label{tab:baselines}
\centering\small\setlength{\tabcolsep}{4pt}
\begin{tabular}{rrccccccccc}\toprule
$K$ & $d$ & Jacobi & FFDIAG & +step & Pham & Thm\,\ref{thm:upper} & pencil & $\Ebar^{\rm w}/n$ & $\Ebar^{\rm PC}/n$ & $\Ebar/n$\\\midrule
8 & 16 & 21.0 & 21.0 & 20.1 & 20.9 & 34.6 & 471 & 19.8 & 18.2 & 17.6\\
8 & 32 & 18.4 & 18.5 & 16.7 & 16.6 & 32.6 & 545 & 18.1 & 16.7 & 16.0\\
8 & 64 & 20.3 & 20.3 & 18.4 & 18.3 & 32.1 & 666 & 19.8 & 18.2 & 17.5\\
8 & 128 & 20.1 & 20.2 & 18.5 & 18.4 & 32.0 & 779 & 19.5 & 17.9 & 17.3\\
8 & 256 & 20.6 & 20.6 & 18.9 & 18.8 & 33.2 & 842 & 20.4 & 18.8 & 18.1\\
4 & 16 & 37.4 & 37.5 & 36.4 & 36.7 & 72.5 & 366 & 34.1 & 31.9 & 31.0\\
4 & 32 & 61.4 & 61.5 & 56.3 & 52.6 & 115.7 & 483 & 42.1 & 39.2 & 38.0\\
4 & 64 & 46.1 & 46.2 & 41.8 & 40.6 & 128.8 & 579 & 40.3 & 37.0 & 35.9\\
4 & 128 & 51.3 & 51.5 & 47.3 & 43.6 & 170.9 & 677 & 43.9 & 40.4 & 39.2\\
4 & 256 & 49.9 & 50.0 & 45.8 & 44.6 & 214.7 & 764 & 46.0 & 42.1 & 40.9\\
\bottomrule\end{tabular}

\end{table}

\paragraph{The linear pencil as an initialisation.} Theorem~\ref{thm:pencil} concerns the eigenvectors of one combination as an estimate. As a starting point for Algorithm~\ref{alg:jacobi} they cost nothing ($K=8$ random signatures, $n=400d$, estimated whitening, $d\in\{\PencilInitDs\}$, \PencilInitReps{} draws each). The eigenvectors of one random combination $\sum_k\alpha_k\Sh_k$, $\alpha\sim N(0,I_K)$, have median error \PencilInitPencilFirst{} at $d=16$ and \PencilInitPencilLast{} at $d=\PencilInitDmax$. One Jacobi sweep from them brings it to \PencilInitSweepOneFirst{} and \PencilInitSweepOneLast, two sweeps to \PencilInitSweepTwoFirst{} and \PencilInitSweepTwoLast, and three sweeps to within \PencilInitSweepThreeMaxExcess{} of the converged run from the identity (\PencilInitConvMin--\PencilInitConvMax) at every $d$. Run to convergence, the two starting points give the same minimiser: the errors differ by less than $\PencilInitConvRelDiffBound$ in relative terms and the squared sines between matched columns are below $\PencilInitMinimiserSinBound$ (\PencilInitAllConverged{} runs converged, within \PencilInitSweepsMax{} sweeps). Starting from the eigenvectors $V_0$ is the same as running the sweeps from the identity on the rotated covariances $V_0^\top\Sh_kV_0$.

\paragraph{Robustness.} \emph{Strong contrast and unequal sizes} (Figure~\ref{fig:extra}(c), means over six draws): for $B=10$ the random signatures are uniform on the slice of $[0.1,1.9]^K$ (ratios bounded by $10$ up to the renormalisation for unequal sizes); unequal sizes have $\pi_k$ geometric from $1$ to $10$, normalised; Jacobi uses the $\pi$-weighted criterion and pooled covariance and is compared with the $\pi$-weighted $\Ebar^{\rm w}/n$, and Pham's algorithm uses the weights $\pi_k$. Over the $B=10$ runs and the unequal-size runs, the largest deviations of Jacobi from $\Ebar^{\rm w}/n$ ($\ContrastJacMax$ at $B=10$, $\UnequalJacMax$ at $B=2$) are both for $K=\WorstK$ with unequal sizes at $n=\WorstRatio d$, where the smallest environment has only $\WorstNmin d$ samples; these two mean errors have relative standard errors $\WorstSElo$ and $\WorstSEhi$. The equal-size $B=2$ point far below the diagonal ($K=4$, $d=16$) has $\Ebar^{\rm w}/n\approx0.24$, outside the first-order regime. From $n=200d$ on, Laplace and uniform sources give the error of Gaussian ones (Figure~\ref{fig:verify}(c)). \emph{Departures from the model} (Table~\ref{tab:stress}; $K=\StressK$ random signatures, $d=\StressD$, $n/d\in\{200,800\}$, \StressReps{} draws of signatures, mixing and data per cell): with AR(1) sources of coefficient $\phi$ in every environment, the variance of an off-diagonal sample covariance, and with it the prediction, is multiplied by $\tau=(1+\phi^2)/(1-\phi^2)$ (Bartlett's formula; \citealp{bartlett1935some}); at $\phi=0.5$ the error is \StressArHalfMin--\StressArHalfMax{} times $\tau\Ebar^{\rm w}/n$, at $\phi=0.8$ ($\tau=\StressArHighTau$) \StressArHighHigh{} times at $n=800d$ and \StressArHighLow{} times at $n=200d$, where the effective sample size is $\StressArHighNeff d$ and \StressArHighConv{} of the \StressReps{} runs converge. Mixings $A(I+\epsilon G_k/\sqrt d)$ that differ across environments ($G_k$ Gaussian) give at most \StressMixSmallMax{} times the prediction for $\epsilon\le0.03$ and leave a floor of about $\StressMixFloor\cdot10^{-3}\approx\StressMixFloorOverEps\,\epsilon^2$ at $\epsilon=0.1$ (fitted from the two sample sizes as $a/n+f$); a shared factor with correlation $r$ between all sources leaves a floor of $\StressCorrFloorLow\cdot10^{-3}$ at $r=0.05$ and $\StressCorrFloorHigh\cdot10^{-3}$ at $r=0.1$. At $B\in\{2,4,10\}$ the ratio is \StressContrastMin--\StressContrastMax. In every cell the predicted per-source errors $E^{\rm w}_j/n$ rank the observed ones (Spearman correlation \StressSpearmanMin--\StressSpearmanMax, mean over draws): the diagnostic still says which source is hard when its constant is off.

\begin{table}[tb]
\caption{The first-order prediction under departures from the model ($K=8$, $d=64$, eight draws per cell, means): mean $\lA^2$ ($\times10^3$), its ratio to the prediction (multiplied by $\tau$ for AR(1) sources), and the Spearman correlation between the predicted and the observed per-source errors.}
\label{tab:stress}
\centering\small
\begin{tabular}{lrcccrccc}\toprule
 & & \multicolumn{3}{c}{$n=200d$} & & \multicolumn{3}{c}{$n=800d$}\\\cmidrule{3-5}\cmidrule{7-9}
departure & & error & error$/$pred. & Spearman & & error & error$/$pred. & Spearman\\\midrule
AR(1), $\phi=$0 & & 45.0 & 1.12 & 0.73 & & 10.4 & 1.04 & 0.69\\
AR(1), $\phi=$0.5 & & 76.9 & 1.17 & 0.65 & & 16.5 & 1.02 & 0.70\\
AR(1), $\phi=$0.8 & & 479.1 & 2.65 & 0.54 & & 51.0 & 1.11 & 0.66\\
mixing, $\epsilon=$0 & & 43.3 & 1.09 & 0.71 & & 10.5 & 1.02 & 0.73\\
mixing, $\epsilon=$0.01 & & 46.6 & 1.17 & 0.70 & & 10.4 & 1.05 & 0.67\\
mixing, $\epsilon=$0.03 & & 45.5 & 1.16 & 0.61 & & 12.1 & 1.22 & 0.68\\
mixing, $\epsilon=$0.1 & & 74.4 & 1.80 & 0.72 & & 35.4 & 3.53 & 0.67\\
shared factor, $r=$0 & & 43.6 & 1.09 & 0.69 & & 9.9 & 1.01 & 0.67\\
shared factor, $r=$0.05 & & 46.1 & 1.19 & 0.69 & & 14.4 & 1.44 & 0.68\\
shared factor, $r=$0.1 & & 48.9 & 1.22 & 0.72 & & 16.6 & 1.72 & 0.74\\
contrast, $B=$2 & & 43.2 & 1.09 & 0.65 & & 10.5 & 1.02 & 0.70\\
contrast, $B=$4 & & 18.0 & 1.06 & 0.72 & & 4.4 & 1.01 & 0.72\\
contrast, $B=$10 & & 12.3 & 1.04 & 0.74 & & 2.9 & 0.99 & 0.78\\
\bottomrule\end{tabular}

\end{table}

\paragraph{How many environments.} One simulated recording of $n=\SegRatio d$ samples, $d=\SegD$, has sources whose log-variances are random combinations of three low-frequency cosines (\SegReps{} draws of trajectories, mixing and data); cutting it into $K\in\{2,4,\dots,\SegKmax\}$ equal segments gives $K$ environments, with the average variance in a segment as the signature (Table~\ref{tab:segments}). Two segments put the signatures on a line and fail ($\lA^2=\SegTwoLoss$); from $K=4$ the error follows $\Ebar^{\rm w}(K)/n$ (ratio \SegFourRatio{} at $K=4$, within \SegFlatRatioMax{} for $8\le K\le64$), which stops decreasing once the segments resolve the dimension of the trajectories (participation ratio of the centred signatures about \SegFlatPRMin), and the error leaves the prediction only for short segments: \SegRiseRatio{} times at $K=\SegRiseK$ ($n/K\approx\SegRiseNk d$) and \SegLastRatio{} times at $K=\SegKmax$ ($n/K\approx\SegLastNk d$).

\begin{table}[tb]
\caption{Segments of one nonstationary recording as environments ($d=32$, $n=800d$, eight draws): mean $\lA^2$ ($\times10^3$) with its standard error, the median first-order prediction $\Ebar^{\rm w}(K)/n$ ($\times10^3$; off scale for $K=2$), the median per-draw ratio, the participation ratio of the centred signature matrix, and the runs that met the tolerance.}
\label{tab:segments}
\centering\small
\begin{tabular}{rrcccccc}\toprule
$K$ & $n/(Kd)$ & mean $\lA^2$ & s.e. & $\Ebar^{\rm w}(K)/n$ & median ratio & participation ratio & converged\\\midrule
2 & 400 & 371.1 & 22.2 & -- & 0.01 & 1.0 & 8/8\\
4 & 200 & 24.6 & 3.0 & 22.0 & 1.07 & 2.7 & 8/8\\
8 & 100 & 6.7 & 0.4 & 6.5 & 0.99 & 4.8 & 8/8\\
16 & 50 & 5.8 & 0.3 & 5.7 & 1.00 & 5.0 & 8/8\\
32 & 25 & 5.7 & 0.3 & 5.5 & 1.01 & 5.1 & 8/8\\
64 & 12.5 & 5.9 & 0.3 & 5.5 & 1.04 & 5.1 & 8/8\\
128 & 6.25 & 6.6 & 0.5 & 5.5 & 1.17 & 5.1 & 8/8\\
256 & 3.125 & 9.7 & 2.2 & 5.5 & 1.34 & 5.1 & 8/8\\
\bottomrule\end{tabular}

\end{table}

\paragraph{Mean risk on fixed configurations.} The theorems bound expected risks, while most curves show medians, so we fix one configuration per $d$, the $K=4$ lattice, and redraw only the mixing and the data (Figure~\ref{fig:larged}(a); at least \LargeRepsMin{} draws per point; all values in the supplementary table). At $n=400d$ the mean error is $\MeanRiskRatioMin$--$\MeanRiskRatioMax$ times $\Ebar^{\rm w}/n$ from $d=\MeanRiskdMin$ to $\MeanRiskdMax$, with standard errors at most $\MeanRiskSEMax$ of the mean and $90\%$ quantiles within $\MeanRiskQNinetyMax$ of it, and at $n=800d$ ($d=64$ to $\LargeDMax$) it is $\MeanRiskEightRatioMin$--$\MeanRiskEightRatioMax$ times. At $n=200d$, where $\Ebar^{\rm w}/n$ is $\MeanRiskTwoPredMin$--$\MeanRiskTwoPredMax$ and most runs stop at $100$ sweeps, the ratio falls from $\MeanRiskTwoFirst$ to $\MeanRiskTwoLast$ as $d$ grows, and the sweep cap does not explain the excess: continued to $1000$ sweeps, the $\ContRuns$ capped runs at $n=200d$ (all for $d\le216$, the first four at $d=343$) change their loss by at most $\ContMaxChange$, the mean of the continued runs at each $d$ by at most $\ContMeanChange$, and $\ContConv$ of them then meet the tolerance. At $n=100d$ ($d=64$ to $\LargeDMax$), where $\Ebar^{\rm w}/n$ is $\MeanRiskHundredPredMin$--$\MeanRiskHundredPredMax$, outside the first-order regime, the ratio is $\MeanRiskHundredRatioMin$--$\MeanRiskHundredRatioMax$, and $\MeanRiskHundredLargeMin$--$\MeanRiskHundredLargeMax$ for $d\ge512$; at $n=50d$ ($d=64$ to $343$) the error is $\MeanRiskFiftyMin$--$\MeanRiskFiftyMax$, and there its ratio to the prediction, which is not meaningful above about $0.3$, says nothing; at $n\le100d$ \MeanRiskLowConv{} of \MeanRiskLowRuns{} runs meet the tolerance within $100$ sweeps. The crossings of Figure~\ref{fig:larged}(d) interpolate $\log$ of the mean linearly in $\log n$ between neighbouring sample sizes: from $d=27$ to $\LargeDMax$ the $n/d$ at which the mean error reaches $0.1$ grows by $\MeanRiskCrossGrowth$ (Section~\ref{sec:experiments}) and the one for $0.2$ by $\LargeCrossTwentyGrowth$, against $\MeanRiskEwGrowth$ for $\Ebar^{\rm w}/d$ and $\MeanRiskProvedGrowth$ for the proved threshold $d^{5/3}$. A run at $d=1000$ takes \LargeSecMin--\LargeSecMax{} minutes on one core. One fixed draw of $K=\RandK$ random signatures with $B=2$ and $K=\CodesK$ balanced codes with $S_{\min}\ge\CodeSmin$ behave alike for $d=\RanddMin$ to $\RanddMax$ (Figure~\ref{fig:larged}(b,c)): at $n=400d$ their mean errors are \RandRatioFourRange{} and \CodesRatioFourRange{} times $\Ebar^{\rm w}/n$, and the $n/d$ for a target error of $\RandCrossTarget$ ($\CodesCrossTarget$ for the codes) grows by $\RandCrossGrowth$ ($\CodesCrossGrowth$), against $\RandEwGrowth$ ($\CodesEwGrowth$) for $\Ebar^{\rm w}/d$.

\paragraph{Attainment.} With known whitening, the Fisher step after Algorithm~\ref{alg:jacobi} takes $M_k=\hat V^\top\Sh_k\hat V$, $\hat\sig_a(k)=(M_k)_{aa}$ and $\hat\delta_k=\hat\sig_a(k)-\hat\sig_b(k)$, estimates the residual angle of each pair by weighted least squares, $\hat\Xi_{ab}=\sum_kw_k\hat\delta_k(M_k)_{ab}/\sum_kw_k\hat\delta_k^2$ with $w_k=1/(\hat\sig_a(k)\hat\sig_b(k))$, and returns $\hat V\exp(-\hat\Xi)$; by Proposition~\ref{prop:ajd}(ii) its first-order variance per pair is $S_{ab}^{-1}$ (Figure~\ref{fig:verify}(d)). \emph{The estimator of Theorem~\ref{thm:upper} below its proved regime} (Table~\ref{tab:step}; known signatures and whitening, random signatures, $n=400d$, $\tilde\gamma=Kd/(nS_{\min})$ from $\StepGammaEightMin$ to $\StepGammaEightMax$ for $K=8$ and up to $\StepGammaFourMax$ for $K=4$ over the draws; Table~\ref{tab:step} reports medians and means): over the draws, for $K=8$ its labelled vectors are $\StepLabEightMin$--$\StepLabEightMax$ times $\Ebar/n$, their polar factor $\StepPolarEightMin$--$\StepPolarEightMax$ times (it averages the two estimates of each rotation angle, from the pencils at $\sig_i$ and $\sig_j$), and one Fisher step from it $\StepStepEightMin$--$\StepStepEightMax$ times; for $K=4$, where random signatures have close pairs, the three are $\StepLabFourMin$--$\StepLabFourMax$, $\StepPolarFourMin$--$\StepPolarFourMax$ and $\StepStepFourMin$--$\StepStepFourMax$ times $\Ebar/n$.

\begin{table}[tb]
\caption{The estimator of Theorem~\ref{thm:upper} on random signatures, $n=400d$, three draws per row (means, $\times10^3$; $\tilde\gamma$: median): labelled vectors, their polar factor, and one Fisher step from it, with the first-order coefficient of the labelled vectors, the Riesz rate, and the ratio of the polar factor's loss to the Riesz rate.}
\label{tab:step}
\centering\small
\begin{tabular}{rrccccccc}\toprule
$K$ & $d$ & $\tilde\gamma$ & labelled & polar & +step & $\Ebar^{\rm qp}/n$ & $\Ebar/n$ & polar$/(\Ebar/n)$\\\midrule
8 & 64 & 1.4 & 35.7 & 20.7 & 18.7 & 22.2 & 18.3 & 1.13\\
8 & 128 & 2.7 & 32.7 & 19.5 & 18.1 & 21.5 & 17.5 & 1.11\\
8 & 256 & 3.0 & 33.1 & 19.7 & 18.4 & 21.7 & 17.8 & 1.11\\
4 & 64 & 18.1 & 143.5 & 71.8 & 46.2 & 43.0 & 40.0 & 1.79\\
4 & 128 & 27.0 & 173.3 & 78.0 & 41.9 & 42.3 & 39.3 & 1.98\\
4 & 256 & 55.1 & 186.3 & 85.0 & 48.9 & 42.9 & 39.9 & 2.13\\
\bottomrule\end{tabular}

\end{table}

\paragraph{Estimated whitening (Theorem~\ref{thm:upper}(b)).} Table~\ref{tab:whiten} runs the estimator of Theorem~\ref{thm:upper} on $K=4$ lattices with known signatures and the sample split into thirds: on parts $1$ and $2$ with the true whitening, and on the same parts whitened by the pooled sample covariance of part $0$, scored by $\rho_j$ of Theorem~\ref{thm:upper}(b); the comparison is with known whitening on the same two thirds. Mixing the same sources by $I+G/\sqrt d$ ($G$ Gaussian, $d=\WhEquivD$, condition number $\WhCond$) instead of $A=I$ changes no $\rho_j$ by more than $10^{\WhEquivExp}$, as the reduction in the proof says. On these lattices $\Ebar^{\rm wh}=\WhEwh\,\Ebar$, so estimating the whitening costs $\WhPredExcessOne$ at first order; the measured excess is $\WhExcessHighMin$--$\WhExcessHighMax$ at $n=25600d$ ($\gamma\le\WhGammaHighMax$) and $\WhExcessLowMin$--$\WhExcessLowMax$ at $n=6400d$ ($\gamma=\WhGammaLowMin$--$\WhGammaLowMax$), and for $d=27$ it falls to $\WhExcessMidOne$ and $\WhExcessTopOne$ at $n=102400d$ and $409600d$: a second-order term, which the factor $(1+C\sqrt{\gamma_j})^2$ of \eqref{eq:upper-wh} absorbs. To resolve the coefficient $2E^{\rm wh}_j$ we take $K=4$, $d=8$, $n=2\cdot10^6d$ and $\WhFoReps$ draws on a configuration chosen for a large $E^{\rm wh}$ (the largest $\Ebar^{\rm wh}/\Ebar$ among $4000$ draws on $\{\sig\in[\frac14,\frac74]^K:\sum_k\sig(k)=K\}$ with $S_{\min}\ge0.05$; $E^{\rm wh}_j/E_j$ up to $\WhFoTiltEwhMax$, $\gamma=\WhFoTiltGamma$): $n\E\rho_j/[\frac32(E^{\rm qp}_j+2E^{\rm wh}_j)]$ is $\WhFoTiltMin$--$\WhFoTiltMax$ for every source (standard errors at most $\WhFoTiltSE$), whereas $E^{\rm wh}_j$ in place of $2E^{\rm wh}_j$ would miss by up to $\WhFoHalfOff$; on a random configuration as in Section~\ref{sec:experiments} ($\Ebar^{\rm wh}=\WhFoRandEwh\,\Ebar$, $\gamma=\WhFoRandGamma$) the ratio is $\WhFoRandMin$--$\WhFoRandMax$.

\begin{table}[tb]
\caption{Theorem~\ref{thm:upper}(b) in simulation: $K=4$ lattices, known signatures, $n/3$ observations per part. The estimator of Theorem~\ref{thm:upper} on parts $1$ and $2$ with the true whitening (known; loss $1-\langle\hat q_j,q_j\rangle^2$) and with the whitening estimated on part $0$ (estimated; loss $\rho_j$ of Theorem~\ref{thm:upper}(b)): means of $\lA^2$ over the stated number of draws ($\times10^3$), their first-order predictions with $n_B=2n/3$, and the ratio of the two means; $\gamma=\max_j\gamma_j$.}
\label{tab:whiten}
\centering\small
\begin{tabular}{rrccccccc}\toprule
$d$ & $n/d$ & draws & $\gamma$ & known & $\Ebar^{\rm qp}/n_B$ & estimated & $(\Ebar^{\rm qp}+2\Ebar^{\rm wh})/n_B$ & estimated$/$known\\\midrule
27 & 6400 & 40 & 0.22 & 7.99 & 7.28 & 8.81 & 7.43 & 1.103\\
27 & 25600 & 40 & 0.060 & 1.87 & 1.82 & 1.95 & 1.86 & 1.042\\
27 & 102400 & 40 & 0.016 & 0.464 & 0.455 & 0.478 & 0.465 & 1.029\\
27 & 409600 & 40 & 0.0043 & 0.114 & 0.114 & 0.117 & 0.116 & 1.023\\
64 & 6400 & 16 & 0.25 & 9.41 & 8.39 & 10.8 & 8.57 & 1.146\\
64 & 25600 & 16 & 0.065 & 2.15 & 2.10 & 2.25 & 2.14 & 1.050\\
125 & 6400 & 8 & 0.30 & 10.6 & 9.10 & 12.3 & 9.30 & 1.154\\
125 & 25600 & 8 & 0.077 & 2.38 & 2.28 & 2.53 & 2.32 & 1.063\\
\bottomrule\end{tabular}

\end{table}

\paragraph{Unknown signatures.} The grid of Algorithm~\ref{alg:grid} is a device for the proof; Table~\ref{tab:unknown} tests a practical variant on $K=4$ lattices with $d=m^3$, known whitening and the sample split into thirds. On part $0$, $\lambda_{\min}\big(\sum_k(\Sh^{(0)}_k-g(k)I)^2\big)$ is computed on a cubic grid of mesh $r_0/4$ in the hyperplane within $[1/2.6,2.6]^K$ ($[1/2.7,2.7]^K$ at $d=216$, whose lattice has $B=8/3$), $r_0$ the smallest distance between signatures; the grid points below all their face neighbours are refined by three alternating steps ($u\leftarrow$ bottom eigenvector, $g(k)\leftarrow u^\top\Sh^{(0)}_ku$), refined probes closer than $r_0/2$ are merged, and the $d$ with the smallest $\lambda_{\min}$ are kept (a global threshold, as in the proof, needs larger $n$). The estimator of Theorem~\ref{thm:upper} then runs on parts $1$ and $2$ at each probe, and its polar factor is compared with the same estimator at the true signatures. From $n=\UnkRatioMin d$ on, all $d$ probes are found in every run, the largest probe error is $\UnkProbeErrMax$ of the nearest-neighbour distance, and the two estimators differ by at most $\UnkRatioOracleMax$; the selection takes $\UnkTimeFirst$\,s at $d=27$ and about $\UnkLargeTime$ minutes at $d=\UnkLargeD$ ($\UnkLargeGrid$ grid points) on one core. At $n=\UnkLowRatio d$ the selection misses up to $\UnkMissMax$ of the $d$ probes, and the loss over all runs, failures included (the estimator at every returned probe, orthonormalised, each source scored against its closest returned vector), is $\UnkEtoELowMin$--$\UnkEtoELowMax$ times the oracle. These probes are far outside the accuracy that Lemma~\ref{lem:probe} assumes ($1/(8B^4)$ of the nearest-neighbour distance), so the lemma is conservative. \emph{Misspecified $s_0$.} In the variant the mesh and the merge radius stand for $s_0$, which in Algorithm~\ref{alg:grid} also sets the acceptance threshold (the variant has none) and the sample size needed: replacing $r_0$ by $cr_0$ corresponds to $s_0\approx c^2S_{\min}$ up to factors of $B$. On the same data ($d\in\{\UnkSensDs\}$, $n=6400d$), $c=1/2$ multiplies the number of grid points by $\UnkSensGridFactor$ and the selection time by about $\UnkSensTimeFactor$ and returns the same probes as $c=1$ in all \UnkSensRuns{} runs, while $c=2$ merges neighbouring sources and no run succeeds: too small an $s_0$ costs time, too large an $s_0$ fails. A Jacobi pilot on an additional independent batch of $n/3$ samples (rotated by a Haar matrix, so that the sweeps do not start at the truth) estimates $r_0$ by the smallest distance $\hat r_0$ between its signatures: in \UnkPilotRuns{} runs ($d\in\{\UnkPilotDs\}$), $\hat r_0/r_0$ lies between $\UnkPilotRatioMin$ and $\UnkPilotRatioMax$ and $\hat S_{\min}/S_{\min}$ between $\UnkPilotSminMin$ and $\UnkPilotSminMax$ (biased low, the safe side, because the lattice has many pairs at distance $r_0$; with a unique closest pair the bias can be upward, which the safety factor must absorb). With $\hat r_0/2$ in place of $r_0$, a safety factor of four on $s_0$, the selection finds all probes \UnkPilotFound{}, and the estimator is within $\UnkPilotDev$ of the oracle.

\begin{table}[tb]
\caption{A practical variant of Algorithm~\ref{alg:grid} (not a test of Theorem~\ref{thm:unknown}): $K=4$ lattices, three draws per row (\UnkLargeRuns{} for $d=\UnkLargeD$). All probes: runs in which the selection returns one probe per source; probe error: largest distance of a probe from its signature over the signature's nearest-neighbour distance (median over those runs); grid and all runs: $\lA^2$ with the selected probes over those runs and over every run, failures included; oracle: true signatures, on the same data; losses and predictions $\times10^3$, with $n_B=2n/3$ the size of parts $1$ and $2$; selection: time on one core.}
\label{tab:unknown}
\centering\small
\begin{tabular}{rrcccccccc}\toprule
$d$ & $n/d$ & all probes & probe error & grid & all runs & oracle & $\Ebar^{\rm qp}/n_B$ & $\Ebar/n_B$ & selection (s)\\\midrule
27 & 1600 & 1/3 & 0.40 & 36.13 & 70.37 & 30.51 & 29.12 & 27.82 & 4\\
27 & 6400 & 3/3 & 0.18 & 6.86 & 6.86 & 6.84 & 7.28 & 6.95 & 3\\
27 & 25600 & 3/3 & 0.10 & 1.81 & 1.81 & 1.81 & 1.82 & 1.74 & 4\\
64 & 1600 & 0/3 & -- & -- & 62.01 & 34.79 & 33.58 & 32.29 & 23\\
64 & 6400 & 3/3 & 0.18 & 8.37 & 8.37 & 8.29 & 8.39 & 8.07 & 22\\
64 & 25600 & 3/3 & 0.08 & 2.06 & 2.06 & 2.06 & 2.10 & 2.02 & 22\\
125 & 1600 & 0/3 & -- & -- & 99.46 & 38.13 & 36.42 & 35.15 & 100\\
125 & 6400 & 3/3 & 0.17 & 9.21 & 9.21 & 9.17 & 9.10 & 8.79 & 100\\
125 & 25600 & 3/3 & 0.07 & 2.17 & 2.17 & 2.16 & 2.28 & 2.20 & 103\\
216 & 6400 & 2/2 & 0.18 & 9.38 & 9.38 & 9.33 & 9.60 & 9.28 & 533\\
\bottomrule\end{tabular}

\end{table}

\paragraph{Real data.} \emph{Data.} HCP S1200 \citep{vanessen2013wuminn}: the \FmriN{} subjects with all fourteen task runs (seven tasks, two runs each; \citealp{barch2013function}) and the first resting-state run, \FmriNrest{} of them with four resting-state runs. Grayordinate time series (MSMAll, from the minimal preprocessing pipelines; \citealp{glasser2013minimal}) were averaged within Schaefer parcels \citep{schaefer2018local} at $d\in\{100,200,400\}$, converted to percent signal change, cleaned per run by regressing out a trend, twelve motion parameters and a discrete-cosine high-pass basis ($0.008$~Hz), and demeaned. \emph{Environments}: the seven tasks and rest, $176$ frames per task run, i.e.\ $\FmriFramesSubjectEnv$ per subject and environment, and a contiguous rest segment of the same length ($\FmriFramesEnv$ frames per environment in the group fit).

\emph{Group fit}. pooled whitening and Algorithm~\ref{alg:jacobi} (at most $\FmriGroupSweeps$ sweeps) on the covariances summed over subjects; the signatures are $\hat\sig_j(k)=\hat b_j^\top\Sh_k\hat b_j/\hat b_j^\top\hat C\hat b_j$, normalised to mean one. Only the eight-task fit at $d=100$ reached the tolerance; at $d=\FmriD$ the pairs are nearly degenerate ($S_{\min}=\FmriSmin$) and Pham's algorithm \citep{pham2001joint} differs from Jacobi by a matching loss of up to $\FmriPhamLossMax$ for such pairs (the energies below are those of the Jacobi solution). \emph{Autocorrelation}: $\tau_{ab}=\sum_t\rho_a(t)\rho_b(t)$ \citep{bartlett1935some} from the estimated sources of \FmriNtau{} subjects (Bartlett window of $L=\FmriBartlettL$ frames; median $\tau_{aa}=\FmriTauDiagMedian$; an AR(2) fit agrees to within \FmriTauARrelMedian{} in median).

\emph{Split halves}. $\FmriRsplit$ random disjoint pairs of halves with $m\in\{\FmriMmin,\dots,\FmriMmax\}$ subjects each (no family information was available; at $m=\FmriMmax$ the two halves use $2\times\FmriMmax$ of the \FmriN{} subjects, so the repeats are far from independent), each fitted with at most $\FmriSplitSweeps$ sweeps (none reached the tolerance), labelled against the group solution and matched by the Hungarian algorithm; $\delta_j=1-\mathrm{corr}^2$ between the two estimates of source $j$ under the pooled covariance, against $\hat\delta_j=2\sum_{a\neq j}v^{\rm w}_{aj}\tau_{aj}/n_{\rm half}$. The first-order region is $\hat\delta_j<\FmriPredLin$: $\hat\delta_j$ adds the errors of the two halves, so this is a per-half first-order error below $\FmriPredLinHalf$, about the regime of Section~\ref{sec:experiments}; with $\hat\delta_j<\FmriPredPointOne$ the fractions below become \FmriFracPointOneMFifty{} ($m=\FmriMsub$) and \FmriFracPointOneMFourHundred{} ($m=\FmriMmax$).

\emph{Results}. the fraction of components in the first-order region grows from \FmriFracLinMTen{} ($m=\FmriMmin$) to \FmriFracLinMTwoHundred{} ($m=200$) and \FmriLinCountMFourHundred{} of \FmriD{} ($m=\FmriMmax$); within it the Spearman correlation between $\hat\delta_j$ and $\delta_j$ is $\FmriSpearmanLinMFifty$ at $m=\FmriMsub$ and $\FmriSpearmanLinMFourHundred$ at $m=\FmriMmax$ (pooled over $m$: $\FmriSpearmanLin$, slope $\FmriSlopeLin$ on log scales), and $\delta_j/\hat\delta_j$ has median $\FmriRatioLinMedian$. The energy ranks better than the nearest-neighbour gap alone: at $m=\FmriMsub$ the Spearman correlation with $\delta_j$ is $\FmriGapSpearmanEnergyLinMsub$ for $E_j$ and $\FmriGapSpearmanNNLinMsub$ for $1/S_{\rm nn}(j)$ within the first-order region (\FmriLinCountMFifty{} components), and $\FmriGapSpearmanEnergyAllMsub$ against $\FmriGapSpearmanNNAllMsub$ over all \FmriD{} components, although the two quantities rank the components alike (Spearman $\FmriEnergyNNSpearman$); the nearest neighbour contributes a median \FmriNNShareMedian{} of $E_j$ (\FmriNNShareQten{} to \FmriNNShareQninety{} between the deciles).

Fitting $\delta_j\approx a_j/n_{\rm half}+f_j$ over $m$ gives, for the \FmriFloorFitted{} components with at least two sample sizes in the region, a floor $f_j$ with median $\FmriFloorMedian$ and coefficients $a_j$ that, with two to six points each, do not rank with the prediction (Spearman $\FmriFloorSpearman$), so the ranking of $\delta_j$ within each $m$ is carried mostly by the floor. Most of these discrepancies are close to their random level (mean $\FmriDeltaBarMFourHundred$ at $m=\FmriMmax$ against a predicted $\FmriPredBarMFourHundred$), so the fitted floor largely measures saturation. The perturbed mixings of Table~\ref{tab:stress} bias every estimate in the same way, so their floor does not show in a split-half comparison.

\begin{figure}[!tb]
\centering
\includegraphics[width=\textwidth]{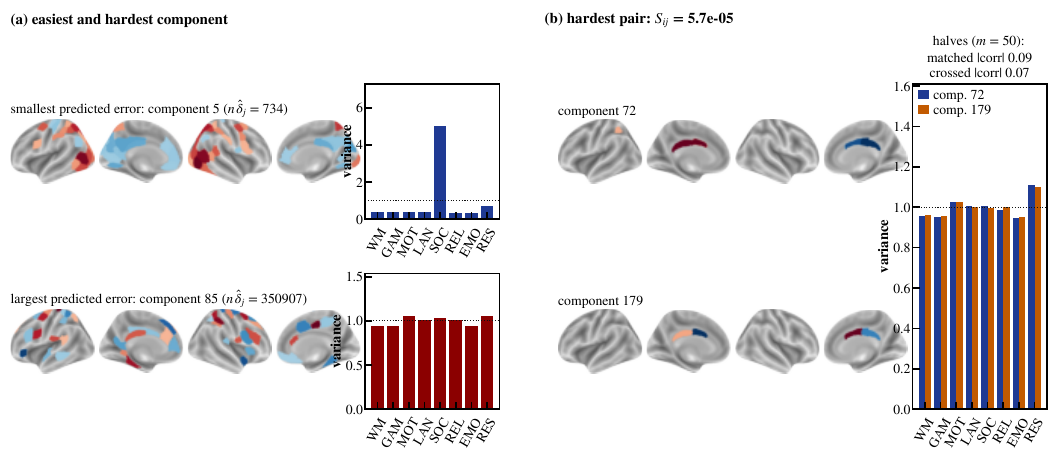}
\caption{\textbf{Real data: components ($d=\FmriD$).} \textbf{(a)}~The components with the smallest and the largest predicted error ($n\hat\delta_j=\FmriPredNmin$ and $\FmriPredNmax$): parcel loadings on the inflated cortical surface and task-variance signatures. \textbf{(b)}~The hardest pair (components \FmriHardestI{} and \FmriHardestJ, $S_{ij}=\FmriSmin$) has nearly identical signatures; between halves of $m=\FmriMsub$ subjects the matched $|\mathrm{corr}|$ of their estimates is $\FmriPairMatchedMFifty$ and the crossed one $\FmriPairCrossedMFifty$.}
\label{fig:fmrimaps}
\end{figure}

Of the $\FmriSubsetCount$ subsets of four environments, the five with the smallest energy ($\Ebar/d\approx\FmriSubsetLowEbar$) give $\delta=\FmriSubsetLowDelta$ at $m=\FmriMsub$ against $\FmriSubsetHighDelta$ for the five with the largest ($\Ebar/d\approx\FmriSubsetHighEbar$), and eight rest segments (participation ratio $\FmriRestPR$) give $\FmriRestDelta$ against $\FmriTaskDeltaSameM$ for the seven tasks and rest, whose signatures differ little from one another (cf.\ the correspondence of task and rest architectures, \citealp{smith2009correspondence}) (same $m$ and number of repeats): at this $m$ all of them are close to saturation. In Figure~\ref{fig:fmri}(b), with two environments the nearest-neighbour distances of the estimated signatures are below their estimation noise, so those energies, and their growth with $d$, are rough. Figure~\ref{fig:fmrimaps} shows the components with the smallest and the largest predicted error and the hardest pair.

\end{document}